\documentclass[10pt]{phdthesis}

\usepackage{amsthm}
\usepackage{float}
\usepackage{amsmath}
\DeclareMathOperator{\pred}{Pred}
\DeclareMathOperator{\layerv}{Layer^v}
\DeclareMathOperator{\layerf}{Layer^f}
\DeclareMathOperator{\messageFN}{Message}
\DeclareMathOperator{\updateFN}{Update}
\DeclareMathOperator{\aggregateFN}{Aggregate}
\DeclareMathOperator{\attendFN}{Attend}

\usepackage{makecell} 

\usepackage[dvipsnames]{xcolor}
\usepackage{graphicx}
\usepackage{epstopdf}
\usepackage{float}
\usepackage{bm,upgreek}
\usepackage{amsfonts}
\usepackage{enumitem}
\usepackage{mathtools}
\usepackage{multirow}
\usepackage{booktabs}
\usepackage[subnum]{cases}
\usepackage{setspace}
\usepackage{color}
\usepackage{blkarray, bigstrut} 
\usepackage[labelformat=simple]{caption}
\usepackage{subcaption}
\usepackage{xcolor}
\usepackage{pgfplots}
\pgfplotsset{compat=1.5,
  tick label style={font=\small},
  label style={font=\small}}
\usepackage{pgfplotstable}
\usepgfplotslibrary{statistics}
\usepackage{tikz}

\usepackage[european]{circuitikz}
\newcommand{\bushere}[2]{%
    coordinate(tmp)
    ++(0,1) node[above]{#1} edge[ultra thick] ++(0,-2)
    ++(0,-2) node[below]{#2}
    (tmp)
}

\newenvironment{customlegend}[1][]{%
    \begingroup
    \csname pgfplots@init@cleared@structures\endcsname
    \pgfplotsset{#1}%
}{%
    \csname pgfplots@createlegend\endcsname
    \endgroup
}%

\def\addlegendimage{\csname pgfplots@addlegendimage\endcsname}

\usepackage{pgfplots}
\usepackage{pgfplotstable}
\pgfplotsset{compat=1.7}
\usepgfplotslibrary{groupplots}

\usepackage{amsthm}
\usepackage{changepage}
\usepackage{fancyhdr}
\usepackage{amssymb,latexsym,pifont}
\usepackage{graphics}
\usepackage{float}
\usepackage{captionhack}
\usepackage{hyperref}
\usepackage{bm,upgreek}
\usepackage{enumitem}
\usepackage{parskip}
\usepackage{titlesec}
\usepackage{multirow}
\usepackage[toc,page]{appendix}
\usepackage{emptypage}
\usepackage{mathtools}
\usepackage{tikz}
\usepackage{setspace}
\usepackage{cite}
\usepackage{microtype}
\usepackage{pgfplots}
\usepackage{pgfplotstable}
\usepgfplotslibrary{statistics}
\pgfplotsset{compat=1.7,
        every axis label={font=\large}, 
        tick label style={
            /pgf/number format/assume math mode=true,
            font=\normalsize,     
        },
}
\usepgfplotslibrary{groupplots}

\usepackage{tikz}
\usetikzlibrary{spy}
\usetikzlibrary{shapes.geometric, arrows}
\usepackage{etoolbox}
\usepackage[mathscr]{euscript}
\usepackage{amsbsy}
\usepackage{euscript}
\usepackage[most]{tcolorbox}
\usepackage{arydshln}
\usepackage{booktabs}
\usepackage{graphicx}

\usepackage[ruled]{algorithm2e}

\usepackage{amsthm}
\usepackage{float}
\usepackage{amsmath}

\usepackage{makecell} 

\usepackage[dvipsnames]{xcolor}
\usepackage{graphicx}
\usepackage{epstopdf}
\usepackage{float}
\usepackage{bm,upgreek}
\usepackage{amsfonts}
\usepackage{enumitem}
\usepackage{mathtools}
\usepackage{multirow}
\usepackage{booktabs}
\usepackage[subnum]{cases}
\usepackage{setspace}
\usepackage{color}
\usepackage{blkarray, bigstrut} 
\usepackage{xcolor}
\usepackage{pgfplots}
\pgfplotsset{compat=1.5,
  tick label style={font=\small},
  label style={font=\small}}
\usepackage{pgfplotstable}
\usepgfplotslibrary{statistics}
\usepackage{tikz}

\usepackage{pgfplots}
\usepackage{pgfplotstable}
\pgfplotsset{compat=1.7}
\usepgfplotslibrary{groupplots}

\usepackage{pdfpages}

\hypersetup{colorlinks=true,citecolor=blue,linkcolor=black}

\titlespacing*{\section}      {0em}{.75em}{0.25em}
\titlespacing*{\subsection}   {0em}{1.00em}{.0em}
\titlespacing*{\subsubsection}{0em}{.50em}{.25em}
\titlespacing*{\paragraph}    {0em}{.25em}{.25em}

\tcbset{colback=blue!5,boxrule=0pt,left=5pt,right=5pt,top=3pt,bottom=3pt,
title filled, oversize,breakable,enhanced,before upper={\parindent15pt}, before
skip=15pt plus 2pt,after skip=15pt plus 2pt}

\newcommand\ph[1]{\mathscr{#1}}

\makeatletter
\def\myline{\pgfutil@ifnextchar[{\my@line}{\my@line[]}}%
\def\my@line[#1](#2)(#3){%
\tikz[overlay] \draw[#1]  (#2)--(#3);}%

\makeatletter
  \renewcommand*\env@matrix[1][*\c@MaxMatrixCols c]{%
    \hskip -\arraycolsep
    \let\@ifnextchar\new@ifnextchar
  \array{#1}}
\makeatother

\newcommand\chap[1]{%
  \chapter*{#1}%
  \addcontentsline{toc}{chapter}{#1}}
  
\makeatletter
\newcommand{\tocfill}{\cleaders\hbox{$\m@th \mkern\@dotsep mu . \mkern\@dotsep mu$}\hfill}
\makeatother

\newenvironment{abbreviations}{\begin{list}{}{%
\setlength{\labelwidth}{3cm}\setlength{\leftmargin}{\labelwidth+\labelsep}%
\setlength{\itemsep}{-3pt}}}{\end{list}}

\pgfplotsset{compat=1.5}
\pgfplotsset{grid style={dotted,gray}}
\pgfplotsset{legend image with text/.style={legend image code/.code={%
\node[anchor=west, align=right] at (0.0cm,0cm) {#1};}},}

\pgfplotsset{
  box plot/.style={/pgfplots/.cd,black,only marks,mark=-,
  mark size=\pgfkeysvalueof{/pgfplots/box plot width},
  /pgfplots/error bars/y dir=plus,/pgfplots/error bars/y explicit,
  /pgfplots/table/x index=\pgfkeysvalueof{/pgfplots/box plot x index},},
  box plot box/.style={
  /pgfplots/error bars/draw error bar/.code 2 args={%
  \draw  ##1 -- ++(\pgfkeysvalueof{/pgfplots/box plot width},0pt) 
  |- ##2 -- ++(-\pgfkeysvalueof{/pgfplots/box plot width},0pt) |- ##1 -- cycle;},
  /pgfplots/table/.cd,
  y index=\pgfkeysvalueof{/pgfplots/box plot box top index},
  y error expr={
  \thisrowno{\pgfkeysvalueof{/pgfplots/box plot box bottom index}}
  - \thisrowno{\pgfkeysvalueof{/pgfplots/box plot box top index}}},
  /pgfplots/box plot},
  box plot top whisker/.style={
  /pgfplots/error bars/draw error bar/.code 2 args={%
  \pgfkeysgetvalue{/pgfplots/error bars/error mark}%
  {\pgfplotserrorbarsmark}%
  \pgfkeysgetvalue{/pgfplots/error bars/error mark options}%
  {\pgfplotserrorbarsmarkopts}%
  \path ##1 -- ##2;},/pgfplots/table/.cd,
  y index=\pgfkeysvalueof{/pgfplots/box plot whisker top index},
  y error expr={
  \thisrowno{\pgfkeysvalueof{/pgfplots/box plot box top index}}
  - \thisrowno{\pgfkeysvalueof{/pgfplots/box plot whisker top index}}},
  /pgfplots/box plot},
  box plot bottom whisker/.style={
  /pgfplots/error bars/draw error bar/.code 2 args={%
  \pgfkeysgetvalue{/pgfplots/error bars/error mark}%
  {\pgfplotserrorbarsmark}%
  \pgfkeysgetvalue{/pgfplots/error bars/error mark options}%
  {\pgfplotserrorbarsmarkopts}%
  \path ##1 -- ##2;},
  /pgfplots/table/.cd,
  y index=\pgfkeysvalueof{/pgfplots/box plot whisker bottom index},
  y error expr={
  \thisrowno{\pgfkeysvalueof{/pgfplots/box plot box bottom index}}
  - \thisrowno{\pgfkeysvalueof{/pgfplots/box plot whisker bottom index}}},
  /pgfplots/box plot},
  box plot median/.style={/pgfplots/box plot,
  /pgfplots/table/y index=\pgfkeysvalueof{/pgfplots/box plot median index},
  semithick,black },
  box plot width/.initial=1em,
  box plot x index/.initial=0,
  box plot median index/.initial=1,
  box plot box top index/.initial=2,
  box plot box bottom index/.initial=3,
  box plot whisker top index/.initial=4,
  box plot whisker bottom index/.initial=5,}

\newcommand{\boxplot}[2][]{
    \addplot [box plot median,#1] table {#2};
    \addplot [forget plot, box plot box,#1] table {#2};
    \addplot [forget plot, box plot top whisker,#1] table {#2};
    \addplot [forget plot, box plot bottom whisker,#1] table {#2};}

\begin{document}

\pagestyle{fancyplain}
\pagenumbering{roman}
\allowdisplaybreaks

\newcommand{\publ}{}
\renewcommand{\sectionmark}[1]{\markright{\it \thesection.\ #1}}
\renewcommand{\chaptermark}[1]{\markboth{\it \thechapter.\ #1}{}}
\lhead[\thepage]{\fancyplain{\publ}{\rightmark}}
\chead[\fancyplain{}{}]{\fancyplain{}{}}
\rhead[\fancyplain{}{\leftmark}]{\fancyplain{}{\thepage}}
\lfoot[]{}
\cfoot[]{}
\rfoot[]{}
\pagenumbering{arabic}


\includepdf[pages=1-1]{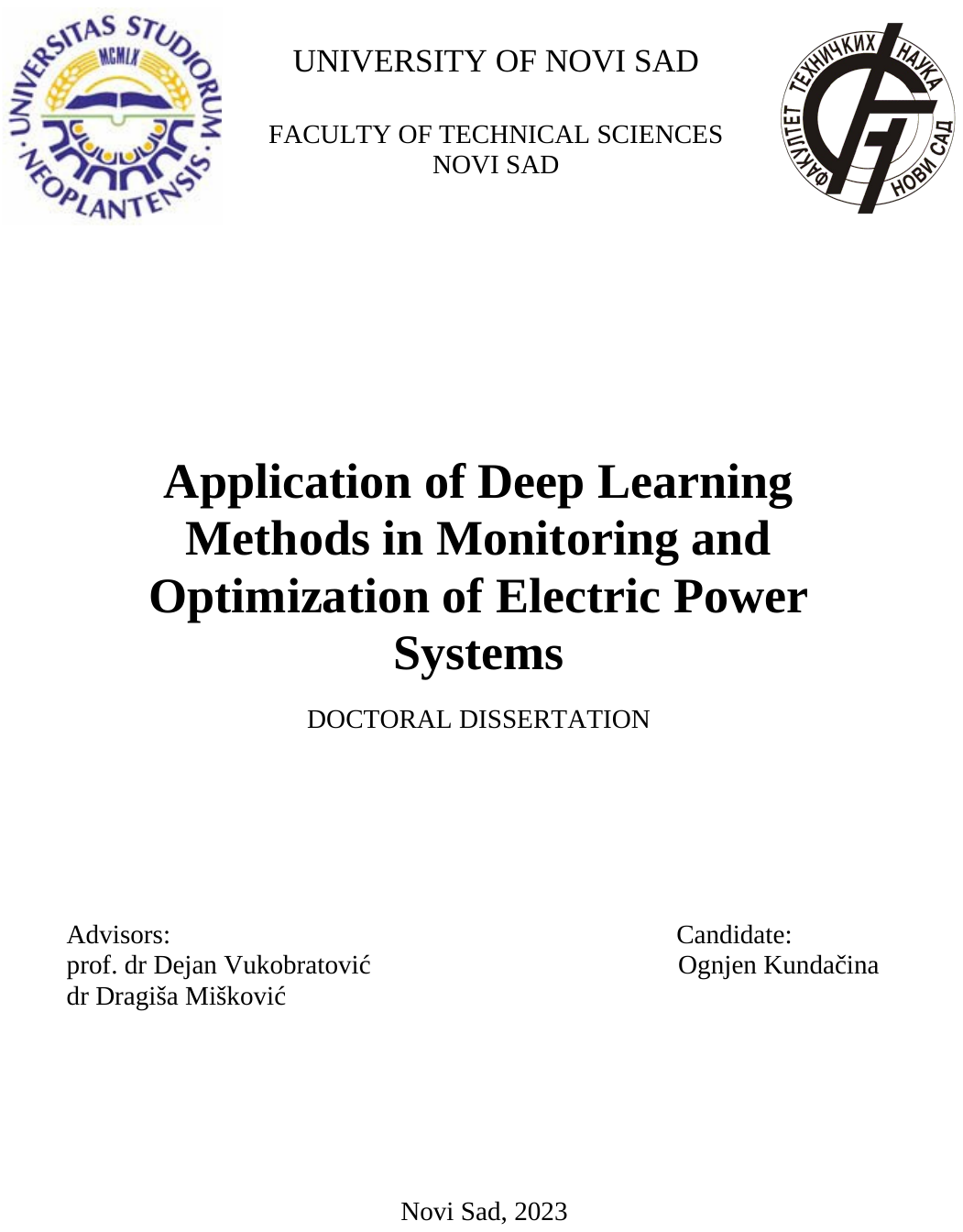}
\setcounter{page}{3} 

\thispagestyle{empty}

\begin{centering}

{\Large Application of Deep Learning Methods \\  in Monitoring and Optimization \\ \vskip 2.1mm  of Electric Power Systems}


{\large by\\
Ognjen Kunda\v{c}ina\\
}

\vskip 0.7cm

M.Sc.El.Comp.Eng. Power, Electronic and Telecommunication Engineering, \\University of Novi Sad, Serbia, 2018.\\
B.Sc.El.Comp.Eng. Power, Electronic and Telecommunication Engineering, \\University of Novi Sad, Serbia, 2017.\\

\vskip 0.7cm

for the degree of \\ \vspace{1.0cm}
{\Large Doctor of Technical Sciences}

\vskip 0.7cm
{\large A dissertation submitted to the\\
\vskip 0.3cm 
Department of Power, Electronics\\
and Communication Engineering,\\ 
Faculty of Technical Sciences,\\
University of Novi Sad,\\
Serbia.\\}

\vspace{1.0cm}


\end{centering}

\vskip 4cm
\pagebreak
\thispagestyle{empty}

{\Large Advisors:}\\
\\
\hspace*{1.0cm} {\Large Dr Dejan Vukobratovi\'c}, {Full Professor}\\  
\hspace*{1.0cm} Department of Power, Electronics and Communication Engineering,\\
\hspace*{1.0cm} University of Novi Sad, Serbia.
\\
\\
\hspace*{1.0cm} {\Large Dr Dragi\v{s}a Mi\v{s}kovi\'c}, {Science Associate}\\
\hspace*{1.0cm} The Institute for Artificial Intelligence Research and Development of Serbia.\\

\vspace*{1cm}
{\Large Thesis Committee Members:} \\
\\
\hspace*{1.0cm} {\Large Dr Tatjana Lon\v{c}ar-Turukalo}, {Full Professor}\\ 
\hspace*{1.0cm} Department of Power, Electronics and Communication Engineering,\\
\hspace*{1.0cm} University of Novi Sad, Serbia.\\
\\
\hspace*{1.0cm} {\Large Dr Milan Rapai\'c}, {Full Professor}\\ 
\hspace*{1.0cm} Department of Computing and Control Engineering,\\
\hspace*{1.0cm} University of Novi Sad, Serbia.\\
\\
\hspace*{1.0cm} {\Large Dr Predrag Vidovi\'c}, {Associate Professor}\\ 
\hspace*{1.0cm} Department of Power, Electronics and Communication Engineering,\\
\hspace*{1.0cm} University of Novi Sad, Serbia.\\
\\
\hspace*{1.0cm} {\Large Dr Mirsad \'Cosovi\'c}, {Assistant Professor},\\ 
\hspace*{1.0cm} Faculty of Electrical Engineering,\\
\hspace*{1.0cm} University of Sarajevo, Bosnia and Herzegovina.\\

\vspace*{1cm}
This research has received funding from the European Union's Horizon 2020 research and innovation programme under Grant Agreement number 856967.

\tableofcontents
\chap{List of Publications} 
\pagestyle{empty}

{\Large Journal Publications:}

\noindent
O. Kundacina, M. Cosovic, D. Miskovic, and D. Vukobratovic, “Graph Neural Networks on Factor Graphs for Robust, Fast, and Scalable Linear State Estimation with PMUs,” in Sustainable Energy, Grids and Networks, 2023.

\noindent
O. Kundacina, P. Vidovic, and M. Petkovic, “Solving dynamic distribution network reconfiguration using deep reinforcement learning,” in Electrical Engineering, 2021.

\vspace{1.2cm}
\noindent
{\Large Conference Publications:}

\noindent
O. Kundacina, M. Cosovic, D. Miskovic, and D. Vukobratovic, “Distributed Nonlinear State Estimation in Electric Power Systems using Graph Neural Networks,” in 2022 IEEE International Conference on Communications, Control, and Computing Technologies for Smart Grids (SmartGridComm), Singapore, 2022, pp. 1–6.

\noindent
O. Kundacina, M. Forcan, M. Cosovic, D. Raca, M. Dzaferagic, D. Miskovic, M. Maksimovic, and D. Vukobratovic, “Near Real-Time Distributed State Estimation via AI/ML-Empowered 5G Networks,” in 2022 IEEE International Conference on Communications, Control, and Computing Technologies for Smart Grids (SmartGridComm), Singapore, 2022, pp. 1–6.

\noindent
O. Kundacina, M. Cosovic, and D. Vukobratovic, “State estimation in electric power systems leveraging graph neural networks,” in 2022 17th International Conference on Probabilistic Methods Applied to Power Systems (PMAPS), online, 2022, pp. 1–6.

\noindent
O. Stanojev, O. Kundacina, U. Markovic, E. Vrettos, P. Aristidou, and G. Hug, “A reinforcement learning approach for fast frequency control in low-inertia power systems,” in 52nd North American Power Symposium (NAPS), online, 2021, pp. 1–6.

\noindent
O. Kundacina, G. Gojic, M. Cosovic, D. Miskovic, and D. Vukobratovic, “Scalability and Sample Efficiency Analysis of Graph Neural Networks for Power System State Estimation,” in Sixth International Balkan Conference on Communications and Networking (BalkanCom), Istanbul, 2023, pp. 1–6.

\noindent
O. Kundacina, G. Gojic, M. Mitrovic, D. Miskovic, and D. Vukobratovic, “Supporting Future Electrical Utilities: Using Deep Learning Methods in EMS and DMS Algorithms,” in 22nd International Symposium INFOTEH-JAHORINA (INFOTEH), Jahorina, 2023, pp. 1–6.

\phantomsection
\cleardoublepage
\addcontentsline{toc}{chapter}{\listfigurename}
\listoffigures

\cleardoublepage
\addcontentsline{toc}{chapter}{\listtablename}
\phantomsection
\listoftables

\chap{Abstract} 
\pagestyle{empty}

Electric power systems consist of generation, distribution, and transmission systems, which are all traditionally coordinated from the corresponding control centres. System operators use specialized software solutions for monitoring and optimization of electric power systems, installed in control centres. Typical algorithms implemented in mentioned software solutions should satisfy near real-time operation requirements, while delivering accurate information for power system monitoring and optimizing its operation.

Modern electric power systems have been increasing in size, complexity, as well as dynamics due to the growing integration of renewable energy resources, which have sporadic power generation. This necessitates the development of near real-time power system algorithms, demanding lower computational complexity regarding the power system size. Considering the growing trend in the collection of historical measurement data and recent advances in the rapidly developing deep learning field, the topic of this dissertation is the application of deep learning algorithms, namely graph neural networks (GNNs) and deep reinforcement learning (DRL), for monitoring and optimization of electric power systems.

The first part of this thesis presents a GNN approach to solving the power system state estimation (SE) problem, which aims to estimate complex bus voltages based on available measurements. Two formulations of the SE problem are considered: the first is a linear SE formulation that uses measurements from phasor measurement units (PMUs), while the second is a nonlinear SE formulation that incorporates both PMU measurements and legacy measurements from the supervisory control and data acquisition (SCADA) system.

As PMUs become more widely used in transmission power systems, a fast state estimation algorithm that can take advantage of their high sampling rates is needed. To accomplish this, we present a method that uses GNNs to solve the linear formulation of SE problem by learning complex bus voltage estimates from PMU voltage and current measurements. We propose an original implementation of GNNs over the power system's factor graph to simplify the integration of various types and quantities of measurements on power system buses and branches. Furthermore, we augment the factor graph to improve the robustness of GNN predictions. The proposed GNN model is highly efficient and scalable, as its computational complexity is linear with respect to the number of nodes in the power system. Training and test examples were generated by randomly sampling sets of power system measurements and annotating them with the exact solutions of linear SE with PMUs, obtained using a traditional weighted least squares-based method. The numerical results demonstrate that the GNN model provides an accurate approximation of the SE solutions. Furthermore, errors caused by PMU malfunctions or communication failures that would normally make the SE problem unobservable have a local effect and do not deteriorate the results in the rest of the power system.

Alongside the linear SE problem formulation, in this thesis, we consider nonlinear SE, which takes into account all types of measurements available in the power system, and is usually solved using the iterative Gauss-Newton (GN) method. The nonlinear SE formulation presents some difficulties when considering inputs from both PMUs and SCADA system. These include numerical instabilities, convergence time depending on the starting point of the iterative method, and the quadratic computational complexity of a single iteration regarding the number of state variables. Analogously to the GNN-based linear SE, we apply GNN over the augmented factor graph of the nonlinear power system SE. Once trained, the proposed regression model has linear computational complexity during the inference time, with a possibility of distributed implementation. Since the method is noniterative and non-matrix-based, it is resilient to the problems that the GN solver is prone to. In addition to good prediction accuracy on the test set, the proposed model demonstrates robustness during the simulation of cyberattacks and unobservable scenarios due to communication irregularities.

In the second part of this thesis, we focus on distribution network reconfiguration (DNR), which is critical for enhancing energy efficiency by coordinating switch operations in the distribution network. The sufficient number of remote switching devices in the distribution network enables dynamic distribution network reconfiguration (DDNR), which determines the optimal network topologies over a specified time interval. To achieve this, we propose a data-driven approach for DDNR using DRL. The proposed DDNR controller aims to minimize the objective function which includes active energy losses and the cost of switching manipulations, while ensuring that all constraints are satisfied. The following constraints are considered: allowed bus voltages, allowed line apparent powers, a radial network configuration with all buses being supplied, and the maximal allowed number of switching operations. This optimization problem is modelled as a Markov decision process by defining the possible states and actions of the DDNR agent (controller) and rewards that lead the agent to minimize the objective function while satisfying the constraints. Switching operation constraints are modelled by modifying the action space definition instead of including the additional penalty term in the reward function, to increase the computational efficiency. The proposed algorithm was tested on three test examples: small benchmark network, real-life large-scale test system and IEEE 33-bus radial system and the results confirmed the robustness and scalability of the proposed algorithm.
\chap{Abbreviations} 
\pagestyle{empty}


\begin{abbreviations}

\item[CNN] Convolutional neural network
\item[DDNR] Dynamic distribution network reconfiguration
\item[DMS] Distribution management system
\item[DNN] Deep neural network
\item[DNR] Distribution network reconfiguration
\item[DRL] Deep reinforcement learning
\item[DQN] Deep Q-network
\item[EMS] Energy management system
\item[GAT] Graph attention network
\item[GN] Gauss-Newton
\item[GNN] Graph neural network
\item[GRU] Gated recurrent unit
\item[LSTM] Long short-term memory
\item[MADRL] Multi-agent deep reinforcement learning
\item[MDP] Markov decision process
\item[MSE] Mean square error
\item[PMU] Phasor measurement unit
\item[ReLU] Rectified linear unit
\item[RNN] Recurrent neural network
\item[RL] Reinforcement learning
\item[SCADA] Supervisory control and data acquisition
\item[SE] State estimation
\item[WAMS] Wide area measurement system
\item[WLS] Weighted least-squares
\item[5G] Fifth-Generation

\end{abbreviations}

\pagestyle{fancyplain}

\chapter{Introduction}\label{ch:introduction}
\addcontentsline{lof}{chapter}{1 Introduction}

Power systems are undergoing a transition due to the increased integration of renewable energy resources, and as a result they are facing new challenges in their operations. These challenges include the unpredictable nature of renewable energy resources, maintaining stability within the power system, managing the impacts of distributed generation, and the challenges presented by reverse power flows \cite{Aguero2017Challenges}. Consequently, the mathematical formulations of traditional algorithms that solve these problems have become increasingly complex and nonlinear, with larger dimensionality, making their practical implementation and real-time operation more challenging. These algorithms are usually implemented as parts of specialized software solutions, such as energy management systems (EMS) for transmission networks and distribution management systems (DMS) used in distribution networks, which are installed in power system control centres and used by power system operators on a daily basis. Some of the algorithms typically used as EMS and DMS functionalities include state estimation (SE), fault detection and localization, demand and generation forecast, voltage and transient stability assessment, voltage control, optimal power flow, economic dispatch, etc. Increasing amounts of data generated by power systems \cite{Rusitschka2010SGDataCloud} and collected by EMS and DMS are enabling the development of new deep learning-based algorithms to overcome the limitations of traditional ones.

Deep learning is a subfield of artificial intelligence that involves training neural network models to find patterns and make predictions based on the available set of data samples \cite{GoodBengCour16}. Some of the advantages of employing deep learning methods in the field of power systems include:

\begin{itemize}
    \item Speed: Once trained, a deep learning algorithm usually operates quickly, even when processing large amounts of data \cite{Sarker2021DeepLA}. This is crucial for applications where fast decision-making is required, as is the case in many power system operation problems.
    \item Accuracy: Universal approximation theorem \cite{hornik1989UniversApprox} states that neural networks can approximate any function to a desired degree of accuracy, if it consists of a sufficient number of trainable parameters. Practically, this implies that neural networks can be employed to tackle a wide range of problems, including those in power systems, and that different network architectures and sizes can be used to adapt to the complexity of the problem.
    \item Adaptability: Deep learning methods are easily adaptable, meaning that they can be retrained when the underlying data generation process changes \cite{datasetShift}. This makes them suitable for dynamic environments, such as when the power system's operating conditions change.
    \item Robustness: Traditional model-based algorithms can encounter problems when faced with uncertain or unreliable power system parameters \cite{antona2018PSUncertainty}. As a model-free alternative, deep learning methods alleviate these problems by not relying on power system parameters.
    \item Automation: Since deep learning algorithms can learn from human responses in various situations given enough training data, they can be used to reduce the need for human intervention in certain power system tasks. For instance, in applications such as predictive maintenance \cite{Zhang2019PredictiveMainta}, which are integral parts of asset management systems, deep learning can be applied within an automated real-time monitoring system.
\end{itemize}

In the continuation, we shortly introduce the basic deep learning terminology, describe the most common deep learning approaches and review their recent applications in the field of monitoring and optimization of electric power systems \cite{kundacinaInfoteh}.

\section{Deep Learning Fundamentals}
\label{sec:dl}
\begin{figure}[htbp]
    \centerline{\includegraphics[width=4.0in,trim={1.8cm 6cm 7.2cm 3.7cm},clip]{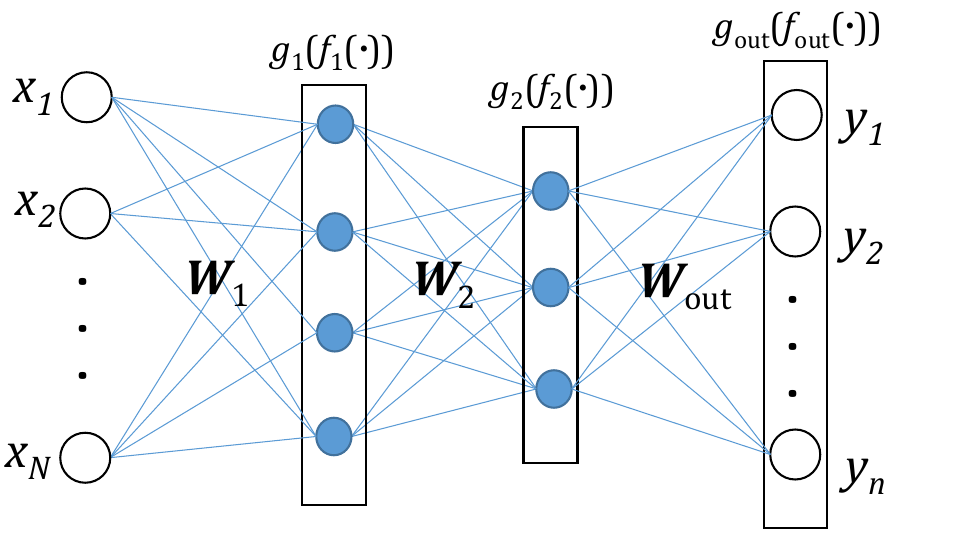}}
    \caption{A simple fully connected neural network containing an input layer, two hidden layers, and an output layer.}
    \label{fig_nn}
\end{figure}

Deep learning is a field of machine learning that involves training neural networks on large datasets \cite{GoodBengCour16}, with a goal of generating accurate predictions on unseen data samples. Therefore, neural networks can be seen as trainable function approximators, composed of interconnected units called neurons, which process and transmit information. In a simple fully connected neural network, the information processing is organized in layers, where input information from the previous layer is linearly transformed using a function $f_i(\cdot)$, where $i$ denotes the layer index. The linear transformation is defined using a matrix of trainable parameters $\mathbf{W_i}$, i.e., the weights of the connections between the neurons, shown in Fig.~\ref{fig_nn}. Trainable parameters also include biases, which are free terms associated with each neuron, and are omitted in the figure. The information is then passed through a nontrainable nonlinear function $g_i(\cdot)$ to create the outputs of that layer. Inputs and outputs of the whole neural network are denoted as $\mathrm{x}_{j}$ and $y_k$ in Fig.~\ref{fig_nn}, where $j$ and $k$ denote the indices of input and output neurons.

Neural network training assumes adjusting the trainable parameters (i.e., weights and biases of the neurons) using the knowledge in the collected data, so that accurate predictions can be performed based on the new inputs. The training process is formulated as an optimization problem which searches through the trainable parameter space to minimize the distance function between the predicted output and the true output. The problem is usually solved using gradient-based optimization methods such as gradient descent, or some of its variants \cite{KingmaB14Adam}.

In practice, when using deep learning to solve a problem, it is common to train multiple instances with different neural network model structures. This structure is defined by hyperparameters, such as the number of layers and the number of neurons in each layer. By finding the optimal set of hyperparameters, the neural network structure that best fits the problem being solved can be identified. The hyperparameter search can be done manually or with the use of specialized optimization methods \cite{Bergstra2013MakingAS}. Commonly, the collected data is split into three sets: a training set, a validation set, and a test set. The training set is used in a neural network training process, the validation set is used to evaluate the performance of a single training instance, and the test set is used to evaluate the overall performance of the trained model.

Adjusting the architecture of a deep learning model to match the structure of the input data can enhance training speed and performance and reduce required training data. For example, convolutional neural networks (CNNs) use shared parameters to process grid data, exploiting local relations between neighboring pixels and achieving spatial translation invariance. Recurrent neural networks (RNNs) use shared parameters to process sequential data, resulting in time translation invariance, while graph neural networks (GNNs) aim for permutation invariance and are particularly efficient when applied to graph structured data. Since ordinary, fully connected neural networks have been widely used for solving power systems problems, we focus on applications of more advanced deep learning architectures.

\section{Convolutional Neural Networks}
Convolutional Neural Networks are a well studied class of deep learning architectures primarily designed for analysing spatial patterns in grid-structured data such as images \cite{GoodBengCour16}. They consist of multiple convolutional layers, each of which acts as a trainable convolutional filter that extracts local information from the image, transforms it into more abstract, grid-shaped representations, and feeds it into the succeeding layer. Applying multiple CNN layers enables CNN to extract useful features from an image, which can then be used for various tasks such as classification or regression.

Although power system data is not inherently arranged in the format of an image, CNNs have been effectively used to address power system problems, mostly involved with processing data sequences. To meet the requirements of CNNs, power system data is transformed and reshaped in various ways, some of which include:
\begin{itemize}
    \item One approach for dealing with the time-varying nature of power systems is to utilize 1D CNNs on univariate time series data. For example, in study \cite{Poudyal2021ConvolutionalNN}, 1D CNNs were used to predict power system inertia using only frequency measurements. The process involves stacking time series of changes in frequency measurements, along with their rates of change, into a one-dimensional array and then processing it using 1D CNNs.
    \item A more effective method is to group signals into a matrix, where each row represents a single univariate signal. By using a 2D CNN to process this matrix, we can perform multivariate time series analysis, which allows us to analyse patterns across multiple time series and how they interact with each other. This approach has been used in recent research, such as in the study \cite{Alqudah2022CnnFaults}, to detect faults in power systems through analysing series of voltage, current, and frequency measurements.
    \item Time series data can be subjected to time-frequency transformation, allowing for analysis of the frequency content of the signal while maintaining its temporal localization. These transformations can be visually represented in two dimensions, and therefore can be analysed using various image processing tools, including CNNs. For instance, in \cite{Scalograms2020} a CNN was trained to classify faults in power systems by analysing 2D scalograms, which were generated by applying the continuous wavelet transform to time series of phasor measurements.
    \item Another approach is to use a CNN over the matrix of electrical quantities created for a single time instance, where each row contains the values of a specific electrical quantity for each power system element. This approach, which does not consider time series data, has been shown to be effective in certain applications. The study \cite{DCOPF_CNN} solves the DC optimal power flow problem by using this approach and taking node-level active and reactive power injections as inputs, with labels obtained using the traditional DC optimal power flow approach.
\end{itemize}

It's important to note that these approaches use only aggregated inputs from all the elements of the power system, without considering the connectivity between them.

\section{Recurrent Neural Networks}
Recurrent neural networks represent a significant development in deep learning algorithms, particularly in the processing of sequential data such as speech, text, and time series. \cite{GoodBengCour16}. Each of the recurrent layers acts as a memory cell that takes in information from previous steps in the sequence, processes it, and generates a hidden state representation that is passed on to the next step. The final hidden state of RNNs encapsulates the information of the entire input sequence and can be applied to tasks such as natural language processing, speech recognition, and time-series prediction. While 1D CNNs are limited to fixed length sequences, meaning that all time series in the training and test samples must have the same number of elements, RNNs are adaptable to varying sequence lengths, making them more versatile and useful for analysing sequential data.

The fundamental building blocks of RNNs are memory units, such as gated recurrent units (GRUs) and long short-term memory units (LSTMs) \cite{LSTMvsGRU}. These architectures are created to tackle the challenge of longer-term dependencies in sequential data. Both GRUs and LSTMs include an internal memory, which allows them to selectively retain or discard information from previous steps in the sequence, thus enhancing their ability to handle inputs of varying lengths. LSTMs are more complex and powerful, capable of handling longer-term dependencies, while GRUs are computationally simpler and faster, yet may not be as effective in certain tasks.

In the field of power demand and generation forecasting, various time series prediction algorithms, including RNNs, have been utilized. One recent study, \cite{MissingDataTolerantPVForecasting} uses LSTM RNNs to predict multistep-ahead solar generation based on recorded measurement history while also addressing missing records in the input time series. RNNs can also be used to predict the flexibility of large consumers' power demand in response to dynamic market price changes, as demonstrated in \cite{SiameseLSTMDemandFlexibility}. This approach combines two LSTM RNNs, one for predicting market price and the other for predicting a consumer's demand flexibility metric, with a focus on uncommon events such as price spikes. An interesting technical aspect of this method is that the two RNNs share some LSTM-based layers, resulting in more efficient and faster training, as well as improved prediction capabilities.

RNNs can also be applied to other data available in DMS and EMS, other than power and energy. The work \cite{BidirectionalLSTMVoltageStability} proposes using an RNN to classify the voltage stability of a microgrid after a fault, using time series of measurement deviations, providing power system operators with valuable information, needed to take corrective actions. The employed RNN architecture is the bidirectional LSTM, which processes the time series data in both forward and backward directions, allowing the RNN to consider both past and future context in each step of the sequence when making predictions. In the study \cite{FELLNER2022100851}, the authors evaluate different deep learning models for detecting misconfigurations in power systems using time series of operational data. They compare GRU RNN, LSTM RNN, the transformer architecture \cite{attentionAllYouNeed}, which has been successful in natural language processing tasks, and a hybrid RNN-enhanced transformer \cite{RTransformer}. The results show that the RNN-enhanced transformer is the most effective architecture, highlighting the potential of attention-based architectures for solving time series problems in power systems.

\section{Graph Neural Networks}
Graph Neural Networks, particularly spatial GNNs that utilize message passing, are an increasingly popular deep learning technique that excels at handling graph structured data, which makes them well-suited for addressing a wide range of power systems problems. Spatial GNNs process graph structured data by repeatedly applying a process called message passing between the connected nodes in the graph \cite{GraphRepresentationLearningBook}. The goal of GNNs is to represent the information from each node and its connections in a higher-dimensional space, creating a vector representation of each node, also known as node embeddings. GNNs are made up of multiple layers, each representing one iteration of message passing. Each message passing iteration is performed by applying multiple trainable functions (implemented as neural networks) such as a message function, an aggregation function, and an update function. The message function calculates the messages being passed between two node embeddings, the aggregation function combines the incoming messages in a specific way to create an aggregated message, and the update function calculates the update to each node's embedding. This process is repeated a predefined number of times, and the final node embeddings are passed through additional neural network layers to generate predictions.

GNNs have several advantages over the other deep learning architectures when used in power systems. One of them is their permutation invariance property, which means that they produce the same output for different representations of the same graph by design. GNNs are able to handle dynamic changes in the topology of power systems and can effectively operate over graphs with varying numbers of nodes and edges. This makes them well suited for real-world power systems, which may have varying topologies. Additionally, GNNs are computationally and memory efficient, requiring fewer trainable parameters and less storage space than traditional deep learning methods applied to graph-structured data, which is beneficial in power system problems where near real-time performance is critical. Spatial GNNs have the ability to perform distributed inference with only local measurements, which makes it possible to use the 5G network communication infrastructure and edge computing to implement this effectively \cite{kundacina5G2022}. This enables real-time and low-latency decision-making in large networks as the computations are done at the network edge, near the data source, minimizing the amount of data sent over the network.

GNNs have recently been applied to a variety of regression or classification tasks in the field of power systems. The work \cite{chen2020FaultLocation} proposes using GNNs over the bus-branch model of power distribution systems, with phasor measurement data as inputs, to perform the fault location task by identifying the node in the graph where the fault occurred. The use of GNNs for assessing power system stability has been explored in \cite{ZHANG2022Stability}, where the problem is formulated as a graph-level classification task to distinguish between rotor angle instability, voltage instability, and stability states, also based on power system topology and measurements. The paper \cite{ARASTEHFAR2022LoadFOrecasting} presents a hybrid neural network architecture which combines GNNs and RNNs to address the Short-Term Load Forecasting problem. The RNNs are used to process historical load data and provide inputs to GNNs, which are then used to extract the spatial information from users with similar consumption patterns, thus providing a more comprehensive approach to forecast the power consumption. In \cite{Zhao2022Transient} the authors propose a GNN approach for predicting the power system dynamics represented as time series of power system states after a disturbance or failure occurs. The GNN is fed with real-time measurements from phasor measurement units that are distributed along the nodes of the graph. In \cite{takiddin2022FalseAttacks} GNNs are applied over varying power system topologies to detect unseen false data injection attacks in smart grids.

In the previously mentioned studies, GNNs have been applied to the traditional bus-branch model of power systems, however, a recent trend in the field has been to apply GNNs over other topologies representing the connectivity in power system data. As it will be further discussed in this thesis, GNNs can be applied in combination with heterogeneous power system factor graphs to solve the SE problem, both linear \cite{kundacina2022state} and nonlinear \cite{kundacina2022NonlinearSE}. In these approaches, measurements are represented using factor nodes, while variable nodes are used to predict state variables and calculate training loss. These approaches are more flexible regarding the input measurement data compared to traditional deep learning-based SE methods because they provide the ability to easily integrate or exclude various types of measurements on power system buses and branches, through the addition or removal of the corresponding nodes in the factor graph. A different approach that does not use the GNN over the traditional bus-branch model is presented in \cite{Yuan2022LearningInterractions}. The proposed method solves the power system event classification problem based on the collected data from phasor measurement units. The approach starts by using a GNN encoder to infer the relationships between the measurements, and then employs a GNN decoder on the learned interaction graph to classify the power system events.

\section{Deep Reinforcement Learning}
So far, we have reviewed deep learning methods that are inherently suited for predicting discrete or continuous variables based on a set of inputs. In contrast, deep reinforcement learning (DRL) methods have a direct goal of long-term optimization of a series of actions that are followed by immediate feedback\cite{Sutton1998}. Therefore, DRL methods are powerful tools for multi-objective sequential decision-making, suitable for application in various EMS and DMS functionalities that involve power system optimization \cite{Glavic2019}. In the DRL framework, the agent interacts with the stochastic environment in discrete time steps and the goal is to find the optimal policy that maximizes the long-term reward while receiving feedback about its immediate performance. The agent receives state variables from the environment, takes an action, receives an immediate reward signal and the state variables for the next time step. The DRL training process involves many episodes that include agent-environment interaction, during which the agent learns by trial and error. Using the collected data from these episodes, the agent is able to predict the long term rewards in various situations using neural networks, and these predictions are then used to generate an optimal decision-making strategy.

There are many studies that apply DRL in the field of power system optimization and control. Some of the examples include distribution network reconfiguration \cite{Gao2019DDNR}, Volt-VAR control in power distribution systems \cite{wang2021SafeVVC}, frequency control in low-inertia power systems \cite{stanojev2021FFC}, and so on. In these studies, an RL agent receives various electrical measurements as state information and takes a single multidimensional action per time step, which includes both discrete and continuous set points on controllable devices within a power system.

A recent trend in the power system research is transitioning from single agent to multi-agent deep reinforcement learning (MADRL), which is based on coordinating multiple agents operating together in a single environment using the mathematical apparatus developed in the field of game theory \cite{MultiAgentRL}. MADRL relies on centralized training and decentralized execution concept, where a centralized algorithm is responsible for training all the agents at once, allowing for coordination and cooperation among the agents. This centralized training approach results in faster real-life execution due to significantly reduced communication delays during decentralized execution, where each agent can act independently based on the knowledge acquired during the centralized training. Reducing these communication delays is particularly important in large transmission power systems where the individual agents may be significantly geographically separated.

For example, a decentralized Volt-VAR control algorithm for power distribution systems based on MADRL is proposed in \cite{MARLVVC}. In this algorithm, the power system is divided into multiple independent control areas, each of which is controlled by a corresponding DRL agent. These agents observe only the local measurements of electrical quantities within their corresponding area, and the action of each agent contains set points on all the reactive power resources in that area. Similarly, in \cite{PowerNet}, a MADRL algorithm is used to solve the secondary voltage control problem in isolated microgrids in a decentralized fashion by coordinating multiple agents, each of which corresponds to a distributed generator equipped with a voltage-controlled voltage source inverter. The action of each agent is a single secondary voltage control set point of the corresponding generator. The fundamental difference compared to \cite{MARLVVC} is that the agent in \cite{PowerNet} uses not only the local measurements of electrical quantities for the state information, but also messages from the neighbouring agents, leading to improved performance. Work \cite{MARL_EconomicDispatch} proposes using a MADRL algorithm to perform the economic dispatch, which minimizes the overall cost of generation while satisfying the power demand. The agent models an individual power plant in a power system, with the action being the active power production set point. Another example of using MADRL for an economic problem in coupled power and transportation networks is given in \cite{MADRL_EVCharging}. A MADRL method is proposed to model the pricing game and determine the optimal charging pricing strategies of multiple electric vehicle charging stations, where each individually-owned EV charging station competes using price signals to maximize their respective payoffs. In all the aforementioned works, multiple agents are trained in a centralized manner to optimize the reward function defined globally based on the nature of the particular problem at hand.

\section{Power System State Estimation using Graph Neural Networks}

The power system state estimation is a problem of determining the state of the power system represented as the set of complex bus voltages, given the available set of measurements \cite{monticelli2000SE}. The dominant part of the input data for the SE model consists of legacy measurements coming from the supervisory control and data acquisition (SCADA) system, which have relatively high variance, high latency, and low sampling rates. Increasingly deployed phasor measurement units (PMUs), provided by the wide area measurement system (WAMS), have low variance and high sampling rates and are a potential enabler of real-time system monitoring. There are two main SE formulations that emerge based on the type of input measurements taken into account:
\begin{itemize}    
    \item \textbf{Nonlinear SE:} Taking into account both legacy and phasor measurements results in the SE model formulated by the system of nonlinear equations and is traditionally solved using the iterative Gauss-Newton (GN) method \cite{monticelli2000SE}. Different approaches can be used to integrate phasor measurements into the well established model with legacy measurements. A standard way to include voltage and current phasors coming from PMUs is to represent them in the rectangular coordinate system \cite{exposito}. The main disadvantage of this approach is related to measurement errors, where measurement errors of a single PMU are correlated, and the covariance matrix does not have diagonal form. Despite that, because of the lower computational effort, the measurement error covariance matrix is usually considered as diagonal matrix, which has the effect on the accuracy of the nonlinear SE. The diagonal form of the covariance matrix could be preserved by representing voltage and current phasors coming from PMUs in the polar coordinate system, which requires a large computational effort with a convergence time significantly depending on the state variables' initialization \cite{manousakis}. Additionally, using magnitudes of branch current measurements can cause numerical instabilities such as undefined Jacobian elements due to the ``flat start" \cite[Sec. 9.3]{abur2004power}. Furthermore, different orders of magnitude of phasor and legacy measurement variances can make the SE problem ill-conditioned by increasing the condition number of the estimator's gain matrix \cite{exposito}. A single iteration of the GN method involves solving a system of linear equations, which results in near $\mathcal{O}(n^2)$ computational complexity for sparse matrices, where $n$ is the number of power system buses.
    \item \textbf{Linear SE:} When a sufficient number of PMUs is installed in a power system, the SE algorithm can consider only phasor measurements as inputs, without the need to include legacy measurements in the calculation. In this case, the SE problem can then be expressed as a system of linear equations if both state variables and phasor measurements are represented in a rectangular coordinate system. This approach provides non-iterative solutions which are faster than the nonlinear SE, and utilize high sampling rates of PMUs more. Solving linear SE is traditionally done by solving a linear weighted least-squares (WLS) problem \cite{exposito}, which involves matrix inversions or factorizations, which can be difficult in cases where the matrix is ill-conditioned due to varying orders of magnitudes of power system parameters. It is common practice to neglect the phasor measurement covariances represented in rectangular coordinates \cite{exposito}. This can make the SE problem much easier to solve, but it also results in a computational complexity of nearly $\mathcal{O}(n^2)$ for sparse matrices. 
\end{itemize}

In both SE problem formulations, real-time monitoring of large power systems can be challenging using traditional approaches due to their high computational complexity of $\mathcal{O}(n^2)$ and mentioned numerical difficulties associated with them. Recent advancements in GNNs \cite{pmlr-v70-gilmer17a, GraphRepresentationLearningBook} open up novel possibilities for developing power system algorithms with linear computational complexity and potential distributed implementation. GNNs (as well as other deep learning methods) can be particularly useful for the SE problem because they are not based on the matrix model of the power system, which eliminates numerical difficulties associated with traditional SE solvers. These approaches, when trained on relevant datasets, are able to provide solutions even when traditional methods fail.

Generally, the popularity of deep learning in the field of power systems analysis has been well-documented in recent research, with several studies showing that it can be used to learn the solutions to computationally intensive algorithms such as power system SE. In \cite{zhang2019}, the authors used a combination of recurrent and feed-forward neural networks to solve the SE problem using measurement data and the history of network voltages. Another study, \cite{zamzam2019}, provides an example of training a feed-forward neural network to initialize the network voltages for a Gauss-Newton power distribution system SE solver.

As the use of GNNs in power systems becomes more common, several studies suggest applying GNNs to power flow problems, which are similar to the SE problem in some aspects. In \cite{donon2019graphneuralsolver} and \cite{bolz2019PFapproximator}, power flows in the system are predicted based on power injection data labelled by a traditional power flow solver. Similarly, \cite{wang2020ProbabPF} suggests using a trained GNN as an alternative to computationally expensive probabilistic power flow methods, which calculate probability density functions of unknown variables. Different approaches propose training a GNN in an unsupervised manner to perform power flow calculations by minimizing the violation of Kirchhoff's law \cite{donon2020graphneuralsolverJOURNAL} or power balance error \cite{LOPEZ2023GNNpowerFlow} at each bus, thus avoiding the need for labelled data from a conventional power flow solver. 

In \cite{pagnier2021physicsinformed}, the authors propose a combined model- and data-based approach using GNNs for power system parameter and state estimation. The model predicts power injections and consumptions in nodes where voltage and phase measurements are taken, but it does not consider branch measurements and other types of node measurements in its calculations. In \cite{Yang2022RobustPSSEDataDrivenPriors}, the authors train a GNN by propagating simulated or measured voltages through the graph to learn the voltage labels from a historical dataset, and then use the GNN as a regularization term in the nonlinear SE loss function. However, the proposed GNN only uses node voltage measurements and does not consider other types of measurements, although they are handled in other parts of the algorithm. Another feed-forward neural network learns the solutions that minimize the SE loss function, resulting in an acceleration of the nonlinear SE solution with $\mathcal{O}(n^2)$ computational complexity at inference time. In \cite{TGCN_SE_2021}, state variables are predicted based on a time-series of node voltage measurements, and the authors solve the nonlinear SE problem using GNNs with gated recurrent units.

\textbf{Contributions:} This thesis proposes specialized GNN models for solving linear and nonlinear SE problems in positive sequence power transmission systems. To provide fast and accurate predictions during the evaluation phase, GNNs is trained using the inputs and solutions from traditional SE solvers. The following are the main contributions of our work regarding GNN-based SE, also published in \cite{kundacina2022state, kundacina2022Elsevier} and \cite{kundacina2022NonlinearSE}:
\begin{itemize}[leftmargin=*]
\item Inspired by \cite{Satorras2021NeuralEB}, we present the first use of GNNs on factor graphs \cite{Kschischang2001FactorGraphs} for the SE problem, instead of using the bus-branch power system model. This enables trivial integration and exclusion of any type and number of measurements on the power system buses and branches, by adding or removing the corresponding nodes in the factor graph, and therefore is applicable to both linear and nonlinear SE problem formulations. Furthermore, the factor graph is augmented by adding direct connections between variable nodes that are $2^{nd}$-order neighbours to improve information propagation during neighbourhood aggregation, particularly in unobservable scenarios when the loss of the measurement data occurs.
\item We present a graph attention network (GAT) \cite{velickovic2018graph} model, with the architecture customized for the proposed heterogeneous augmented factor graph, to solve the SE problem. GNN layers that aggregate into factor and variable nodes have separate sets of trainable parameters. Furthermore, separate sets of parameters are used for variable-to-variable and factor-to-variable message functions in GNN layers that aggregate into variable nodes.
\item Given the sparsity of the power system's graph, and the fact that node degree does not increase with the total number of nodes, the proposed approach has $\mathcal{O}(n)$ computational complexity, making it suitable for large-scale power systems. The inference of the trained GNN is easy to distribute and parallelize. Even in the case of centralized SE implementation, the processing can be done using distributed computation resources, such as graphical-processing units.
\item We demonstrate that the number of trainable parameters in the proposed GNN-based SE model is constant, while it grows quadratically with the number of measurements in conventional deep learning approaches.
\item We evaluated the performance of the proposed method by testing on various data samples, including unobservable cases caused by communication errors or measurement device failures, and scenarios corrupted by malicious data injections. Furthermore, we study the local-processing nature of the proposed model and show that significant degradation of results in these scenarios affects only the local neighbourhood of the node where the failure or malicious data injection occurred.
\item In addition to the standalone application, the proposed GNN-based nonlinear SE can be used as a fast and accurate initializer of the GN method by providing it with a starting point near the exact solution.
\end{itemize}

\section{Dynamic Distribution Network Reconfiguration based on Deep Reinforcement Learning}  

Distribution network reconfiguration (DNR) is a widely used distribution system optimization procedure, which has the goal of finding the optimal topology of the distribution network by manipulating the statuses of switching devices. Primarily, DNR is used to achieve objectives such as the minimization of power loss and voltage deviations \cite{Samman2020FastNr, Fathi2020Reconfig}, and secondarily, it can be used for load balancing, Volt-VAR optimization, supply restoration, etc. \cite{Mishra2017ACR}. Therefore, DNR is an important feature of software systems used for distribution network management. In the research literature, DNR is a common name for a static formulation of the DNR problem, which is determined for the fixed operation point, defined by load and generation in the one time instance. In the cases of limited distribution network automation, static DNR is performed once per interval ranging up to one year, due to the time-variability of the distribution system state. The number of possible solutions to the static DNR problem is $2^{N_{sw}}$, where $N_{sw}$ is the number of switches available for network reconfiguration. The other formulation of the DNR problem is the dynamic distribution network reconfiguration (DDNR), which optimizes the network operations over the specified time period. This makes DDNR suitable for real-time applications due to the time-varying nature of load, generation, and other network conditions. Using DDNR instead of static DNR can result in larger benefits and increased network operation performance, with the drawback of requiring the use of the fully automated distribution network. Since the aim of developing DDNR is to be executed more often than the static DNR, it must consider the number of switching manipulations in the cost and the constraint functions. The large number of switching manipulations can reduce the life span of switching devices and cause instability in the distribution network operation in the case of complex topology changes. Therefore, DDNR resolves the trade-off between performing the optimal reconfiguration too often and changing the network topology less frequently to reduce the number of switching manipulations. The DDNR problem introduces discretization of the optimization horizon into $T$ time intervals, resulting in increased problem complexity with the total of $2^{N_{sw}T}$ possible solutions. Since operation planning based on DDNR can be performed daily or every hour, the aim of this work is to develop a DDNR algorithm that optimizes the operation of the distribution network and is fast during the evaluation time, making it applicable to real-world scenarios.

The first studies on DNR address problems such as power loss reduction \cite{civanlarFeederReconfiguration, shormohamadi1989Reconfig, baran1989Recondig} and load balancing \cite{baran1989Recondig} among distribution feeders in the scope of static DNR. Some extensions of the standard DNR problem for the optimization over time period have been developed for the specialized cases like energy loss reduction \cite{taleski1997EnergyLosses}, and operation cost reduction \cite{Zhou1997OperationCosts}. The static DNR is often solved using the standard deterministic optimization tools from the classical optimization theory. For example, DNR based on the mixed-integer linear programming is presented in \cite{Borghetti2015MixedInteger, Ahmadi2015DistribSystemOptim, Lavorato2012Imposing, Jabr2012Minimal, Haghighat2016Distribution}, with the main advantage of finding the global optimum using the standard solvers, with the expense of computational complexity. The "path-to-node" concept for DNR proposed in \cite{ramos2005PathBased} efficiently models the radiality of the distribution network and solves the formulated problem using a mixed-integer linear programming solver. The main drawback of this concept is the significant increase in the number of decision variables with an increase in the optimization problem dimension. In \cite{khodr2009Banders}, DNR integrated with the optimal power flow based on the Benders decomposition approach is presented. 

Heuristic methods utilize physics and engineering knowledge about a specific problem to produce practically effective solutions. One type of these methods used for the DNR problem performs a heuristic search through the distribution network topologies by opening switches with minimal current flows obtained from power flow solutions \cite{shormohamadi1989Reconfig, merlin1975SearchFor, borozan1995ImprovedMethod}. Branch exchange methods \cite{civanlarFeederReconfiguration, baran1989Recondig, HUDDLESTON1990Reconfiguration, Roytelman1996MultiObjective} provide a faster alternative by performing only a local heuristic topology search starting from the current distribution network topology. These methods are often used in practice because of their robustness and explainability; however, they do not provide a theoretical guarantee of optimality because they search only through the subset of all possible network topologies.

In addition to performing local and heuristic searches, or applying classical optimization techniques or exhaustive searches, one can use stochastic optimization algorithms that perform approximate searches through the whole search space. Therefore, a set of tools often used for solving static DNR are nature-inspired metaheuristic algorithms, which are more computationally efficient than the classical optimization approaches and can provide better solutions compared to the heuristic algorithms. Representative metaheuristic algorithms used to solve the DNR problem are the genetic algorithm \cite{Haghighat2016Distribution, GUPTA2014664, kara1992ImplementationGenetic, Lin2000DistributionFR}, evolutionary algorithms \cite{tsai2010Gray, Delbem2005MainChain}, particle swarm optimization \cite{wu2011Application, Sivanagaraju2008DiscretePSO}, simulated annealing algorithm \cite{jeon2002Annealing, CHANG1994SimulatedAnnealing}, and others. The main drawbacks of these methods are their stochastic nature and suboptimal solutions when the problem dimensionality increases.

Ref. \cite{KARIMIANFARD2019105943} uses a different methodology that performs DNR without solving optimization based or power flow-based programs. It presents a simple and fast strategy specific to the DNR problem for selecting candidate solutions using only primitive network topology information. Several supervised learning approaches to static DNR using artificial neural networks have been presented in \cite{kim1993ArtificialNeural, Salazar2006ArtificialNeural}. In both references, neural networks output the radial distribution network topology given the input set of variables describing the power system state, directly or indirectly. Neural networks are trained on datasets labelled by classical optimization-based DNR solvers and try to mimic them during the evaluation time. Outputs of these methods are deterministic, and the evaluation time is fast, since it is determined by the computational complexity of several matrix-vector multiplications. The main drawbacks of these approaches include the need for DNR solver to the training set, and significantly deteriorated results in the case where inputs significantly differ from the training set samples.

Besides the typical DNR problem formulation, some extensions regarding the optimization function and the constraints are emerging. The study \cite{Ahmadi2018Novel} presented the DNR for reducing power loss with a budget limit as a hard constraint for the planning purposes of distribution networks. The modern distribution network is being transformed from passive to active due to increasingly deployed renewable energy resources and, consequently, the use of distributed generators for the DNR problem is becoming a study of importance \cite{Fathi2020Reconfig, Liu2019DDNR}.

A significantly lower number of studies tackle the dynamic DNR problem formulation. Ref. \cite{Golshannavaz2014Smart} formulates day-ahead scheduling as a mixed-integer nonlinear optimization problem that minimizes total operational costs. The scheduling problem consists of control of distributed generations and responsive loads, as well as per-hour network reconfiguration with switching manipulation constraints, and is solved using the genetic algorithm.
Study \cite{li2008Novel} also solves DDNR with switching manipulation constraints in a multi-agent fashion, by dividing the problem into multiple time intervals, generating multiple instances of a problem solved separately by particle swarm optimization based agents. Both of these studies are numerically tested on small distribution networks and have the problem of computation time increasing significantly with the increase of the problem dimension. DNR can reduce active power losses by being performed hourly, daily, or monthly, as in \cite{Broadwater1993Time, chen1993Energy, Lopez2004Online}; however, these references do not consider the limits of the number of switching operations in the mentioned time periods. DNR studies for annual network reconfiguration that consider variable loads are presented in \cite{SHARIATKHAH20121, ZIDAN2013698}. The study \cite{SHARIATKHAH20121} additionally deals with minimizing the cost of switching operations using dynamic programming combined with the harmony search algorithm. While solving the annual DDNR using the genetic algorithm, the study \cite{ZIDAN2013698} considers the stochastic power generation of distributed generators. The study \cite{MILANI201368} also attempts to solve DDNR using the genetic algorithm, by calculating the optimal intervals between the two topology changes. Ref. \cite{mazza2015Determination} presents the DDNR based on the rule-based algorithm which ranks per-hour DNR solutions and using that finds the optimal time for the network reconfiguration. Multi-objective DDNR using the combination of the heuristic exchange market algorithm and the population-based wild goats algorithm with the possibility of parallel implementation is presented in \cite{JAFARI2020106146}. The proposed method optimizes active power loss and reliability indexes while satisfying radiality, bus voltage, and branch apparent power constraints, where it does not consider switching manipulation constraints. References \cite{Kovacki2018ScalableAF, kovavcki2018operativno} proposed single- and multi-objective formulation of DDNR based on the Lagrange relaxation approach. In the single-objective formulation, the objective function models the active power loss reduction, while in the multi-objective formulation, the objective function minimizes the costs of energy losses, network reliability, and switching operations. The study \cite{Gao2019DDNR} presents a data-driven DDNR for active power loss reduction without using the network parameter information. DDNR is formulated as a Markov decision process (MDP) and is solved using an off-policy reinforcement learning algorithm trained on a historical operation data set.

\textbf{Contributions:} This thesis proposes DDNR based on the DRL algorithm. The proposed expression of the DDNR problem in the RL framework, that is, the definition of the state variables, leads to lower observability requirements compared to the approach proposed in \cite{Gao2019DDNR}. The amount of information needed for the algorithm execution is decreased since the topology information and the information about the power flows in the network are compressed into a single set of variables. This reduces the number of telemetered measurements needed for the possible execution of the algorithm in the real world. The reduced state size is also convenient from the algorithm training perspective, since it decreases the required size of the neural network. We also propose a way of considering switching operation constraints that improves the algorithm training computational efficiency. The proposed approach assumes selecting the actions from the available subset of the action set, which is updated during the episode, so that switching operation constraints are not violated. This approach simplifies the reward function when compared to the approach that allows actions that violate constraints but penalizes them with a large amount of negative reward. This way of selecting actions can be used for optimization problem constraints whose violation can be detected without the feedback from the environment (by evaluating only the agent’s action) and it can be applied to similar power system control and optimization problems treated with RL such as Volt-VAR optimization, energy storage scheduling, supply restoration, etc. The total cost benefits and execution times of the proposed algorithm are compared with the state-of-the-art method from \cite{Kovacki2018ScalableAF}. The main contributions of our work regarding DDNR, published in \cite{kundavcina2022solving}, are:
\begin{itemize}[leftmargin=*]
\item Suggested multi-objective and scalable DRL-based approach is computationally efficient during the algorithm execution, with the expense of high computation cost during the algorithm training.
\item We introduce a novel definition for the state variables of the RL agent, resulting in decreased observability requirements.
\item We proposed a computationally efficient way of considering switching operation constraints by creating the available subset of the action set and updating it during the episode.
\end{itemize}
In Chapter \ref{ch:ddnr} we introduce the main idea of DNR and formulate the DDNR problem. Chapter \ref{ch:rl} presents the theoretical foundations of MDPs and RL. Chapter \ref{ch:rl_ddnr} presents the expression of the DDNR problem in the RL framework, description and discussion of the numerical experiments, and the conclusion along with the possible future work directions.

\part{State Estimation and Graph Neural Networks}

\chapter{Power System State Estimation}	\label{ch:psse}
\addcontentsline{lof}{chapter}{2 Power System State Estimation}

In this chapter, we review two most common formulations of the power transmission system SE problem. The SE algorithm is a key component of the energy management system that provides an accurate and up-to-date representation of the current state of the power system. Its purpose is to estimate complex bus voltages using available measurements, power system parameters, and topology information \cite{monticelli2000SE, phdthesisMirsad}. In this sense, the SE can be seen as a problem of solving large, noisy, sparse, and generally nonlinear systems of equations. The measurement data used by the SE algorithm usually come from two sources: the SCADA system and the WAMS system. The SCADA system provides low-resolution measurements that cannot capture system dynamics in real-time, while the WAMS system provides high-resolution data from PMUs that enable real-time monitoring of the system. The SE problem that considers measurement data from both WAMS and SCADA systems is formulated in a nonlinear way and traditionally solved in a centralized manner using the Gauss-Newton method \cite{monticelli2000SE}. On the other hand, the SE problem that considers only PMU data provided by WAMS has a linear formulation, providing faster, non-iterative solutions. In the following sections, we provide a detailed description of both linear and nonlinear SE problem formations.

\section{Foundational Concepts}	\label{sec:General_Concepts}
This section provides the fundamentals of SE state variables and input data, which are necessary for specific SE problem formulations. For clarity, all variables are expressed in per unit and all transformer ratios are normalized to unity (cancelled out). Additionally, without loss of generality, we make the assumption that the power system does not include phase-shifting transformers.

As mentioned, the outputs of the SE algorithm, i.e., the state variables consist of voltage phasors of all the buses in the power system, where each voltage phasor is represented using a complex number. Let $\mathcal{H} = \{1,\dots,n \}$ represents the set of buses, where $n$ is the number of buses in the power system. The complex bus voltages can be represented both in polar and rectangular coordinate system:
\begin{equation}
   	\begin{aligned}
    \ph V_i = V_{i}\mathrm{e}^{\mathrm{j}\theta_{i}} = 
    \Re {(\ph V_i)} + \mathrm{j} \Im{(\ph V_i)},
   	\end{aligned}
   	\label{state_vol}
\end{equation} 
where $i \in \mathcal{H}$ represents the bus index. $V_{i}$ and $\theta_{i}$ represent magnitude and phase angle, while $\Re {(\ph V_i)}$ and $\Im{(\ph V_i)}$ represent real and imaginary parts of the complex bus voltage $\ph V_i$. Since the state variable vector $\mathbf{x}$ is a vector of real numbers, it can also be represented in polar:
	\begin{equation}
   	\begin{aligned}
    \mathbf x  &=[\bm \uptheta,\mathbf V]^{\mathrm{T}}\\
    \bm \uptheta&=[\theta_1,\dots,\theta_n]\\
    \mathbf V&=[V_1,\dots, V_n],
   	\end{aligned}
   	\label{polar_coord}
	\end{equation} 
as well as in rectangular coordinate system:
	\begin{equation}
   	\begin{aligned}
    \mathbf x &=[\mathbf{V}_\mathrm{re},\mathbf{V}_\mathrm{im}]^{\mathrm{T}}\\
    \mathbf{V}_\mathrm{re}&=\big[\Re(\ph{V}_1),\dots,\Re(\ph{V}_n)\big]\\
	\mathbf{V}_\mathrm{im}&=\big[\Im(\ph{V}_1),\dots,\Im(\ph{V}_n)\big].     
   	\end{aligned}
   	\label{rect_coord}
	\end{equation} 
For simplicity, we omit the concept of the slack bus whose angle value is fixed and acts as a reference value, as it does not impact the way in which the GNN-based SE will be realized in Chapter \ref{ch:gnn_se}.

Traditionally, the input data for the SE algorithm consists of the network topology and parameters, and measured values obtained from the measurement devices spread across the power system. The power system network topology is described by the bus-branch model and can be represented using a graph $\mathcal{G} =(\mathcal{H},\mathcal{E})$, where the set of nodes in the graph is equal to the already defined set of buses in the power system $\mathcal{H}$, while the set of edges $\mathcal{E} \subseteq \mathcal{H} \times \mathcal{H}$ represents the set of branches of the power network. Power system parameters are characteristics of a power system, such as impedance, admittance, etc., that describe the system's behaviour. These parameters are used to build a set of equations that describe the power system via the two-port $\pi$-model of branches in the network. More precisely, the branch $(i,j) \in \mathcal{E}$ between buses $\{i,j\} \in \mathcal{H}$ can be modelled using complex expressions:
\begin{equation}
  \begin{bmatrix}
    \ph{I}_{ij} \\ \ph{I}_{ji}
  \end{bmatrix} =
  \begin{bmatrix}
    y_{ij} + y_{\text{s}i} & -{y}_{ij}\\
    -{y}_{ij} & {y}_{ij} + y_{\text{s}j}
  \end{bmatrix}  
  \begin{bmatrix}
    {\ph{V}}_{i} \\ {\ph{V}}_{j}
  \end{bmatrix},
  \label{unified}
\end{equation} 
where the parameter $y_{ij} = g_{ij} + \text{j}b_{ij}$ represents the branch series admittance, while branch shunt admittances are given as $y_{si} = g_{si} + \text{j}b_{si}$ and $y_{sj} = g_{sj} + \text{j}b_{sj}$. The complex expressions $\ph{I}_{ij}$ and $\ph{I}_{ji}$ define branch currents from the bus $i$ to the bus $j$, and from the bus $j$ to the bus $i$, respectively. The complex bus voltages at buses $\{i,j\}$ are given as $\ph{V}_i$ and $\ph{V}_j$, respectively. 

Input measurements can be placed on various elements in the power system and measure different electrical quantities. Each measurement is associated with the measurement value $z_i$, the measurement variance $v_i$, and the measurement function $f_i(\mathbf{x})$. Measurement functions are mathematical models that express individual measurements in terms of state variables $\mathbf{x}$ using the physical laws in the power system, and can be derived using equations given in \eqref{unified}. A typical set of input measurements includes:
\begin{itemize}
    \item Legacy measurements: The set of legacy measurements provided by SCADA includes active and reactive power flow and injection, branch current magnitude, and bus voltage magnitude measurements. These measurements have low sampling rates and therefore are not suitable for real-time SE. Measurement functions that express these measurements are generally nonlinear, regardless of the coordinate system in which they are represented.
    \item Phasor  measurements: The WAMS supports PMUs and provides phasor measurements of bus voltages and branch currents \cite[Sec. 5.6]{phadke}. More precisely, phasor measurement is formed by a magnitude, equal to the root-mean-square value of the signal, and phase angle \cite[Sec.~5.6]{phadke}. The PMU placed at the bus measures bus voltage phasor and current phasors along all branches incident to the bus \cite{exposito}. Phasor measurements have high sampling rates, with values around 50 samples per second, and also have lower variances compared to legacy measurements. When phasor measurements and state variables are expressed in rectangular coordinate system, the corresponding measurement functions are linear; otherwise they are nonlinear.
\end{itemize}

Values of both legacy and phasor measurements can be stacked together in a vector of measurement values $\mathbf{z} = [z_1,\dots,z_m]^{\mathrm{T}}$. Corresponding measurement functions form their own vector $\mathbf{f}(\mathbf{x})=$ $[f_1(\mathbf{x})$, $\dots$, $f_m(\mathbf{x})]^{\mathrm{T}}$, where $m$ denotes the number of measurement values.

\begin{figure}[htbp]
    \centering
    
    \begin{tikzpicture}[thick, scale=0.7, transform shape]
        \tikzset{
            factorSCADA/.style={draw=black,fill=green!80, minimum size=4mm},
            factorPMU/.style={draw=black,fill=blue!80, minimum size=4mm}}
        \draw (0,0) 
            node[factorPMU, label=below:$\ph V_{m1} \mathrm{=} V_{m1}
            \text{e}^{\text{j}\theta_{m1}}$](L11){} 
            -- ++(0,2)
            -- ++(1,0)
            \bushere{$\ph{V}_1 = V_1 \text{e}^{\text{j}\theta_1}$}{Bus 1} -- ++(1.5,0)
            node[factorPMU, label=below:$\ph I_{12} \mathrm{=} I_{12}  \text{e}^{\text{j}\theta_{I_{12}}}$](L1){}
            -- ++(2,0)
            node[factorSCADA, label=below:$P_{12}$](L2){} 
            to[] ++(3,0)
            \bushere{$\ph{V}_2 = V_2 \text{e}^{\text{j}\theta_2}$}{Bus 2}
            -- ++(1,0) 
            -- ++(0,-2) 
            node[factorSCADA, label=below:$V_{m2}$](L22){} 
            ;
    \end{tikzpicture}

\caption{Simple two-bus power system containing a PMU at the bus 1, one legacy active power flow measurement, and one legacy voltage magnitude measurement at the bus 2.}
    \label{TwoBusPowerSyst}
    
\end{figure}
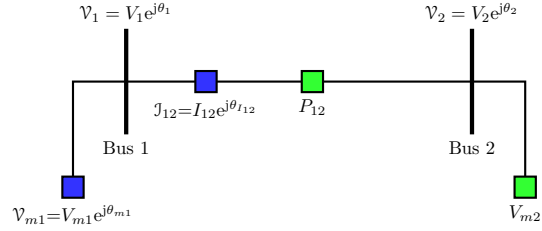

An example of a simple two-bus power system is given in Fig.\ref{TwoBusPowerSyst}. Its state variables consist of two complex bus voltages, $\ph{V}_1$ and $\ph{V}_2$, and the state variable vector is given in polar coordinates as:
\begin{equation}
\mathbf x =[\theta_1, \theta_2, V_1, V_2]^{\mathrm{T}}.
\end{equation}
The system has a PMU placed at the bus 1 which measures the voltage phasor given on that bus $\ph V_{m1} = V_{m1}\text{e}^{\text{j}\theta_{m1}}$ and the current phasor $\ph I_{12} = I_{12}  \text{e}^{\text{j}\theta_{I_{12}}}$ on the branch which connects the two buses. The system also contains a legacy active power flow measurement $P_{12}$ on the same branch and the legacy voltage magnitude measurement $V_{m2}$ on the bus 2. The vector of measurement values of this system can be given as:
\begin{equation}
\mathbf z =[V_{m1}, \theta_{m1}, I_{12}, \theta_{I_{12}}, V_2, P_{12}]^{\mathrm{T}}.
\end{equation}

\section{Linear State Estimation} \label{linearSEbackground}

Since measurement functions corresponding to phasor measurements can be expressed as a linear combination of state variables when represented in rectangular coordinate system, the SE problem formulation which considers only phasor measurements is linear. This formulation is viable in cases when the power system is fully observable using PMUs, and a fast SE solver is then needed to fully utilize their high sampling rates.

PMUs measure complex bus voltages and complex branch currents, and originally output phasor measurements in polar coordinates. In addition, PMU outputs can be observed in the rectangular coordinates with real and imaginary parts of the bus voltage and branch current phasors. In that case, the vector of state variables $\mathbf{x}$ can also be given in rectangular coordinates $\mathbf x \equiv[\mathbf{V}_\mathrm{re},\mathbf{V}_\mathrm{im}]^{\mathrm{T}}$. Using rectangular coordinates, we obtain the linear system of equations defined by voltage and current measurements. The measurement functions corresponding to the bus voltage phasor measurement on the bus $i \in \mathcal{H}$ are simply equal to: 
\begin{equation}
    \begin{aligned}
        f_{\Re\{\ph V_i\}}(\mathbf x) = \Re\{\ph V_i\}\\
        f_{\Im\{\ph V_i\}}(\mathbf x) = \Im\{\ph V_i\}.
    \end{aligned}  
    \label{measFunction1}
\end{equation}
According to the two-port $\pi$ branch model \eqref{unified}, functions corresponding to the branch current phasor measurement are given as:
	\begin{equation}
   	\begin{aligned}
	f_{\Re (\ph{I}_{ij})}(\cdot) & = (g_{ij}+g_{\mathrm{s}i})\Re(\ph{V}_{i}) -
	(b_{ij}+b_{\mathrm{s}i})\Im(\ph{V}_{i}) - g_{ij}\Re(\ph{V}_{j}) +
	b_{ij}\Im(\ph{V}_{j})  
	\\
	f_{\Im (\ph{I}_{ij})}(\cdot) & = (b_{ij}+b_{\mathrm{s}i})\Re(\ph{V}_{i}) +
	(g_{ij}+g_{\mathrm{s}i})\Im(\ph{V}_{i}) - b_{ij}\Re(\ph{V}_{j}) -
	g_{ij}\Im(\ph{V}_{j}).      
	\end{aligned}
   	\label{mf_current_pmu_linear}%
	\end{equation}
The presented model represents the system of linear equations, where the solution can be found by solving the linear weighted least-squares problem: 
\begin{equation}
    \left(\mathbf J^{T} \mathbf \Sigma^{-1} \mathbf J \right) \mathbf x =
		\mathbf J^{T} \mathbf \Sigma^{-1} \mathbf z,    
	\label{SE_system_of_lin_eq}
\end{equation}
where the Jacobian matrix $\mathbf {J} \in \mathbb {R}^{m \times 2n}$ is defined according to measurement functions \eqref{measFunction1}-\eqref{mf_current_pmu_linear}, $m$ is the total number of linear equations, the measurement error covariance matrix is given as $\mathbf {\Sigma} \in \mathbb {R}^{m \times m}$, and the vector $\mathbf z \in \mathbb {R}^{m}$ contains measurement values given in rectangular coordinate system. 

The main disadvantage of this approach is that measurement errors are originally given in polar coordinates (i.e., magnitude and angle errors); therefore, the covariance matrix must be transformed from polar to rectangular coordinates \cite{zhou2016phasorsMeasSE}. As a result, measurement errors are correlated and the covariance matrix $\mathbf {\Sigma}$ does not have a diagonal form. Despite that, because of the lower computational effort, the non-diagonal elements of the covariance matrix $\mathbf {\Sigma}$ are usually neglected, which has an effect on the accuracy of the SE \cite{exposito}. Using the classical theory of propagation of uncertainty \cite{PropagationOfUncertainty}, the variance in the rectangular coordinate system can be obtained using variances in the polar coordinate system. For example, let us observe the voltage phasor measurement at the bus $i$, where PMU outputs the voltage magnitude measurement value $z_{|V_i|}$ with corresponding variance $v_{|V_i|}$, and voltage phase angle measurement $z_{\theta_i}$ with variance $v_{\theta_i}$. Then, variances in the rectangular coordinate system can be obtained as:
\begin{equation}
    \begin{aligned}
        v_{\Re\{V_i\}} &= v_{|V_i|} (\cos z_{\theta_i})^2 + v_{\theta_i} (z_{|V_i|} \sin z_{\theta_i})^2\\
        v_{\Im\{V_i\}} &= v_{|V_i|} (\sin z_{\theta_i})^2 + v_{\theta_i} (z_{|V_i|} \cos z_{\theta_i})^2.
    \end{aligned}
\end{equation}
Analogously, we can easily compute variances related to current measurements $v_{\Re\{\ph{I}_{ij}\}}$, $v_{\Im\{\ph{I}_{ij}\}}$ or $v_{\Re\{\ph{I}_{ji}\}}$, $v_{\Im\{\ph{I}_{ji}\}}$. 
We will refer to the solution of \eqref{SE_system_of_lin_eq} in which measurement error covariances are neglected to avoid the computationally demanding inversion of the non-diagonal matrix $\mathbf {\Sigma}$ as an \textit{approximative WLS SE solution}. 

In this work, we will investigate if the GNN model trained with measurement values, variances, and covariances labelled with the exact solutions of \eqref{SE_system_of_lin_eq} is more accurate than the approximative WLS SE, which neglects the covariances. Inference performed using the trained GNN model scales linearly with the number of power system buses, making it significantly faster than both the approximate and the exact solver of \eqref{SE_system_of_lin_eq}.

\section{Nonlinear State Estimation}	

Today's power systems are often not fully monitored with PMUs, therefore, SE that incorporates both phasor and legacy measurements is required. As previously discussed, that SE formulation is nonlinear and uses state variables expressed in a polar coordinate system $\mathbf x  =[\bm \uptheta,\mathbf V]^{\mathrm{T}}$.

Below, we present expressions for measurement functions corresponding to legacy measurements:
\begin{itemize}
    \item Measurement function for the bus voltage magnitude measurements is simply given as voltage magnitude state variable corresponding to that bus:
 		\begin{equation}
        \begin{aligned}
        f_{V_{i}}(\mathbf x) = V_i.
        \end{aligned}
        \label{mf_voltage_leg}
		\end{equation} 

    \item Measurement functions for active and reactive power flow measurements on branches are given as:
    \begin{equation}
       	\begin{aligned}
        f_{P_{ij}}(\mathbf x)&=
        {V}_{i}^2(g_{ij}+g_{si})-{V}_{i}{V}_{j}(g_{ij}\cos\theta_{ij}
        +b_{ij}\sin\theta_{ij}) \\
        f_{Q_{ij}}(\mathbf x)&=
        -{V}_{i}^2(b_{ij}+b_{si})-{V}_{i}{V}_{j}(g_{ij}\sin\theta_{ij}
        -b_{ij}\cos\theta_{ij}).
       	\end{aligned}
	\end{equation}

    \item Measurement function for current magnitude measurements on branches are:
    \begin{equation}
	f_{I_{ij}}(\mathbf x) = 
    [A_\mathrm{c} V_i^2 + B_\mathrm{c} V_j^2 - 2 V_iV_j
    (C_\mathrm{c} \cos \theta_{ij}-D_\mathrm{c} \sin \theta_{ij})]^{1/2},
    \label{mf_curmag}
	\end{equation}	
where the coefficients of the function are given as: 		
	\begin{equation}
    \begin{aligned}
    A_\mathrm{c}&=(g_{ij}+g_{\mathrm{s}i})^2+(b_{ij}+b_{\mathrm{s}i})^2;&
    B_\mathrm{c}&=g_{ij}^2+b_{ij}^2\\
    C_\mathrm{c}&=g_{ij}(g_{ij}+g_{\mathrm{s}i})+b_{ij}(b_{ij}+b_{\mathrm{s}i});&
    D_\mathrm{c}&=g_{ij}b_{\mathrm{s}i}-b_{ij}g_{\mathrm{s}i}.
    \end{aligned}
    \nonumber
	\end{equation}

    \item Measurement functions for active and reactive power injection measurements are described as:
    \begin{equation}
   	\begin{aligned}
    f_{P_{i}}(\mathbf x) &={V}_{i}\sum\limits_{j \in {\mathcal{N}_i} \mathop{\cup} i} {V}_{j}
    (G_{ij}\cos\theta_{ij}+B_{ij}\sin\theta_{ij})\\
    f_{Q_{i}}(\mathbf x) &={V}_{i}\sum\limits_{j \in {\mathcal{N}_i} \mathop{\cup} i} {V}_{j}
    (G_{ij}\sin\theta_{ij}-B_{ij}\cos\theta_{ij}),
   	\end{aligned}
	\end{equation}
where $\mathcal{N}_i \in \mathcal{H}$ is the set containing first order neighbours of the bus $i$. $G_{ij}$ and $B_{ij}$ are the elements of bus admittance matrix often used in power system analysis \cite{kundur_power_nodate}, and can be calculated using:
	\begin{equation}
   	\begin{aligned}
  	Y_{ij}= G_{ij} + \mathrm{j}B_{ij} = 
  	\begin{cases}
   	\sum\limits_{j\in \mathcal{N}_i} 
   	({y}_{ij}+{y}_{\mathrm{s}i}), & \text{if} \;\; i=j 
   	 \;\;(\mathrm{diagonal\;element})\\
   	-{y}_{ij}, & \text{if} \;\; i \not = j
   	 \;\;(\mathrm{non-diagonal\;element).}
	\end{cases}
	\end{aligned}
   	\label{adm_mat_ele}
	\end{equation}
\end{itemize}

Next, we provide expressions for measurement functions corresponding to phasor measurements expressed in polar coordinate system:
\begin{itemize}
    \item Measurement functions for bus voltage phasors measurements are given as:
     \begin{equation}
    \begin{aligned}
    	f_{{V}_{i}}(\mathbf x) = V_i\\
    	f_{\theta_i}(\mathbf x) = \theta_i.
    \end{aligned}    
    \label{vol_pha_meas_fun}
    \end{equation} 

    \item The measurement function for the magnitude of the branch current phasor is given in \eqref{mf_curmag}, while the function for the measured angle of the branch current phasor is:
    \begin{equation}
	f_{\theta_{I_{12}}}(\mathbf x) =\mathrm{arctan}\Bigg[ 
	\cfrac{(A_\mathrm{a} \sin\theta_i
    +B_\mathrm{a} \cos\theta_i)V_i 
    - (C_\mathrm{a} \sin\theta_j + D_\mathrm{a}\cos\theta_j)V_j}
   	{(A_\mathrm{a} \cos\theta_i
    -B_\mathrm{a} \sin\theta_i)V_i 
    - (C_\mathrm{a} \cos\theta_j - D_\mathrm{a} \sin\theta_j)V_j} \Bigg],    
    \label{mf_curang}
	\end{equation}
where the function's coefficients are as follows: 	
	\begin{equation}
    \begin{aligned}
    A_\mathrm{a}&=g_{ij}+g_{\mathrm{s}i};&
    B_\mathrm{a}&=b_{ij}+b_{\mathrm{s}i}\\
    C_\mathrm{a}&=g_{ij};&
    D_\mathrm{a}&=b_{ij}.
    \end{aligned}
    \nonumber
	\end{equation}	
\end{itemize}

Finally, the SE model can be expressed as the following system of nonlinear equations:
\begin{equation}
    \mathbf{z} = \mathbf{f}(\mathbf{x}) + \mathbf{u},
    \label{nonlinear_model}
\end{equation}
where $\mathbf{u} \in \mathbb{R}^{m}$ is a vector of uncorrelated measurement errors, where $u_i \sim \mathcal{N}(0, v_{i})$ represents a zero-mean Gaussian distribution with variance $v_{i}$. The GN method is typically used to solve the nonlinear SE model \eqref{nonlinear_model}, where the measurement functions $\mathbf{f}(\mathbf{x})$ precisely follow the physical laws derived on the basis of \eqref{unified}:
		\begin{subequations}
        \begin{gather}  
		\Big[\mathbf J (\mathbf x^{(\nu)})^\mathrm{T} \mathbf \Sigma \, 
		\mathbf J (\mathbf x^{(\nu)})\Big] \Delta \mathbf x^{(\nu)} =
		\mathbf J (\mathbf x^{(\nu)})^\mathrm{T}
		\mathbf \Sigma \, \mathbf r (\mathbf x^{(\nu)})\label{AC_GN_increment}\\
		\mathbf x^{(\nu+1)} = 
		\mathbf x^{(\nu)} + \Delta \mathbf x^{(\nu)}, \label{AC_GN_update}
        \end{gather}
        \label{AC_GN}%
		\end{subequations}
where $\nu = \{0,1,\dots,\nu_{\max}\}$ is the iteration index and $\nu_{\max}$ is the number of iterations, $\Delta \mathbf x^{(\nu)} \in \mathbb {R}^{2n}$ is the vector of increments of the state variables, $\mathbf J (\mathbf x^{(\nu)})\in \mathbb {R}^{m\mathrm{x}2n}$ is the Jacobian matrix of measurement functions $\mathbf f (\mathbf x^{(\nu)})$ at $\mathbf x=\mathbf x^{(\nu)}$, $\mathbf{{\Sigma}}\in \mathbb {R}^{m\mathrm{x}m}$ is in this case a diagonal matrix containing inverses of measurement variances, and $\mathbf r (\mathbf x^{(\nu)}) =$ $\mathbf{z}$ $-\mathbf f (\mathbf x^{(\nu)})$ is the vector of residuals. Note that the nonlinear SE represents a nonconvex problem arising from nonlinear measurement functions $\mathbf{f}(\mathbf{x})$ \cite{ilic}. Due to the fact that the values of the state variables $\mathbf{x}$ usually fluctuate in narrow boundaries, the GN method can be used.

The SE model \eqref{nonlinear_model} that considers both legacy and phasor measurements, where the vector of state variables $\mathbf{x} = [\mathbf{V}, \bm \uptheta]^T$ and phasor measurements are represented in the polar coordinate system, is known as simultaneous. The simultaneous SE model takes measurements provided by PMUs in the same manner as legacy measurements. More precisely, the PMU generates measurements in the polar coordinate system, which delivers more accurate state estimates than the other representations \cite{exposito}, but requires more computing time \cite{manousakis} and produces ill-conditioned problems \cite{exposito}. To address these issues, we propose a non-matrix-based and noniterative GNN base SE, which can be used as a standalone approach to solve \eqref{nonlinear_model}, or as a fast and accurate initializer of the GN method \eqref{AC_GN}.

\chapter{Graph Neural Networks}	\label{ch:gnn}
\addcontentsline{lof}{chapter}{3 Graph Neural Networks}
Graph neural networks are an increasingly popular deep learning method used for efficient learning over graph-structured data. Various real-world objects and phenomena can be represented as graphs; therefore, GNNs found application in a wide variety of domains, such as chemistry for molecular property prediction \cite{pmlr-v70-gilmer17a}, antibiotic discovery \cite{STOKES2020Antibiotic}, social sciences for fake news detection \cite{Monti2019FakeND}, complex physics simulations \cite{Sanchez2020Learning}, wireless communications \cite{He2021AnOO}, analysis and optimization of electrical power systems \cite{Liao2022ARO}, etc. In this chapter, we provide a short overview of machine learning on graphs, the foundation of the GNN theory used in the rest of the thesis, and list some practical aspects in using GNNs in real-world applications.

\section{Overview of Machine Learning on Graphs}

The main goal of this section is to provide the context necessary to understand GNNs. Firstly, we introduce the definition of a graph and categorize the most common machine learning on graphs tasks. We provide a short reference to the traditional machine learning on graphs methods to emphasize the necessity for graph representation learning. Finally, we provide an overview of graph representation learning methods, including deep learning-based GNNs.

\subsection{Graphs} \label{subsec:Graphs}

\begin{figure}[htbp]
    \centering
    \begin{tikzpicture} [scale=1.0, transform shape]
        \tikzset{
            varNode/.style={circle,minimum size=3mm,fill=white,draw=black},
            edge/.style={very thick,black}}
        \begin{scope}[local bounding box=graph]
            \node[varNode] (f1) at (-3, 1.5 * 2) {1};
            \node[varNode] (v1) at (-1, 1 * 2) {2};
            \node[varNode] (v3) at (1, 1 * 2) {3};
            \node[varNode] (f4) at (3, 1.5 * 2) {4};
            \node[varNode] (f2) at (-1, 2 * 2) {5};
            \node[varNode] (f5) at (1, 2 * 2) {6};
            
            \draw[edge] (f1) -- (f2);
            \draw[edge] (f2) -- (f5);
            \draw[edge] (f1) -- (v1);
            \draw[edge] (f4) -- (v3);
            \draw[edge] (f2) -- (v1);
            \draw[edge] (f5) -- (v3);
            \draw[edge] (v1) -- (v3);

        \end{scope}

    \end{tikzpicture}
    \caption{Example of the simple undirected graph containing six nodes and seven edges.}
    \label{simpleGraph}
    
\end{figure}
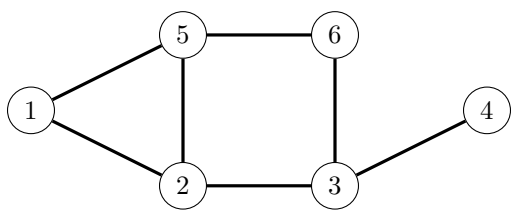

Graphs are often used to describe a set of entities and the relationships between them. Formally, a graph is defined as a tuple $(\mathcal{V}, \mathcal{E})$, where $\mathcal{V}$ denotes the set of nodes, and $\mathcal{E}$ denotes the set of edges between the nodes. The graph is commonly represented with the corresponding adjacency matrix $\mathbf{A} \in \mathbb{R}^{|\mathcal{V}| \times |\mathcal{V}|}$. If there is an edge between the nodes $a,b \in \mathcal{V}$, then a matrix element $\mathbf{A}[a,b]$ is equal to one; otherwise, it is equal to zero. Graphs can contain self-loops, i.e., edges that connect nodes to themselves, resulting in diagonal elements of the adjacency matrix equal to one. An undirected graph assumes bidirectional connections between all the nodes, resulting in a symmetric adjacency matrix. An example of a simple undirected graph is given in ~Fig.\ref{simpleGraph}, with the corresponding adjacency matrix given in \eqref{eq:ajdMatrix}:
\begin{equation} \label{eq:ajdMatrix}
\mathbf{A} = 
\begin{bmatrix}
0 & 1 & 0 & 0 & 1 & 0\\
1 & 0 & 1 & 0 & 1 & 0\\
0 & 1 & 0 & 1 & 0 & 1\\
0 & 0 & 1 & 0 & 0 & 0\\
1 & 1 & 0 & 0 & 0 & 1\\
0 & 0 & 1 & 0 & 1 & 0
\end{bmatrix}.
\end{equation}

There are many extensions of the simplest form of graph presented above. Some of them are the following:
\begin{itemize}
    \item Directed graph, in which the adjacency matrix is generally not symmetric since a connection from one node to another does not imply the existence of a connection in the reverse direction;
    \item Heterogeneous and multi-relational graphs, where there can be multiple types of the nodes and the edges;
    \item Weighted graphs, which have weights associated with the edges, and consequently their adjacency matrices contain real number scalars instead of only zeros and ones.
\end{itemize}

Real-world graphs can contain a large number of nodes, reaching hundreds of millions in the cases of the most popular social networks. The adjacency matrices of these graphs are sparse and space-inefficient, necessitating a more compact graph storage. Therefore, in practise large and sparse graphs are usually stored in adjacency lists, which are implemented as unordered lists of different sizes containing neighbours of each node.

Additional input data can be incorporated into the graph data structure, usually at the node level via real-valued feature vectors, introducing the need for machine learning methods on graphs.  Less often, the input data are provided using the edge-level features, or at the graph level using a single feature vector. In the case of supervised learning on nodes, edges, and graphs, training labels are usually concatenated into the corresponding feature vectors. From the graph signal processing perspective, node-level features can be viewed as signals on a graph. A vector containing one scalar feature per node defines a one-channel graph signal $\mathbf x \in \mathbb {R}^{|\mathcal{V}|}$\footnote{In Chapters \ref{ch:psse} and \ref{ch:gnn_se}, $\mathbf x$ was used as notation for state variables. In this chapter, it is used to denote signals on a graph. No overlap in meaning occurs between chapters, avoiding any potential confusion.}, while multichannel graph signals can be represented with a matrix $\mathbf X \in \mathbb {R}^{|\mathcal{V}| \cdot N_c}$, where $N_c$ denotes the number of input features per node.

Another matrix that represents fundamental properties of a graph and will be referred to throughout this chapter is the graph Laplacian matrix $\mathbf{L} \in \mathbb{R}^{|\mathcal{V}| \times |\mathcal{V}|}$, defined as:
\begin{equation} \label{eq:laplacianMatrixDef}
\mathbf{L} = \mathbf{D} - \mathbf{A},
\end{equation}
where $\mathbf{D} \in \mathbb{R}^{|\mathcal{V}| \times |\mathcal{V}|}$ is diagonal and represents the node degree matrix, whose diagonal elements are equal to corresponding node degrees. This matrix is positive semi-definite, and it can always be eigendecomposed. The Laplacian matrix of the graph is given in Fig.~\ref{simpleGraph} is:
\begin{equation} \label{eq:laplacianMatrix}
\mathbf{L} = 
\begin{bmatrix}
2 & -1 & 0 & 0 & -1 & 0\\
-1 & 3 & -1 & 0 & -1 & 0\\
0 & -1 & 3 & -1 & 0 & -1\\
0 & 0 & -1 & 1 & 0 & 0\\
-1 & -1 & 0 & 0 & 3 & -1\\
0 & 0 & -1 & 0 & -1 & 2
\end{bmatrix}.
\end{equation}

The normalized version of the graph Laplacian is also often used:
\begin{equation} \label{eq:normLaplacianMatrix}
\mathbf{L} = \mathbf{I} - \mathbf{D}^{-\frac{1}{2}} \mathbf{A} \mathbf{D}^{-\frac{1}{2}}.
\end{equation}

The eigendecomposition of the Laplacian matrix, also known as the spectrum of the graph Laplacian, is given as follows:
\begin{equation} \label{eq:eigendecomp}
\mathbf{L} = \mathbf{U} \mathbf{\Lambda} \mathbf{U}^{\mathrm T},
\end{equation}
where $\mathbf{\Lambda} \in \mathbb{R}^{|\mathcal{V}| \times |\mathcal{V}|}$ is a diagonal matrix containing the eigenvalues, while the columns of $\mathbf{U} \in \mathbb{R}^{|\mathcal{V}| \times |\mathcal{V}|}$ contain the eigenvectors ordered by their corresponding eigenvalues. The Laplacian spectrum reveals some information about the grouping of the graph's nodes. The multiplicity of null eigenvalues is equal to the number of connected components in a graph, while the second-smallest eigenvalue and the corresponding eigenvector can be used to perform the optimal node clustering into two.

\subsection{Common Tasks of Machine Learning on Graphs} \label{subsec:commonTasks}

Generally, learning problems on graphs can be reduced to supervised and unsupervised problems; however, due to graph-related specificities there is a need for a more detailed categorization. Before introducing graph representation learning concepts, we provide descriptions of most common tasks of machine learning on graphs: 

\begin{itemize}
    \item Node-level tasks, in which classification or regression is performed on individual nodes, based on a dataset containing nodes labelled with target values. Typical supervised learning approaches perform poorly in this task, as the nodes in a graph are not independent and identically distributed.
    \item Edge-level tasks, with the most common being prediction of edge presence in a graph, also known as link prediction. Models for these tasks are trained on graphs with an incomplete set of edges to predict the missing edges between the pairs of nodes. By minimizing the loss similar to logistic regression, this task is usually reduced to the classification problem, given data from the node pairs as inputs. Less common tasks are classification and regression of individual edges.
    \item Node clustering, often also called community detection, is a form of unsupervised learning with the goal of grouping similar nodes according to their features and connectivity information.
    \item Graph-level tasks can also be formulated as supervised and unsupervised learning problems, once a useful set of graph features is extracted. Supervised learning on graphs assumes predicting the class or a real-number value associated with a graph, whereas unsupervised learning on graphs usually involves tasks that calculate a measure of similarity between pairs of graphs, such as graph clustering.
    \item Influence maximization, often used for viral marketing purposes, is defined as the problem of finding the subset of nodes in a graph such as a social network that maximizes the spread of influence. The main goal of influence maximization is finding small subsets that provide a high number of affected nodes in the rest of the graph.
\end{itemize}

\subsection{The Need for Graph Representation Learning}
In the standard applications, trivial usage of common machine learning models such as neural networks expect input structured as multidimensional arrays, making the usage of adjacency lists more difficult. An additional problem with using common machine learning models for graph learning problems is that they expect all the node and connectivity data as an input, yielding machine learning models with a high number of parameters, which are inefficient from the storage perspective and hard to train as well. Furthermore, common neural networks are not permutation invariant; the same graph topology can be represented with multiple different adjacency matrices or lists, but it can not be ensured that all of them can be mapped to the same output \cite{sanchez-lengeling2021Distill}. Therefore, machine learning algorithms specialized for operating on graphs that preserve permutation invariance become necessary.

Traditional approaches to machine learning on graphs methods are out of the scope of this thesis. Based on the overview given in \cite{GraphRepresentationLearningBook}, we briefly refer to some of these methods to motivate the need to develop graph representation learning algorithms:
\begin{itemize}
    \item Node-level tasks can be solved by extracting multiple node-level features using common node statistics such as node degree, node centrality, clustering coefficient, etc., and feeding them to the inputs of the common machine learning algorithms.
    \item Solving graph-level tasks traditionally involves extracting graph-level features that can be later used in common machine learning models. One of the trivial graph-level feature extractions is defined as a simple aggregation of node-level features, which can miss some important global information about the graph because it is solely based on local node statistics. More advanced graph kernel methods \cite{Kriege2020ASO}, perform iterative neighbourhood aggregation of node-level features, to capture global information about the graph. Additional graph-level information can be provided by counting the number of small subgraph structures or by analysing various types of paths in the graph. Information about paths is created by collecting the node statistics along the shortest paths or random walks on the graph \cite{kashima2003Marginalized}.
    \item Neighbourhood overlap detection methods quantify how much two nodes are related by analysing similarities between their corresponding neighbourhoods. These statistics can be used for edge-level relationship prediction tasks \cite{LU2011LinkPRediction}. In addition to trivial $k \in \mathbb{N}$-hop neighbourhoods, more advanced random walk and shortest path-based neighbourhood functions can be used.
    \item  Node clustering tasks can be traditionally solved using graph Laplacian matrix and the spectral methods. Many node clustering methods rely on determining and analysing its eigenvalues and eigenvectors, that is, its spectrum \cite{Luxburg2007SpectralClustering}. As mentioned in \ref{subsec:Graphs}, the spectrum of the graph Laplacian can be used to perform some variants of node clustering. Additionally, the Laplacian spectrum can be used to create vector representations of nodes in a graph, which can be used as input to a typical clustering algorithm.
\end{itemize}
The main drawback of these methods is the need for manual feature engineering, which can be an expensive and time-consuming process. Additionally, designed feature extractors are inflexible and cannot generalize well on new graphs with different topologies. In the Subsection \ref{subsec:GraphRepreLearning} we will consider the most common approaches to learning node vector representations, instead of extracting them manually.

\subsection{Graph Representation Learning} \label{subsec:GraphRepreLearning}

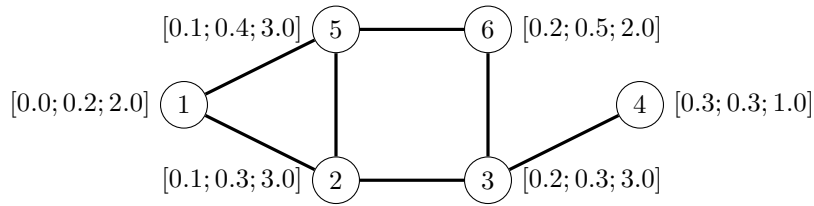
\begin{figure}[htbp]
    \centering
    \begin{tikzpicture} [scale=1.0, transform shape]
        \tikzset{
            varNode/.style={circle,minimum size=3mm,fill=white,draw=black},
            edge/.style={very thick,black}}
        \begin{scope}[local bounding box=graph]
            \node[varNode, label=left:$\mathrm{[0.0; 0.2; 2.0]}$] (f1) at (-3, 1.5 * 2) {1};
            \node[varNode, label=left:$\mathrm{[0.1; 0.3; 3.0]}$] (v1) at (-1, 1 * 2) {2};
            \node[varNode, label=right:$\mathrm{[0.2; 0.3; 3.0]}$] (v3) at (1, 1 * 2) {3};
            \node[varNode, label=right:$\mathrm{[0.3; 0.3; 1.0]}$] (f4) at (3, 1.5 * 2) {4};
            \node[varNode, label=left:$\mathrm{[0.1;0.4; 3.0]}$] (f2) at (-1, 2 * 2) {5};
            \node[varNode, label=right:$\mathrm{[0.2; 0.5; 2.0]}$] (f5) at (1, 2 * 2) {6};
            
            \draw[edge] (f1) -- (f2);
            \draw[edge] (f2) -- (f5);
            \draw[edge] (f1) -- (v1);
            \draw[edge] (f4) -- (v3);
            \draw[edge] (f2) -- (v1);
            \draw[edge] (f5) -- (v3);
            \draw[edge] (v1) -- (v3);

        \end{scope}

    \end{tikzpicture}
    \caption{Node embeddings - a simplified example of graph representation learning algorithm's outputs.}
    \label{simpleGraphWithEmbeddings}
    
\end{figure}

The most common objective of a graph representation learning is to create the representation vectors called node embeddings, that encode the information about graph's local structure\footnote{In this thesis we do not consider less common graph representation learning methods that create edge and graph-level embeddings.}. This process can also be interpreted as a transformation of the graph data into the latent feature space, also known as the embedding space. The distances between the points in the embedding space reflect the node similarity with respect to the relative positions of the nodes. Node embeddings can be a direct output of a graph representation learning algorithm or an intermediate result, as in the case of end-to-end GNNs. In either case, node embeddings, their pairs, or aggregated node embeddings of the whole graph can be used as an input to node, edge, or graph-level tasks mentioned in Subsection \ref{subsec:commonTasks}. Fig.~\ref{simpleGraphWithEmbeddings} displays the simplified output of a typical graph representation learning algorithm, where each node is assigned a corresponding node embedding of size three.

Prior to introducing deep learning-based graph representation learning methods, we will briefly list the types of shallow embedding methods which fundamentally create unique node embeddings for all the nodes, based on their identifiers and their neighbourhood structure. All of these methods follow the idea of the graph encoder-decoder framework, containing an encoder function which maps each node into the real-number vector representation, and a decoder function which maps learned node embedding vectors to the structural information of that node. We will consider the most often used pairwise decoders, which take two node embeddings as an input, and predict some measure of similarity between the two nodes. Encoder-decoder models defined in this way are trained on pairs of similar nodes by minimizing the discrepancy between the similarity of the nodes (e.g. the neighbourhood overlap) and the similarity of the node embeddings obtained as outputs of the decoder. If the graph structure information can be successfully decoded, the learned node embeddings represent the graph well, and can be used as inputs to some of the common machine learning algorithms. The most common types of shallow embedding methods that fit into the encoder-decoder framework are:
\begin{itemize}
    \item Factorization-based methods that express the encoder-decoder loss in matrix form and use matrix-factorization algorithms to minimize it. In these methods, the decoder is usually defined as an L2-norm of the difference between the two node embeddings \cite{Belkin2002LaplacianEigenmaps}, or their inner product \cite{Ahmed2013Large}. An example of the measure of similarity between two nodes in a matrix form is the Leicht-Holme-Newman similarity \cite{Leicht2006VertexSI}, which provides the expected number of paths of all lengths between two nodes by solving the geometric series of the adjacency matrix.
    \item Random walk embedding methods use stochastic node similarity measures based on random walk statistics, since similar pairs of nodes should occur together in short random walks \cite{Perozzi2014DeepWalt, Grover2016Node2Vec}.
\end{itemize}

For example, the DeepWalk shallow embedding algorithm \cite{Perozzi2014DeepWalt} embeds the nodes based on the random walk sequences, similarly like the word2vec algorithm \cite{word2vec} embeds words based on the set of sentences. After a dataset of short fixed-length random walks is generated, the model is optimized to closely embed the nodes that co-occur in the same random walks:
\begin{equation}
    \max\limits_\mathbf H \sum\limits_{u \in \mathcal{V}} \sum\limits_{v \in N_{RW}(u)} log \Pr(v|u).
    \label{DeepWalk}
\end{equation}
$N_{RW}(u)$ denotes the multiset of nodes visited on random walks starting from the node $u$, while $\mathcal{V}$ denotes the set of all nodes. $\mathbf H \in \mathbb {R}^{d \cdot |\mathcal{V}|}$ is the trainable node embedding matrix, containing the node embeddings for individual nodes $\mathbf h_v  \in \mathbb {R}^{d}, v \in \mathcal{V}$, where $d$ denotes the embedding size. Probability that the node $v$ is in the neighbourhood of the node $u$ is parametrized using the softmax function and the node embedding inner products in the following way:
\begin{equation}
    \Pr(v|u) = \frac{\exp{{\mathbf h_u}^{\mathrm{T}}} \mathbf h_v}{ \sum_{n \in \mathcal{V}} \exp{{\mathbf h_u}^{\mathrm{T}}} \mathbf h_n}.
    \label{DeepWalk2}
\end{equation}
In other words, by going through the pairs of nodes that co-occurred in random walks, the algorithm maximizes the inner product of the node embedding pairs, maximizing the probability that the trained algorithm will categorize those node pairs as neighbours. The softmax function also enforces the minimization of inner products of embeddings of nodes that did not co-occur in random walks, making them further apart in the embedding space.

The main drawback of the listed shallow embedding approaches is that they are trained to create unique vector representations of all the nodes of one graph, meaning that they do not have the ability to generalize to new graphs. A more flexible approach would be to learn a function that encodes the local neighbour structure and can be trained and used on graphs with different topologies simultaneously. Another drawback of shallow embedding methods is that they can encode only the graph structure, without taking into account node, edge, and graph-level input features. In the following subsection, we will introduce deep learning-based encoders, which map local node neighbourhoods and all the input feature vectors they contain to node embeddings.

\subsection{Graph Representation Learning using GNNs}

GNNs are becoming the most popular tool for machine learning on graphs problems because of their successful application in various domains. Throughout this chapter we have been motivating the development of specialized machine learning methods for graphs and methods for learning vector representations of nodes in the graphs, until we have motivated the need for training deep learning functions to encode the graph data. Unlike shallow, node embeddings are not the final output of the GNN algorithm, but an intermediate result in end-to-end machine learning on a graph task, either supervised or unsupervised.

We make an introduction to GNNs from the deep learning model's architecture perspective by comparing them to the other common deep learning approaches in Table~\ref{tblInductiveBias}. Adjusting the model's architecture to the specific structure of the input data can increase the training speed and performance and reduce the amount of needed training data. This way of exploiting the regularity of the input data space by imposing the structure of the trainable function space is known by the term relational inductive bias \cite{Battaglia2018Relational}. One of the most successful examples of exploiting relational inductive biases are CNN layers, producing algorithms that surpass human experts in many computer vision tasks. CNNs use the same set of trainable parameters (known as the convolutional kernel) to operate over parts of the input grid data independently, achieving locality and spatial translation invariance. Locality exploits the fact that neighbouring grid elements are more related than further ones, while spatial translation invariance is the ability to map various translations if the input data into the same output. Similarly, recurrent units utilize trainable parameter sharing to process the segments of the sequential data, resulting in a time translation invariant algorithm. From an inductive bias perspective, the main goal of GNNs is to achieve permutation invariance, so that various adjacency matrix representations of the same graph map into the same output. An additional goal of GNNs is to achieve permutation equivariance for node and edge-level tasks, so that node and edge permutations in the input data should manifest only in the corresponding outputs.

\begin{table}[]
\caption{Comparison of various deep learning models from the inductive bias perspective.}
\label{tblInductiveBias}
\begin{tabular}{|c|c|c|c|}
\hline
\begin{tabular}[c]{@{}c@{}}\textbf{Neural network}\\ \textbf{layer type}\end{tabular} & \begin{tabular}[c]{@{}c@{}}\textbf{Input data} \\ \textbf{structure}\end{tabular} & \begin{tabular}[c]{@{}c@{}}\textbf{Relational} \\ \textbf{inductive bias}\end{tabular}            & \textbf{Property}                                                                           \\ \hline
Fully connected                                                     & Arbitrary                                                       & \begin{tabular}[c]{@{}c@{}}Input elements \\ weakly related\end{tabular} & -                                                                                  \\ \hline
Convolutional                                                       & Grids, images                                                   & Local relation                                                                  & \begin{tabular}[c]{@{}c@{}}Spatial translation\\ invariance\end{tabular}           \\ \hline
Recurrent                                                           & Sequences                                                     & Sequential relation                                                             & \begin{tabular}[c]{@{}c@{}}Time translation \\ invariance\end{tabular}             \\ \hline
GNN layer                                                           & Graphs                                                          & Arbitrary relation                                                              & \begin{tabular}[c]{@{}c@{}}Permutation invariance\\  and equivariance\end{tabular} \\ \hline
\end{tabular}
\end{table}

The main classification of GNN methods is the following:
    \begin{itemize}
        \item Spectral GNNs are based on trainable graph convolutions in the spectral domain, achieved using the graph Fourier transform, and involve the eigendecomposition of the graph Laplacian \cite{Bruna2014SpectralNA}. In spectral domain, the graph convolution reduces to element-wise multiplication of the trainable convolution filter with the graph signal. These methods have some important theoretical implications, but also a few drawbacks that make them less applicable in practise; therefore they will not be studied in detail in this thesis. One of the drawbacks is the high computational cost for large graphs, since the eigendecomposition has a $\mathcal{O}(|\mathcal{V}|^3)$ computational complexity, and the fact that the number of trainable parameters grows with the input graph size. Additionally, spectral filters on which they rely cannot localize in the original domain of the graph, and they cannot generalize to new graphs whose eigendecompositions are different from the graphs the model was trained on. Finally, these methods are limited to undirected graphs which have symmetric Laplacian matrices, and cannot include edge-level input features.
        \item Spatial GNNs are a widely used class of GNN methods based on trainable neighbourhood aggregation of node input features, performed in the original (spatial) domain of the graph. The neighbourhood aggregation process is applied locally and independently over the parts of the input graph, making the spatial GNNs easily generalizable to new graphs. Since the spatial GNNs act as local graph filters, the number of trainable parameters does not grow with the input graph size and their inference can be distributed, making them convenient for large scale applications.
    \end{itemize}
    
Before presenting spatial GNNs in detail, we will give a theoretical overview of graph convolution operation in spectral domain and derive a spectral GNN layer using it. The Laplacian matrix eigendecomposition $\mathbf{L} = \mathbf{U} \mathbf{\Lambda} \mathbf{U}^{\mathrm T}$ defines a graph Fourier transform $\mathcal{F}(\cdot)$ providing a way to project a graph signal $\mathbf x \in \mathbb {R}^{|\mathcal{V}|}$ into the spectral domain:
\begin{equation}
\mathcal{F}(\mathbf x) = \mathbf{U}^{\mathrm T} \mathbf x.
\end{equation}
The graph convolutional filter $\mathbf g_\theta \in \mathbb {R}^{|\mathcal{V}|}$ has the same size as the graph signal, and when employed in spectral GNNs, its elements are trainable, i.e. learned from data. In spectral domain, the graph convolution reduces to element-wise multiplication of the graph signal with the filter $\mathcal{F}(\mathbf x) \odot \mathcal{F}(\mathbf g_\theta)$. By performing the inverse Fourier transform $\mathcal{F}^{-1}(\cdot)$ to the signal filtered in the spectral domain, we obtain the result of the graph convolution $\ast_G$ in the original domain:
\begin{equation}
\mathbf x \ast_G \mathbf g_\theta = \mathcal{F}^{-1}(\mathcal{F}(\mathbf x) \odot \mathcal{F}(\mathbf g_\theta)) = \mathbf{U}(\mathbf{U}^{\mathrm T}\mathbf x \odot \mathbf{U}^{\mathrm T}\mathbf g_\theta).
\end{equation}
This expression can further be simplified by expressing the filter in the spectral domain $\mathbf{U}^{\mathrm T}\mathbf g_\theta$ as a diagonal matrix $\mathbf{\Theta} = diag(\mathbf{U}^{\mathrm T}\mathbf g_\theta)$:
\begin{equation}
\mathbf x \ast_G \mathbf g_\theta = \mathbf{U} \mathbf{\Theta} \mathbf{U}^{\mathrm T} \mathbf x.
\end{equation}

A spectral GNN layer transforms a multichannel input into the multichannel output by performing multiple graph convolutions and summing them per output channel. This process is repeated $K$ times, starting with the input multichannel graph signal $\mathbf X \in \mathbb {R}^{|\mathcal{V}| \cdot N_c}$, and ending with the final node embedding matrix $\mathbf X \in \mathbb {R}^{|\mathcal{V}| \cdot f_K}$, where $f_K$ denotes the size of the final node embeddings. Generally, each spectral GNN layer can have a different number of input and output channels, and a separate set of trainable parameters. Operations performed in $k^\mathrm{th}$ GNN layer can be described as:
\begin{equation}
\mathbf H_{:, j}^{k} = \sigma ( \sum\limits_{i=1}^{f_{k-1}} \mathbf{U} \mathbf{\Theta}_{i,j}^{k} \mathbf{U}^{\mathrm T} \mathbf H_{:, i}^{k-1}),  \; \; \; \;  j = 1, 2, \dots f_k.
\end{equation}
$\sigma$ represents some nonlinear function, while $f_{k-1}$ and $f_k$ denote the number of GNN layer's input and output channels. $\mathbf H^{k-1} \in \mathbb{R}^{|\mathcal{V}| \cdot f_{k-1}}$ is the input graph signal for $k^\mathrm{th}$ layer, with $\mathbf H^{0} = \mathbf X$. Finally, $\mathbf{\Theta}_{i,j}^{k}$ contains trainable filter's parameters for every input-output channel combination. The final node embeddings are used as inputs of the additional trainable functions which perform node, edge, or graph-level tasks, and the whole model is trained in an end-to-end fashion.

To make spectral GNNs applicable to large graphs, approximations using Chebyshev polynomials of the diagonal matrix of eigenvalues \cite{chebNetPaper} are often employed. Some of these approximations exhibit spatial GNN properties like localization, which blurs the border between spatial and spectral GNNs. The best example are graph convolutional networks \cite{KipfW17_GCN}, whose neighbourhood aggregation process is theoretically equivalent to the graph convolutional filtering in the original spectral version of the algorithm.

In the next section, we will give the theoretical foundations of spatial GNN methods, which will be applied to the power system state estimation problem in Chapter \ref{ch:gnn_se}.

\section{Theoretical Foundations of Spatial Graph Neural Network}

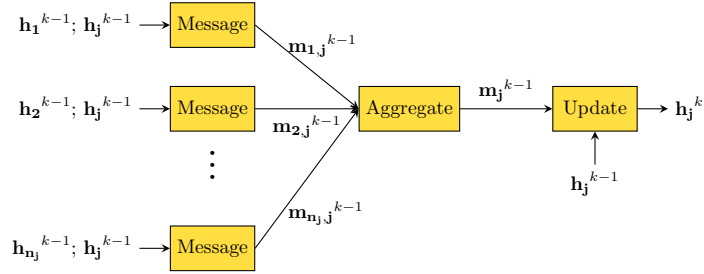
\begin{figure}[!t]
\centering

\begin{tikzpicture} [scale=0.74, transform shape]

\tikzset{
    box/.style={draw, fill=Goldenrod, minimum width=1.5cm, minimum height=0.8cm}}
    
\begin{scope}[local bounding box=graph]

\node [box]  (message1) at (-0.5, 1.5) {Message};
\node [box]  (message2) at (-0.5, 0) {Message};
\node [below of=message2, font=\Huge, rotate=90] {...};
\node [box]  (messageN) at (-0.5, -2.5) {Message};
\node [box]  (gat) at (3, 0) {$\aggregateFN$};
\node [box]  (update) at (6.3, 0) {$\updateFN$};


\draw[-stealth] (-1.8, 0) -- (message2.west) node[at start,left]{$\mathbf{h_{2}}^{k-1}$; $\mathbf{h_{j}}^{k-1}$};

\draw[-stealth] (-1.8, 1.5) -- (message1.west) node[at start,left]{$\mathbf{h_{1}}^{k-1}$; $\mathbf{h_{j}}^{k-1}$};

\draw[-stealth] (-1.8, -2.5) -- (messageN.west) node[at start,left]{$\mathbf{h_{n_j}}^{k-1}$; $\mathbf{h_{j}}^{k-1}$};

\draw[-stealth] (message1.east) -- (gat.west) node[near start,right]{$\mathbf{m_{1,j}}^{k-1}$};

\draw[-stealth] (message2.east) -- (gat.west) node[midway,below]{$\mathbf{m_{2,j}}^{k-1}$};

\draw[-stealth] (messageN.east) -- (gat.west) node[near start,right]{$\mathbf{m_{{n_j},j}}^{k-1}$};

\draw[-stealth] (gat.east) -- (update.west) node[midway,above]{$\mathbf {m_j}^{k-1}$};

\draw[-stealth] (update.east) -- (7.6, 0) node[at end,right]{$\mathbf{h_{j}}^{k}$};

\draw[-stealth] (6.3, -1.0) -- (update.south) node[at start,below]{$\mathbf{h_{j}}^{k-1}$};

\end{scope}
\end{tikzpicture}
\caption{A GNN layer, which represents a single message passing iteration, includes multiple trainable functions, depicted as yellow rectangles. The number of first-order neighbours of the node $j$ is denoted as $n_j$.}
    \label{GNNlayerDetails}
\end{figure}

The spatial GNNs perform recursive neighbourhood aggregation, also known as message passing \cite{pmlr-v70-gilmer17a}, over the local subsets of graph-structured inputs to create a meaningful representation of the connected pieces of data. More precisely, a GNN acts as a trainable local graph filter which has a goal of transforming the inputs from each node and its connections to a higher dimensional space, resulting in a $s$-dimensional vector embedding $\mathbf h \in \mathbb {R}^{s}$ per node. In other words, the goal of the node embedding is to represent the information about the node's position in the graph, as well as its own and the input features of the neighbouring nodes. The GNN layer, which implements one iteration of the recursive neighbourhood aggregation, consists of several differentiable functions that can be represented using a trainable set of parameters, usually in the form of the feed-forward neural networks. These functions co-operate to produce updated versions of node embeddings based on the previous ones, as shown in Fig.~\ref{GNNlayerDetails}. We will discuss the role of each of the functions and the intermediate values that they exchange in the continuation of the text. In the rest of the section, nodes into which the messages are aggregated will be denoted with index $j$, while their 1-hop neighbours, which are the sources of the messages, will be denoted with index $i$.

The message function $\messageFN(\cdot|\theta^{\messageFN}): \mathbb {R}^{2s} \mapsto \mathbb {R}^{u}$ outputs the message $\mathbf m_{i,j} \in \mathbb {R}^{u}$ between the embeddings of a pair of connected nodes, $\mathbf h_i$ and $\mathbf h_j$. Many GNN architectures do not explicitly define this function, but instead simply use node embeddings of the 1-hop neighbours $i$ as messages: $\messageFN(\mathbf h_i, \mathbf h_j|\theta^{\messageFN}) = \mathbf h_i$. The expressive power of a GNN model can be increased by including a set of trainable parameters $\theta^{\messageFN}$ to the message function definition \cite{pmlr-v70-gilmer17a}. The message function can also be defined in a way to include the data from the edge input features; however, that consideration is out of the scope of this thesis.

The aggregation function $\aggregateFN(\cdot|\theta^{\aggregateFN}): \mathbb {R}^{\textrm{deg}(j) \cdot u} \mapsto \mathbb {R}^{u}$ defines in which way incoming neighbouring messages are combined, and outputs the aggregated messages denoted as $\mathbf {m_j} \in \mathbb {R}^{u}$ for node $j$. The aggregation function is designed to take the set of messages as an input, which makes it permutation invariant. Some of the commonly used are element-wise average, sum, minimum, and maximum, optionally followed by some kind of trainable function. In some GNN architectures, the incoming messages are weighted before being aggregated. For example, in graph convolutional networks \cite{KipfW17_GCN}, the messages are normalized by the product of node degrees of source and target nodes. In some of the more advanced aggregation functions, weights for the messages are learned. One of the most popular examples are GATs \cite{velickovic2018graph}, which will be discussed in greater detail in Subsection \ref{subsec:GraphAttentionNetworks}.

The output of one iteration of the neighbourhood aggregation process is the updated node embedding obtained by applying the update function $\updateFN(\cdot|\theta^{\updateFN}): \mathbb {R}^{u+s} \mapsto \mathbb {R}^{s}$ on the aggregated messages concatenated with the embedding of the node $j$ prior to the update. In this way, the update of the node embedding does not rely only on the aggregated messages, but also on its previous values. When the number of recursive neighbourhood aggregations is large, this can help a GNN model distinguish node embeddings of similar nodes\footnote{However, this problem, known as over-smoothing, limits the use of deep GNN models and is still an open area of GNN research \cite{Chen_2020_oversmoothing}.}. The update function can be implemented using GRUs or LSTM units, in which node embedding values are maintained as a hidden state while aggregated messages are taken as new inputs during multiple neighbourhood aggregations \cite{yuLi2015GatedGNN, Selsam2019LearningAS}.

The recursive neighbourhood aggregation process is repeated a predefined number of iterations $K$, also known as the number of GNN layers, where the initial node embedding values are equal to the $l$-dimensional node input features, linearly transformed to the initial node embedding $\mathbf {h_j}^0 \in \mathbb {R}^{s}$. The iteration that the node embeddings and calculated messages correspond to is indicated by the superscript. One iteration of the neighbourhood aggregation process for the $k^{th}$ GNN layer, depicted in Fig.~\ref{GNNlayerDetails}, can also be described analytically by equations \eqref{gnn_equations}:
\begin{equation}
    \begin{gathered}
        \mathbf {m_{i,j}}^{k-1} = \messageFN( \mathbf {h_i}^{k-1}, \mathbf {h_j}^{k-1})\\
        \mathbf {m_j}^{k-1} = \aggregateFN(\{{\mathbf m_{i,j}}^{k-1} | i \in \mathcal{N}_j\})\\
        \mathbf {h_j}^k = \updateFN(\mathbf {m_j}^{k-1}, \mathbf {h_j}^{k-1})\\
        k \in \{1,\dots,K\},
    \end{gathered}
\end{equation}
where $\mathcal{N}_j$ denotes the $1$-hop neighbourhood of the node $j$, and the vector superscript corresponds to the message passing iteration. Either the same or different trainable parameters can be used across different GNN layers; will consider only the former since it results in a smaller GNN model, and also has a regularization effect (i.e., reduces overfitting) due to parameter sharing.

As an example, we present a simple GNN layer architecture in which the message passing process is described using a single equation:
\begin{equation}
    \begin{gathered}
        \mathbf{h_j}^k = \sigma \left(\mathbf{W}_{\mathrm{self}}^{(k)} \mathbf{h_j}^{k-1} + \mathbf{W}_{\mathrm{neigh}}^{(k)} \mathlarger{\sum\limits_{i \in \mathcal{N}_j}} \mathbf{h_i}^{k-1} \right).
    \end{gathered}
    \label{gnn_equations}
\end{equation}
This GNN layer uses node embeddings of the 1-hop neighbours $\mathbf{h_i}^{k-1}$ as messages, while the aggregation function is defined as the sum of messages linearly transformed using the matrix $\mathbf{W}_{\mathrm{neigh}}^{(k)}$ which contains trainable parameters. Finally, the update function is defined by applying the nonlinear function $\sigma(\cdot)$ element-wise on the sum of the aggregated messages and the current embedding of the node $j$ linearly transformed using an additional trainable matrix $\mathbf{W}_{\mathrm{self}}^{(k)}$.

The outputs of the message passing process are final node embeddings $\mathbf {h_j}^{K}$ which can be used for the classification or regression over the nodes, edges, or the whole graph, or can be used directly for the unsupervised node or edge analysis of the graph. In the case of supervised learning over the nodes, the final embeddings are passed through the additional nonlinear function, creating the outputs that represent the predictions of the GNN model for the set of inputs fed into the nodes and their neighbours. GNN training is performed by optimizing the model parameters using variants of the gradient descent algorithm \cite{Rumelhart1986LearningRB}, with the loss function being some measure of the distance between the labels and the predictions. We refer the reader to \cite{GraphRepresentationLearningBook} for a more comprehensive introduction to graph representation learning and GNNs.

It is important to note that since nearby nodes have a significant overlap of the corresponding $k-$hop neighbourhoods, GNN's message passing process results in similar node embeddings for those nodes by design, even for suboptimal values of trainable parameters. Works \cite{velickovic2018deep, Liu2021Simplifying} report that untrained, randomly initialized GNNs can match the performance of trained random walk-based shallow embedding methods.

\subsection {Graph Attention Networks} \label{subsec:GraphAttentionNetworks}

An important decision to be made while creating the GNN model is to select the GNN's aggregation function. Aggregation functions that are commonly used include sum, average, minimum and maximum pooling, and graph convolution \cite{KipfW17_GCN}. One common drawback of these approaches is that incoming messages from all the node's neighbours are weighted equally, or using weights calculated using the structural properties of the graph (e.g., node degrees) prior to training. GATs \cite{velickovic2018graph} propose using the attention-based aggregation, in which the weights that correspond to the importance of each neighbour's message are learned from their corresponding embeddings, increasing the representational capacity of a GNN model. The weights are calculated using the attention mechanism \cite{Bahdanau2015NeuralMT} which is traditionally used in transformer models for sequential data \cite{attentionAllYouNeed} and has achieved significant success in the field of natural language processing.
 
The attention mechanism introduces an additional set of trainable parameters in the aggregation function, usually implemented as a feed-forward neural network $\attendFN(\cdot|\theta^{\attendFN}): \mathbb {R}^{2s} \mapsto \mathbb {R}$ applied over the concatenated embeddings of the node $j$ and each of its neighbours:
 
\begin{equation}
    e_{i,j} = \attendFN(\mathbf {h_i}, \mathbf {h_j} | \theta^{\attendFN}).
    \label{unnormalized_attention}
\end{equation}

We obtain the final attention weights $a_{i,j} \in \mathbb {R}$ by normalizing the output $e_{i,j} \in \mathbb {R}$ using the softmax function:
 \begin{equation}
    a_{i,j} = \frac{exp({e_{i,j}})}{\sum_{i' \in \mathcal{N}_j} exp({e_{i',j}})}.
    \label{normalized_attention}
\end{equation}

To add a regularization effect and stabilize the learning process, GAT implementations usually include the multi-head attention concept \cite{velickovic2018graph}. It introduces calculating multiple attention weights for each message using separate attention layers, which are trained independently. Multiple replicas of the same message are then aggregated in some way, usually using concatenation followed by a trainable function such as feed-forward neural network.

\section{Practical Aspects of Graph Neural Networks}

This section discusses some of the aspects of applying GNNs to real-world problems, like scalability, mini-batch training, and application to various types of graphs. Additionally, we discuss graph augmentation methods when the graphs are too sparse or lack input features, as well as applying GNNs in a semi-supervised and self-supervised setting, which is useful in the cases of sparsely labelled data.

The GNN model presented in the previous section may suffer from scalability issues when applied to large graphs in which some of the nodes have very high degrees. An example of this kind of graph would be social networks, in which nodes that correspond to popular members have a large number of connections. The problem arises during the $K$- hop neighbourhood aggregation process, which would aggregate neighbours of high-degree nodes multiple times, resulting in large computational graphs and high training and inference computational complexity. GraphSAGE \cite{GraphSAGE} is one of the first GNN models successfully applied to large graphs, with one of his variants deployed at the Pinterest's recommendation system over the graph containing 3 billion nodes and 18 billion edges \cite{PINSAGE}. During both the training and inference processes, GraphSAGE employs stochastic neighbourhood aggregation, in which only a subset of neighbouring messages is calculated during each message passing iteration. However, sampling is not done uniformly at random, but with a strategy in which important neighbours are sampled with a higher probability instead of numerous low-degree nodes. As a side effect, stochastic neighbourhood aggregation improves robustness to the changes in the graph during the inference time, but also increases the variance of the training process, making it less stable.

Deep learning models are usually trained using a mini-batch gradient descent algorithm, which calculates a loss function and model updates over a group (i.e., a mini-batch) of training examples rather than over individual examples or a whole training dataset. This strategy can be easily applied in the case of GNNs when graphs have small sizes, allowing multiple graphs from the dataset to be grouped together in mini-batches without violating memory constraints of the hardware being used for the training. The issue of how to split up graph elements into mini-batches arises in the case of large graphs that require more memory than is available. Making the subgraphs using graph cuts is a simple solution, but doing so results in the loss of connectivity information between the nodes in different subgraphs, which disables message passing between them. GraphSAGE authors \cite{GraphSAGE} propose grouping $K$-hop neighbourhoods of nodes into mini-batches, in which each mini-batch can include multiple nodes with their neighbourhoods. Although this approach may result in the use of overlapping neighbourhoods, it preserves the message passing in the original graph while allowing the training process to meet the memory requirements.

The GNN methods previously presented can be tailored to specific types of graphs that may arise in practise.
Without going into too much detail, here are some examples of how the neighbourhood aggregation process needs to be modified for the specific use cases: 
\begin{itemize}
    \item Directed graphs introduce a notion of direction to each edge. In social networks, an example would be the concept of following, in which one member of the network can be connected to another but not vice versa. In these cases, the process of aggregation into a node is typically carried out by gathering only the incoming messages defined by the edge directions.
    \item Temporal graphs are used to model dynamic networks in which nodes and edges appear and disappear over time. Temporal graph networks \cite{tgn_icml_grl2020} introduce the spatio-temporal aggregation which gathers the messages from the current and past 1-hop neighbours in each time step. Every node is given a memory component that serves as a representation of its interaction history, much like the hidden state concept in recurrent neural networks. The message between two nodes is determined by the current values of their memory components, the interaction time step, and, if present, the edge features. The aggregation function employs the attention mechanism, which computes the importance score for all 1-hop neighbours throughout the history.    
    \item Heterogeneous and multi-edge graphs contain different types of nodes and edges. GNNs can use distinct sets of trainable parameters for aggregation and update functions that operate over different types of nodes. Likewise, different message functions can be used for different edge types.
    \item Graph-structured data can contain hierarchical information, like in the cases of networks of molecules \cite{networksOfMolecules} or object-oriented data models. These data structures can be efficiently represented as hypernode-based nested graphs \cite{nestedGraphs1994}, which are composed of hypernodes, which are themselves graphs. Learning over these types of graphs is done using multiple levels of GNNs, which operate at different hierarchy levels and exchange information at particular message passing iterations.
\end{itemize}

Because data in real-world problems is typically sparsely labelled due to high labelling costs, techniques that make the best use of both unlabelled and labelled data samples are required. Semi-supervised learning techniques fall somewhere between supervised and unsupervised learning, with the goal of making predictions on unseen data based on small sets of labelled data while utilizing large amounts of unlabelled data. The self-training method \cite{selfTraining}, also known as self-labelling, is a simple form of semi-supervised learning for classification problems in which an initial classifier is trained on a small labelled dataset and then used to label the rest of the data using the most confident predictions. Then, using the larger labelled dataset, a new classifier is trained, and the process is repeated several times. More advanced approaches employ generative models to learn the distribution of real-world data from unlabelled samples \cite{semiSupervisedGAN2016Odena}. Generative models are then used to enrich the labelled dataset by generating new data samples with the desired label. GNNs can adapt well to the semi-supervised concept when dealing with node-level prediction problems in graph-structured data, where labels are only available to a small subset of nodes \cite{KipfW17_GCN}. In this scenario, the GNN model is trained by backpropagating the supervised learning loss function calculated using only the labelled nodes. However, because all nodes and their incident edges participate in the message passing process, the node input features and connectivity data from unlabelled nodes also contribute to prediction generation.

Self-supervised learning \cite{selfSupervised2021Liu} is another type of methods that falls somewhere between supervised and unsupervised learning. It uses unlabelled data samples to create a supervised learning objective using pseudo-labels, resulting in useful data representations (i.e., embeddings) that can be used as inputs for other, so-called downstream tasks. Similarly to semi-supervised learning, it became popular because of its ability to avoid annotating large amounts of data. Because no manual data labelling is required, some classifications consider self-supervised learning to be a subset of unsupervised learning. One of the most common forms of self-supervised learning is contrastive learning \cite{chen2020ContrastiveLearning}, in which data samples are associated with positive and negative pseudo-labels automatically. A subset of data samples, for example, can be intentionally corrupted by adding some noise and labelled as negative, whereas uncorrupted data samples are labelled as positive. Models defined in this manner are trained using the binary classification loss (e.g., cross-entropy) as a learning objective and produce data representations that can be reused for downstream tasks as an intermediate result. 

To make use of unlabelled graph-structured data, GNNs can also be used in self-supervised setting. Link prediction can be thought of as a simple form of self-supervised contrastive learning based on GNNs. It is essentially a binary classification task in which pairs of nodes are sampled, and classification labels are generated based on the presence of direct connections between the nodes. This task can be generalized so that positive pseudo-labels indicate that a pair of nodes is close in terms of some neighbourhood metric (e.g., $k$-hop or random walk-based), and negative pseudo-labels indicate that they are not. Node embeddings obtained using these positive and negative label definitions can be useful for tasks which rely on local graph information. To obtain more relevant node embeddings for global tasks, contrastive learning methods that distinguish between original and corrupted graphs can be used \cite{velickovic2018deep}.

We conclude the overview of the practical aspects of GNNs by discussing the augmentation methods that are commonly used with graph-structured data. In real-world problems, the original, raw graph data is rarely used as an input for GNNs without being augmented in some way. Since nodes in a graph can lack input features, feature augmentation in the form of expanding the node input feature vectors is often used. Let's consider a case of a graph in which nodes do not have any input features. A trivial form of feature augmentation is adding a constant feature (e.g., a scalar with a value of $1.0$) to every node. As a result, the GNN is able to learn the structural information of a graph using the aggregation function, which is not possible when aggregating zeroes. More advanced types of feature augmentation, such as one-hot vector encoding, cycle counts, node centralities, and clustering coefficients, can reveal more information about the graph structure and increase the expressive power of GNN. Aside from feature augmentation, graph augmentation is used when the input graph is either too sparse, making the message passing process inefficient, or has long-range dependencies between nodes. Some common graph augmentation methods include connecting pairs of nodes using virtual edges and adding virtual nodes that connect all the nodes in a certain subset.

\chapter{Graph Neural Network-based State Estimation}	\label{ch:gnn_se}
\addcontentsline{lof}{chapter}{4 Graph Neural Network-based State Estimation}

In this chapter, we explain how GNNs can be applied to both linear and nonlinear SE, using the power system factor graph-like structures. Since we used similar methodologies for both linear and nonlinear formulations, they are presented simultaneously, with specific differences highlighted as necessary. First, we present the augmentation techniques for the power system's factor graph, then the details of the proposed GNN architecture, and analyse the computational complexity and the distributed implementation of the GNN model's inference. Finally, we demonstrate the effectiveness of our proposed method through numerical evaluations on various test cases.

\section{Power System Factor Graph Augmentation} 

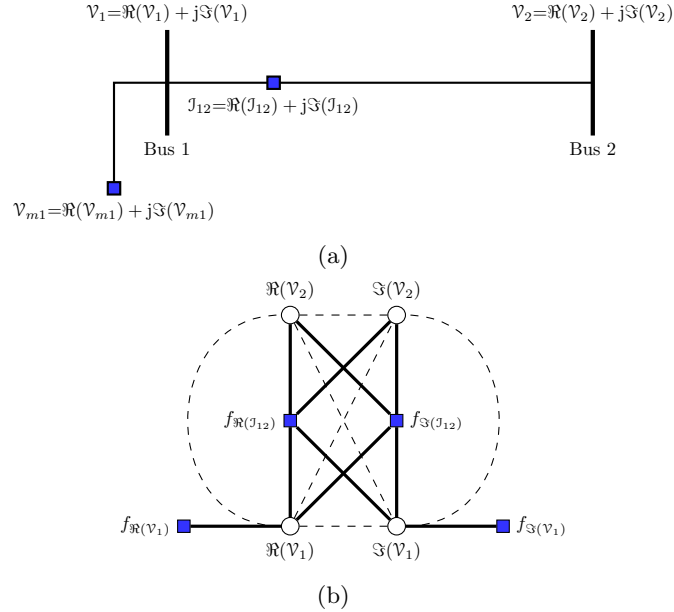
\begin{figure}[htbp]
    \centering
    \subfloat[]{
    \centering
    \begin{tikzpicture}[thick, scale=0.7, transform shape]
        \tikzset{
            factorVoltage/.style={draw=black,fill=blue!80, minimum size=2mm},
            factorCurrent/.style={draw=black,fill=blue!80, minimum size=2mm}}
        \draw (0,0) 
            node[factorVoltage, label=below:$\ph V_{m1} \mathrm{=} \Re({\ph V}_{m1}) + \mathrm{j} \Im({\ph V}_{m1}) $](L11){} 
            -- ++(0,2)
            -- ++(1,0)
            \bushere{${\ph V}_1 \mathrm{=} \Re {(\ph V_1)} + \mathrm{j} \Im{(\ph V_1)}$}{Bus 1}
            -- ++(2,0)
            node[factorCurrent, label=below:$\ph I_{12} \mathrm{=} \Re {(\ph I_{12})} + \mathrm{j} \Im{(\ph I_{12})}$](L2){} 
            to[] ++(6,0)
            \bushere{${\ph V}_2 \mathrm{=} \Re {(\ph V_2)} + \mathrm{j} \Im{(\ph V_2)}$}{Bus 2}
            ;
    \end{tikzpicture}
    }
    \hfil
    \subfloat[]{
    \begin{tikzpicture} [scale=0.7, transform shape]
        \tikzset{
            varNode/.style={circle,minimum size=2mm,fill=white,draw=black},
            factorVoltage/.style={draw=black,fill=blue!80, minimum size=2mm},
            factorCurrent/.style={draw=black,fill=blue!80, minimum size=2mm},
            edge/.style={very thick,black},
            edge2/.style={dashed,black}}
        \begin{scope}[local bounding box=graph]
            \node[factorVoltage, label=left:$f_{\Re({\ph V}_1)}$] (f1) at (-3, 1 * 2) {};
            \node[factorVoltage, label=right:$f_{\Im({\ph V}_1)}$] (f4) at (3, 1 * 2) {};
            \node[varNode, label=below:$\Re({\ph V}_1)$] (v1) at (-1, 1 * 2) {};
            \node[varNode, label=below:$\Im({\ph V}_1)$] (v3) at (1, 1 * 2) {};
            \node[factorCurrent, label=left:$f_{\Re({\ph I}_{12})}$] (f2) at (-1, 2 * 2) {};
            \node[factorCurrent, label=right:$f_{\Im({\ph I}_{12})}$] (f5) at (1, 2 * 2) {};
            \node[varNode, label=above:$\Re({\ph V}_2)$] (v2) at (-1, 3 * 2) {};
            \node[varNode, label=above:$\Im({\ph V}_2)$] (v4) at (1, 3 * 2) {};
            
            \draw[edge] (f1) -- (v1);
            \draw[edge] (f4) -- (v3);
            \draw[edge] (f2) -- (v1);
            \draw[edge] (f5) -- (v3);
            \draw[edge] (f5) -- (v1);
            \draw[edge] (f2) -- (v3);
            \draw[edge] (f2) -- (v2);
            \draw[edge] (f2) -- (v4);
            \draw[edge] (f5) -- (v2);
            \draw[edge] (f5) -- (v4);
            \draw[edge2] (v1) to [out=180,in=180,looseness=1.5] (v2);
            \draw[edge2] (v1) -- (v3);
            \draw[edge2] (v1) -- (v4);
            \draw[edge2] (v2) -- (v3);
            \draw[edge2] (v2) -- (v4);
            \draw[edge2] (v3) to [out=0,in=0,looseness=1.5] (v4);
            
        \end{scope}
    \end{tikzpicture}
    }
    \caption{Subfigure (a) shows a simple two-bus power system with two phasor measurements from a PMU placed at the bus 1. Subfigure (b) displays the corresponding factor graph (full-line edges) and augmented factor graph (all edges). Variable nodes are depicted as circles, and factor nodes are as squares.}
    \label{toyFactorGraph}
    
\end{figure}

\begin{figure}[htbp]
    \centering
    \subfloat[]{
    \centering
    \begin{tikzpicture}[thick, scale=0.7, transform shape]
        \tikzset{
            factorSCADA/.style={draw=black,fill=green!80, minimum size=4mm},
            factorPMU/.style={draw=black,fill=blue!80, minimum size=4mm}}
        \draw (0,0) 
            node[factorPMU, label=below:${\ph V}_{m2} \mathrm{=} V_{m1}
            \text{e}^{\text{j}\theta_{m1}}$](L11){} 
            -- ++(0,2)
            -- ++(1,0)
            \bushere{$\ph{V}_1 = V_1 \text{e}^{\text{j}\theta_1}$}{Bus 1} -- ++(1.5,0)
            node[factorPMU, label=below:$\ph I_{12} \mathrm{=} I_{12}  \text{e}^{\text{j}\theta_{I_{12}}}$](L1){}
            -- ++(2,0)
            node[factorSCADA, label=below:$P_{12}$](L2){} 
            to[] ++(3,0)
            \bushere{$\ph{V}_2 = V_2 \text{e}^{\text{j}\theta_2}$}{Bus 2}
            -- ++(1,0) 
            -- ++(0,-2) 
            node[factorSCADA, label=below:$V_{m2}$](L22){} 
            ;
    \end{tikzpicture}
    }
    \hfil
    \subfloat[]{
    \begin{tikzpicture} [scale=0.81, transform shape]
        \tikzset{
            varNode/.style={circle,minimum size=2mm,fill=white,draw=black},
            factorPMU/.style={draw=black,fill=blue!80, minimum size=2mm},
            factorSCADA/.style={draw=black,fill=green!80, minimum size=2mm},
            edge/.style={very thick,black},
            edge2/.style={dashed,black}}
        \begin{scope}[local bounding box=graph]
            \node[factorPMU, label=left:$f_{V_{m1}}$] (f1) at (-3, 1 * 2) {};
            \node[factorPMU, label=right:$f_{\theta_{m1}}$] (f4) at (3, 1 * 2) {};
            \node[varNode, label=below:$V_1$] (v1) at (-1, 1 * 2) {};
            \node[varNode, label=below:$\theta_1$] (v3) at (1, 1 * 2) {};
            \node[factorPMU, label=left:$f_{I_{12}}$] (f2) at (-1, 2 * 2) {};
            \node[factorPMU, label=right:$f_{\theta_{I_{12}}}$] (f5) at (1, 2 * 2) {};
            \node[varNode, label=above:$V_2$] (v2) at (-1, 3 * 2) {};
            \node[varNode, label=above:$\theta_2$] (v4) at (1, 3 * 2) {};
            
            \node[factorSCADA, label=left:$f_{V_{m2}}$] (f6) at (-3, 3 * 2) {};
            \node[factorSCADA, label=right:$f_{P_{12}}$] (f7) at (4, 2 * 2) {};
            
            \draw[edge] (f1) -- (v1);
            \draw[edge] (f4) -- (v3);
            \draw[edge] (f2) -- (v1);
            \draw[edge] (f5) -- (v3);
            \draw[edge] (f5) -- (v1);
            \draw[edge] (f2) -- (v3);
            \draw[edge] (f2) -- (v2);
            \draw[edge] (f2) -- (v4);
            \draw[edge] (f5) -- (v2);
            \draw[edge] (f5) -- (v4);
            
            \draw[edge] (f6) -- (v2);
            \draw[edge] (f7) -- (v1);
            \draw[edge] (f7) -- (v2);
            \draw[edge] (f7) -- (v3);
            \draw[edge] (f7) -- (v4);
            \draw[edge2] (v1) to [out=180,in=180,looseness=1.5] (v2);
            \draw[edge2] (v1) -- (v3);
            \draw[edge2] (v1) -- (v4);
            \draw[edge2] (v2) -- (v3);
            \draw[edge2] (v2) -- (v4);
            \draw[edge2] (v3) to [out=0,in=0,looseness=1.5] (v4);
            
        \end{scope}
    \end{tikzpicture}
    }
    \caption{Subfigure (a) shows a simple two-bus power system containing a PMU at the bus 1, one legacy active power flow measurement, and one legacy voltage magnitude measurement at the bus 2. Subfigure (b) displays the corresponding factor graph (full-line edges) and augmented factor graph (all edges). Variable nodes are represented as circles, and factor nodes are depicted as squares, coloured differently to distinguish between phasor and legacy measurements.}
    \label{toyFactorGraphNonlinear}
    
\end{figure}
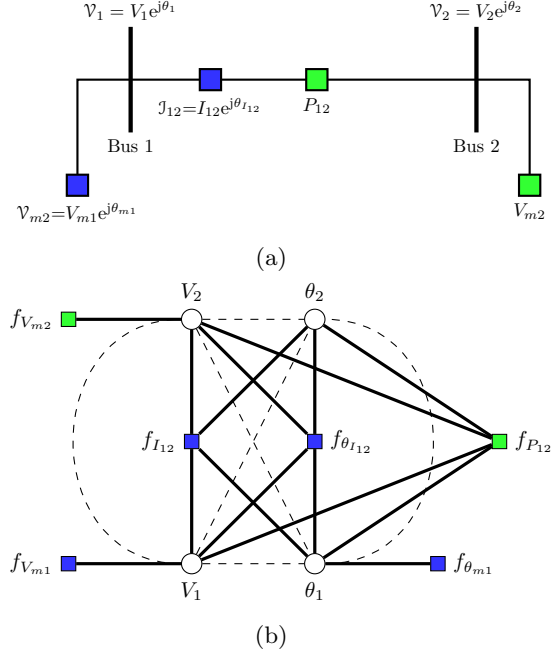

Inspired by recent work on using probabilistic graphical models for power system SE \cite{cosovic2019bpse}, we first create a GNN over a graph with a factor graph topology. This bipartite graph consists of factor and variable nodes connected by edges, and it is established in accordance with different SE problem formulations:
\begin{itemize}
    \item \textbf{Linear SE:} Variable nodes are used to calculate a $s$-dimensional node embedding for all real and imaginary parts of the bus voltages, $\Re(\ph V_i)$ and $\Im(\ph V_i)$, $i=1,\dots,n$, using which state variable predictions can be generated. Factor nodes, two per each measurement phasor, serve as inputs for the measurement values, variances, and covariances, also given in rectangular coordinates. These values are then transformed and sent to variable nodes via GNN message passing. Unlike the approximative WLS SE defined in \ref{linearSEbackground}, which neglects measurement covariances, the GNN includes them, leading to accurate solutions without increasing the computational complexity. The upper subfigure of Fig.~\ref{toyFactorGraph} illustrates a two-bus power system, with a PMU on the first bus, containing one voltage and one current phasor measurement. The nodes connected by full lines represent the corresponding factor graph in the lower subfigure. 

    \item \textbf{Nonlinear SE:} In this case, pairs of variable nodes generate $s$-dimensional node embeddings for magnitudes ${V}_i$ and phase angles ${\theta}_i$ of complex bus voltages $\ph V_i = V_{i}\mathrm{e}^{\mathrm{j}\theta_{i}}$. The inputs to the factor nodes are the values and variances of both phasor and legacy measurements. Phasor measurements are expressed in polar coordinates, which eliminates any correlation between measurement errors. Therefore, in contrast to linear SE, in nonlinear SE measurement covariances do not need to be included as inputs into factor nodes. When creating the factor graph from the bus-branch power system model, each phasor measurement generates two factor nodes, while each legacy measurement generates one factor node. As an example, in the upper subfigure of Fig.~\ref{toyFactorGraphNonlinear} we consider a simple two-bus power system, in which we placed one voltage phasor measurement on the first bus and one legacy voltage magnitude measurement on the second bus. Additionally, we placed one current phasor measurement and one legacy active power flow measurement on the branch connecting the two nodes. The factor graph of this simple power system consists of the generated factor and variable nodes, connected by full-line edges, as shown in the lower subfigure of Fig.~\ref{toyFactorGraphNonlinear}.
\end{itemize}

Unlike approaches that apply GNNs over the bus-branch power system model, such as in \cite{pagnier2021physicsinformed, TGCN_SE_2021}, the using GNNs over factor-graph-like topology allows for the incorporation of various types and quantities of measurements on both power system buses and branches. The ability to simulate the addition or removal of various measurements can be easily achieved by adding or removing factor nodes from any location in the graph. In contrast, using a GNN over the bus-branch power system model would require allocating a single input vector to each bus that includes all potential measurement data for that bus and its neighbouring branches. This can cause problems, such as having to fill elements in the input vector with zeros when not all possible measurements are available, and making the output sensitive to the order of measurements in the input vector. This can make it difficult to accurately model the system and generate reliable results.

Augmenting the factor graph topology by connecting the variable nodes in the $2$-hop neighbourhood significantly improves the model's prediction quality in unobservable scenarios. This is because the graph should remain connected even when we remove factor nodes to simulate measurement loss. This will allow the messages to be still propagated in the whole $K$-hop neighbourhood of the variable node. In other words, a factor node corresponding to a branch current measurement can be removed while still preserving the physical connection between the power system buses. This requires an additional set of trainable parameters for the variable-to-variable message function. Although the augmented factor graph displayed with both full and dashed lines in Figs.~\ref{toyFactorGraph} and \ref{toyFactorGraphNonlinear} is not a factor graph because it is no longer bipartite, we will still refer to the nodes as factor and variable nodes for simplicity.

To achieve better representation of node's neighbourhood structure, we perform variable node feature augmentation using binary index encoding. Since variable nodes have no additional input features, this encoding allows the GNN to better capture the relationships between nodes. Compared to one-hot encoding used in \cite{kundacina2022state}, binary index encoding significantly reduces the number of input neurons and trainable parameters, as well as training and inference time.

\section{Proposed GNN Architecture} \label{subsec:gnnBasedSe}
Since the proposed GNN operates on a heterogeneous graph, we use two different types of GNN layers: one for aggregation in factor nodes, and one for variable nodes. These layers, denoted as $\layerf(\cdot|\theta^{\layerf}): \mathbb {R}^{\textrm{deg}(f)  \cdot s} \mapsto \mathbb {R}^{s}$ and $\layerv(\cdot|\theta^{\layerv}): \mathbb {R}^{\textrm{deg}(v) \cdot s} \mapsto \mathbb {R}^{s}$, have their own sets of trainable parameters  $\theta^{\layerf}$ and $\theta^{\layerv}$, allowing their message, aggregation, and update functions to be learned separately. Additionally, we use different sets of trainable parameters for variable-to-variable and factor-to-variable node messages, $\textrm{Message}\textsuperscript{f$\rightarrow$v}(\cdot|\theta^{\textrm{Message}\textsuperscript{f$\rightarrow$v}}): \mathbb {R}^{2s} \mapsto \mathbb {R}^{u}$ and $\textrm{Message}\textsuperscript{v$\rightarrow$v}(\cdot|\theta^{\textrm{Message}\textsuperscript{v$\rightarrow$v}}): \mathbb {R}^{2s} \mapsto \mathbb {R}^{u}$, in the $\layerv(\cdot|\theta^{\layerv})$ layer. In both GNN layers, we use two-layer feed-forward neural networks as message functions, single layer neural networks as update functions and the attention mechanism in the aggregation function. Furthermore, we apply a two-layer neural network $\pred(\cdot|\theta^{\pred}): \mathbb {R}^{s} \mapsto \mathbb {R}$ to the final node embeddings $\mathbf h^K$ of variable nodes only, to create state variable predictions $\mathbf{x^{pred}}$. For factor and variable nodes with indices $f$ and $v$, neighbourhood aggregation and state variable prediction can be described as: 
\begin{equation}
    \begin{gathered}
        \mathbf {h_v}^k = \layerv(\{\mathbf {h_i}^{k-1} | i \in \mathcal{N}_v\} | \theta^{\layerv})\\
        \mathbf {h_f}^k = \layerf(\{\mathbf {h_i}^{k-1} | i \in \mathcal{N}_f\} | \theta^{\layerf})\\
        {x_v}^{pred} = \pred(\mathbf {h_v}^K|\theta^{\pred})\\
        k \in \{1,\dots,K\},
    \end{gathered}
    \label{embeddingds_and_predictions}
\end{equation}
where $\mathcal{N}_v$ and $\mathcal{N}_f$ denote the $1$-hop neighbourhoods of the nodes $v$ and $f$. All the trainable parameters $\theta$ of the GNN are updated by applying gradient descent, using backpropagation, to a loss function calculated over a mini-batch of graphs. This loss function is the mean-squared difference between the predicted state variables and their corresponding ground-truth values:
\begin{equation} \label{loss_function}
    \begin{gathered}
        L(\theta) = \frac{1}{2nB} \sum_{i=1}^{2nB}({{x_i}^{pred}} - {{x_i}^{label}})^2 \\
        \theta = \{\theta^{\layerv} \mathop{\cup} \theta^{\layerf} \mathop{\cup} \theta^{\pred}\} \\
        \theta^{\layerv} = \{\theta^{\textrm{Message}\textsuperscript{f$\rightarrow$v}} \mathop{\cup} \theta^{\textrm{Message}\textsuperscript{v$\rightarrow$v}} \mathop{\cup} \theta^{\textrm{Aggregate}\textsuperscript{v}} \mathop{\cup} \theta^{\textrm{Update}\textsuperscript{v}}\}\\
        \theta^{\layerf} = \{\theta^{\textrm{Message}\textsuperscript{v$\rightarrow$f}} \mathop{\cup} \theta^{\textrm{Aggregate}\textsuperscript{f}} \mathop{\cup} \theta^{\textrm{Update}\textsuperscript{f}}\},
    \end{gathered}
\end{equation}
where the total number of variable nodes in a graph is $2n$, and the number of graphs in the mini-batch is $B$. In this work, we chose the loss function for training the GNN based on the fundamental SE problem, where state variables are obtained from available measurement information. However, if there is a requirement to include additional constraints in the SE calculation, it is possible to achieve this by adding new terms to the loss function defined in \eqref{loss_function}. For example, the loss function can be augmented with the power balance error at each bus where the constraints are imposed, in addition to minimizing the prediction error from the labels. A similar approach has been proposed in \cite{LOPEZ2023GNNpowerFlow}, where the power flow problem is solved using GNN by minimizing the power balance errors at each bus. Adding additional constraints (e.g., zero injection constraints) to the GNN SE loss function can improve the SE results, especially in distributed power systems with limited measurement coverage.

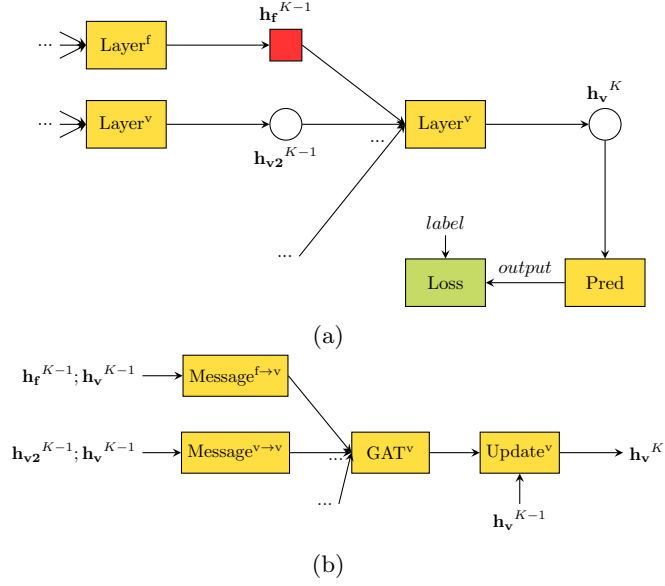
\begin{figure}[!t]
\centering
\subfloat[]{
\begin{tikzpicture} [scale=0.7, transform shape]

\tikzset{
    varNode/.style={circle,minimum size=6mm,fill=white,draw=black},
    factorVoltage/.style={draw=black,fill=red!80, minimum size=6mm},
    factorCurrent/.style={draw=black,fill=orange!80, minimum size=6mm},
    box/.style={draw, fill=Goldenrod, minimum width=1.5cm, minimum height=0.9cm}}

\begin{scope}[local bounding box=graph]

\node [box]  (layerF0) at (-3, 1.5) {$\layerf$};
\node [box]  (layerV0) at (-3, 0) {$\layerv$};

\node[factorVoltage, label=above:$\mathbf{h_{f}}^{K-1}$] (facReV1) at (0, 1.5) {};
\node[varNode, label=below:$\mathbf{h_{v2}}^{K-1}$] (varImV1) at (0, 0) {};
\node[varNode, label=above:$\mathbf{h_{v}}^{K}$] (varReV1) at (6, 0) {};

\node [box]  (layerV1) at (3, 0) {$\layerv$};
\node [box]  (pred) at (6, -3) {$\pred$};
\node [box, fill=SpringGreen]  (loss) at (3, -3) {Loss};

\draw[-stealth] (-4.25, 1.75) -- (layerF0.west);
\draw[-stealth] (-4.25, 1.5) -- (layerF0.west) node[at start,left]{$...$};
\draw[-stealth] (-4.25, 1.25) -- (layerF0.west);

\draw[-stealth] (-4.25, 0.25) -- (layerV0.west);
\draw[-stealth] (-4.25, 0) -- (layerV0.west) node[at start,left]{$...$};
\draw[-stealth] (-4.25, -0.25) -- (layerV0.west);

\draw[-stealth] (layerF0.east) -- (facReV1.west);
	
\draw[-stealth] (layerV0.east) -- (varImV1.west);
	
\draw[-stealth] (varImV1.east) -- (layerV1.west);
	
\draw[-stealth] (0.25, -2.5) -- (layerV1.west)
	node[at start,left]{$...$} node[very near end,left]{$...$};
	
\draw[-stealth] (facReV1.east) -- (layerV1.west);
	
\draw[-stealth] (layerV1.east) -- (varReV1.west);
	
\draw[-stealth] (varReV1.south) -- (pred.north);
	
\draw[-stealth] (pred.west) -- (loss.east)
	node[midway,above]{$output$};
	
\draw[-stealth] (3, -2.1) -- (loss.north)
	node[at start,above]{$label$};

\end{scope}

\end{tikzpicture}
}
\hfil
\subfloat[]{

\begin{tikzpicture} [scale=0.68, transform shape]

\tikzset{
    box/.style={draw, fill=Goldenrod, minimum width=1.5cm, minimum height=0.8cm}}
    
\begin{scope}[local bounding box=graph]

\node [box]  (message1) at (0, 1.5) {Message\textsuperscript{f$\rightarrow$v}};
\node [box]  (message2) at (0, 0) {Message\textsuperscript{v$\rightarrow$v}};
\node [box]  (gat) at (3, 0) {\textrm{GAT}\textsuperscript{v}};
\node [box]  (update) at (5.5, 0) {\textrm{Update}\textsuperscript{v}};


\draw[-stealth] (-1.8, 0) -- (message2.west) node[at start,left]{$\mathbf{h_{v2}}^{K-1}; \mathbf{h_{v}}^{K-1}$};

\draw[-stealth] (-1.8, 1.5) -- (message1.west) node[at start,left]{$\mathbf{h_{f}}^{K-1}; \mathbf{h_{v}}^{K-1}$};

\draw[-stealth] (message1.east) -- (gat.west);

	
\draw[-stealth] (message2.east) -- (gat.west);
	
\draw[-stealth] (2, -1) -- (gat.west)
	node[at start,left]{$...$} node[very near end,left]{$...$};
	
\draw[-stealth] (gat.east) -- (update.west);

\draw[-stealth] (update.east) -- (7.5, 0) node[at end,right]{$\mathbf{h_{v}}^{K}$};

\draw[-stealth] (5.5, -1.0) -- (update.south) node[at start,below]{$\mathbf{h_{v}}^{K-1}$};

\end{scope}

\label{layerDetails}

\end{tikzpicture}
}
\caption{Subfigure (a) shows a high-level computational graph that starts with the loss function for the output of a variable node $v$. Subfigure (b) depicts the detailed structure of a single GNN $\layerv$. Functions with trainable parameters are highlighted in yellow.}
    \label{computationalGraph}
\end{figure}

Fig.~\ref{computationalGraph} shows the high-level computational graph for the output of a variable node from the augmented factor graph given in Fig.~\ref{toyFactorGraph}. For simplicity, only one unrolling of the neighbourhood aggregation is shown, as well as only the details of the parameters $\theta^{\layerv}$.

\subsection{Computational Complexity and Distributed Inference} 
Because the node degree in the power system graph does not increase with the total number of nodes, the same is true for the node degrees in the augmented factor graph. This means that the inference time per variable node remains constant, as it only requires information from the node's $K$-hop neighbourhood, whose size also does not increase with the total number of nodes. This implies that the computational complexity of inferring all state variables is $\mathcal{O}(n)$. To avoid the over-smoothing problem in GNNs \cite{Chen_2020_oversmoothing}, a small value is assigned to $K$, thus not affecting the overall computational complexity of the inference.

To make the best use of the proposed approach for large-scale power systems, the inference should be performed in a computationally and geographically distributed manner. This is necessary because the communication delays between the PMUs and the central processing unit can hinder the full utilization of the PMUs' high sampling rates. The distributed implementation is possible as long as all the measurements within a node's $K$-hop neighbourhood in the augmented factor graph are fed into the computational module that generates the predictions. For any arbitrary $K$, the GNN inference method only requires measurements that are physically located within the $\lceil K/2 \rceil$-hop neighbourhood of the power system bus.

\section{Numerical results}
In this section, we conduct comprehensive numerical tests to evaluate the effectiveness of proposed augmented factor graph-based GNN approaches for linear and nonlinear SE problems. We used the IGNNITION framework \cite{pujolperich2021ignnition} for building and utilizing GNN models, with the hyperparameters presented in Table~\ref{tbl_hyperparameters}, the first three of which were obtained with the grid search hyperparameter optimization using the Tune tool \cite{liaw2018tune}. All the results presented in this section are normalized using the corresponding nominal voltages in the test power systems and a base power of 100 MVA.

\begin{table}[!t]
\caption{List of GNN hyperparameters.}
\label{tbl_hyperparameters}
\begin{center}
    \begin{tabular}{ | l | c | }
        \hline
        \textbf{Hyperparameters} & \textbf{Values} \\ 
        \hline
        Node embedding size $s$ & $64$ \\
        \hline
        Learning rate & $4\times 10^{-4}$ \\
        \hline
        Minibatch size $B$ & $32$ \\
        \hline
        Number of GNN layers $K$ & $4$ \\
        \hline
        Activation functions & ReLU \\
        \hline
        Gradient clipping value & $0.5$ \\
        \hline
        Optimizer & Adam \\
        \hline
        Batch normalization type & Mean \\
        \hline
    \end{tabular}
\end{center}
\vspace{-6mm}
\end{table}

\subsection{Linear State Estimation} \label{subsec:linearSEresults}
To evaluate the proposed GNN-based linear SE, we create various scenarios using the IEEE 30-bus system, the IEEE 300-bus system, and the ACTIVSg 2000-bus system \cite{ACTIVSg2000}, on which the GNN model is trained and tested. Training, validation, and test sets are obtained using WLS solutions of the system described in (\ref{SE_system_of_lin_eq}) to label various samples of input measurements. Measurements are obtained by adding Gaussian noise to the exact power flow solutions, with each calculation performed using a different, randomly sampled load profile to capture a wide range of power system states. Due to the strong interpolation and extrapolation abilities of GNNs \cite{xu2021how}, our method of randomly sampling from a wide diversity of loads for training examples is effective for generalizing the GNN algorithm for state estimation under varying load conditions. GNN models are tested in three different situations: i) optimal number of PMUs (minimal measurement redundancy, for which the WLS SE offers a solution); ii) underdetermined scenarios; iii) scenarios with maximal measurement redundancy. We also compare the proposed approach with more conventional deep neural network (DNN)-based SE algorithms and assess its scalability, sample efficiency, and robustness to outliers.

\subsubsection{Power System With Optimally Placed PMUs} \label{subsec:optPlacedPMUs}

In this subsection, we conduct a series of experiments on the IEEE 30-bus power system, using measurement variances of $10^{-5}$, $10^{-3}$, and $10^{-1}$ for the creation of the training, validation, and test sets. The number and positions of PMUs are fixed and determined using the optimal PMU placement algorithm \cite{optimalPMUPlacement}, which finds the smallest set of PMUs that make the system observable. This algorithm has resulted in a total of $10$ PMUs and $50$ measurement phasors, $10$ of which are voltage phasors and the rest are current phasors.

Table \ref{tbl_MSEsIEEE30} shows the $100$-sample test set results for all the experiments on the IEEE 30-bus power system, in the form of average mean square errors (MSEs) between the GNN predictions and the test set labels. These results are compared with the average MSE between the labels and the approximate WLS SE solutions defined in Sec. \ref{linearSEbackground}. The results show that for systems with optimally placed PMUs and low measurement variances, GNN predictions have very small deviations from the exact WLS SE, although they are outperformed by the approximate WLS SE. For higher measurement variances, GNN has a lower estimation error than the approximate WLS SE, while also having lower computational complexity in all cases.

\begin{table}[htbp]
\caption{Comparison of GNN and approximative SE test set MSEs for various measurement variances.}
\begin{center}
    \begin{tabular}{ | c | c | c | }
        \hline
        \textbf{Variances} & \textbf{GNN}  & \textbf{Approx. SE} \\
        \hline
        $10^{-5}$ & $2.48\times 10^{-6}$ & $1.87\times 10^{-8}$ \\
        \hline
        $10^{-3}$ & $8.21\times 10^{-6}$ & $2.25\times 10^{-6}$ \\
        \hline
        $10^{-1}$ & $7.47\times 10^{-4}$ & $2.27\times 10^{-3}$ \\
        \hline
    \end{tabular}
\label{tbl_MSEsIEEE30}
\end{center}
\end{table}

\begin{figure}[htbp]
    \centering
    \pgfplotstableread[col sep = comma,]{chapter_04/dataLinear/PredictionsPerNode0Excluded.csv}\CSVPredVSObservedZeroExcludedEdgesVarNodes
    
        \begin{tikzpicture}
        
        \begin{groupplot}[group style={group size=1 by 2, vertical sep=0.5cm}, ]
        \nextgroupplot[
        yscale=0.55,
        legend cell align={left},
        legend columns=1,
        legend style={
          fill opacity=0.8,
          font=\small,
          draw opacity=1,
          text opacity=1,
          at={(0.75,1.1)},
          anchor=south,
          draw=white!80!black
        },
        scaled x ticks=manual:{}{\pgfmathparse{#1}},
        tick align=outside,
        tick pos=left,
        x grid style={white!69.0196078431373!black},
        xmin=-1.45, xmax=30.45,
        xtick style={color=black},
        xticklabels={},
        y grid style={white!69.0196078431373!black},
        ylabel={\(\displaystyle \Re({\ph V}_i)\) [p.u.]},
        ymin=0.93, ymax=1.08,
        ytick style={color=black},
        xmajorgrids=true,
        ymajorgrids=true,
        grid style=dashed
        ]
        \addplot [red, mark=square*, mark options={scale=0.5}, mark repeat=2,mark phase=1]
        table [x expr=\coordindex, y={predictionsRE}]{\CSVPredVSObservedZeroExcludedEdgesVarNodes};
        \addlegendentry{Predictions}
        
        \addplot [blue, mark=square*, mark options={scale=0.5}, mark repeat=2,mark phase=2]
        table [x expr=\coordindex, y={true_valuesRE}]{\CSVPredVSObservedZeroExcludedEdgesVarNodes};
        \addlegendentry{Labels}
        
        
        
        
        \nextgroupplot[
        yscale=0.55,
        tick align=outside,
        tick pos=left,
        x grid style={white!69.0196078431373!black},
        xlabel={Bus index ($i$)},
        xlabel style={yshift=-1pt},
        xmin=-1.45, xmax=30.45,
        xtick style={color=black},
        y grid style={white!69.0196078431373!black},
        ylabel={\(\displaystyle \Im({\ph V}_i)\) [p.u.]},
        ymin=-0.35, ymax=0.05,
        ytick style={color=black},
        xmajorgrids=true,
        ymajorgrids=true,
        grid style=dashed
        ]
        
        \addplot [red, mark=square*, mark options={scale=0.5}, mark repeat=2,mark phase=1]
        table [x expr=\coordindex, y={predictionsIM}]{\CSVPredVSObservedZeroExcludedEdgesVarNodes};
        
        \addplot [blue, mark=triangle*, mark options={scale=0.9}, mark repeat=2,mark phase=2]
        table [x expr=\coordindex, y={true_valuesIM}]{\CSVPredVSObservedZeroExcludedEdgesVarNodes};
        
                
        
        
        \end{groupplot}
        
        \end{tikzpicture}
    \caption{GNN predictions and labels for one test example with optimally placed PMUs.}
    \label{PredictionsPerNode0Excluded}
\end{figure}
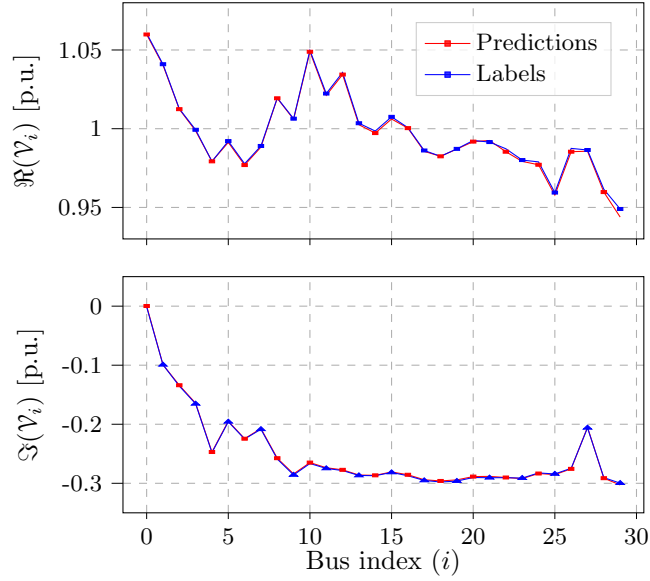

In Fig.~\ref{PredictionsPerNode0Excluded}, we present the predictions and labels for each of the variable nodes for one of the samples from the test set created with measurement variance $10^{-5}$. The results for the real and imaginary parts of the complex node voltages (shown in the upper and lower plots, respectively) indicate that GNNs can be used as accurate SE approximations.

\subsubsection{Performance in a Partially Observable Scenario} \label{subsecPartiallyObservable}
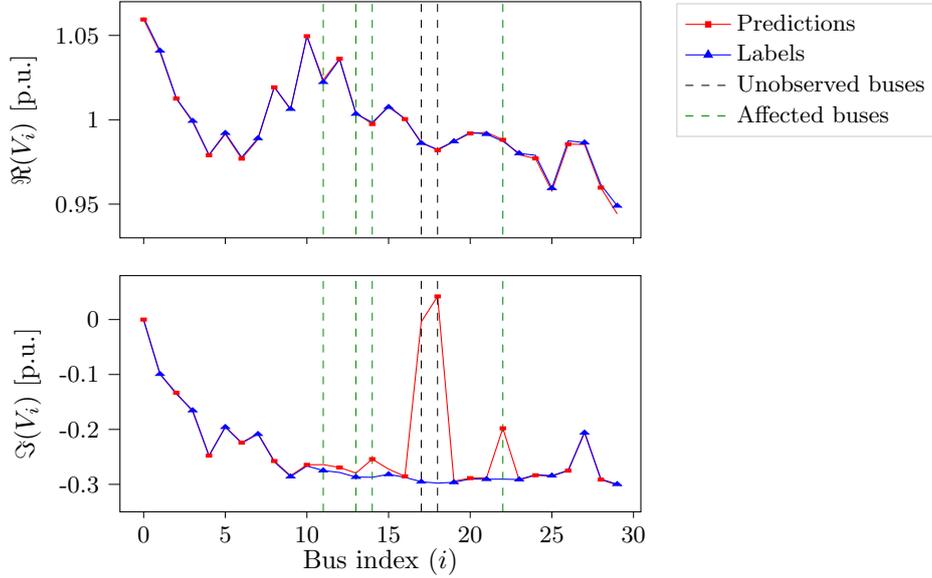
\begin{figure}[htbp]
    \centering
    \pgfplotstableread[col sep = comma,]{chapter_04/dataLinear/PredictionsPerNode2PMUExcluded.csv}\CSVPredVSObservdTwoPMUExcludedEdgesVarNodes
    
        \begin{tikzpicture}
        
        \begin{groupplot}[group style={group size=1 by 2, vertical sep=0.5cm}, ]
        \nextgroupplot[
        yscale=0.55,
        legend cell align={left},
        legend columns=1,
        legend style={
          fill opacity=1,
          font=\small,
          draw opacity=1,
          text opacity=1,
          at={(1.32,17.9)},
          anchor=south,
          draw=white!80!black,
          nodes={scale=0.98, transform shape}
        },
        scaled x ticks=manual:{}{\pgfmathparse{#1}},
        tick align=outside,
        tick pos=left,
        x grid style={white!69.0196078431373!black},
        xmin=-1.45, xmax=30.45,
        xtick style={color=black},
        xticklabels={},
        y grid style={white!69.0196078431373!black},
        ylabel={\(\displaystyle \Re({V}_i)\) [p.u.]},
        ymin=0.93, ymax=1.07,
        ytick style={color=black}
        ]
        \addplot [red, mark=square*, mark options={scale=0.5}, mark repeat=2,mark phase=1]
        table [x expr=\coordindex, y={predictionsRE}]{\CSVPredVSObservdTwoPMUExcludedEdgesVarNodes};
        \addlegendentry{Predictions}
        
        \addplot [blue, mark=triangle*, mark options={scale=0.9}, mark repeat=2,mark phase=2]
        table [x expr=\coordindex, y={true_valuesRE}]{\CSVPredVSObservdTwoPMUExcludedEdgesVarNodes};
        \addlegendentry{Labels}

        

        \addplot [very thin, black, dashed]
        table {%
        17 -2.0
        17 2.0
        };
        \addlegendentry{Unobserved buses}

        \addplot [very thin, green!55!black, dashed]
        table {%
        14 -2
        14 2
        };
        \addlegendentry{Affected buses}

        \addplot [very thin, green!55!black, dashed, forget plot]
        table {%
        11 -2.0
        11 2.0
        };
        \addplot [very thin, green!55!black, dashed, forget plot]
        table {%
        13 -2.0
        13 2.0
        };
        \addplot [very thin, green!55!black, dashed, forget plot]
        table {%
        22 -2.0
        22 2.0
        };

        \addplot [very thin, black, dashed, forget plot]
        table {%
        18 -2.0
        18 2.0
        };

        \nextgroupplot[
        yscale=0.55,
        legend cell align={left},
        legend columns=1,
        legend style={
          fill opacity=0.8,
          font=\small,
          draw opacity=1,
          text opacity=1,
          at={(0.785,4.249)},
          anchor=south,
          draw=white!80!black,
          nodes={scale=0.98, transform shape}
        },
        tick align=outside,
        tick pos=left,
        x grid style={white!69.0196078431373!black},
        xlabel={Bus index ($i$)},
        xlabel style={yshift=-1pt},
        xmin=-1.45, xmax=30.45,
        xtick style={color=black},
        y grid style={white!69.0196078431373!black},
        ylabel={\(\displaystyle \Im({V}_i)\) [p.u.]},
        ymin=-0.35, ymax=0.08,
        ytick style={color=black}
        ]
        
        \addplot [red, mark=square*, mark options={scale=0.5}, mark repeat=2,mark phase=1, forget plot]
        table [x expr=\coordindex, y={predictionsIM}]{\CSVPredVSObservdTwoPMUExcludedEdgesVarNodes};
        
        \addplot [blue, mark=triangle*, mark options={scale=0.9}, mark repeat=2,mark phase=2, forget plot]
        table [x expr=\coordindex, y={true_valuesIM}]{\CSVPredVSObservdTwoPMUExcludedEdgesVarNodes};
        
        
        
        \addplot [very thin, green!55!black, dashed, forget plot]
        table {%
        14 -2
        14 2
        };
        
        \addplot [very thin, green!55!black, dashed, forget plot]
        table {%
        11 -2.0
        11 2.0
        };
        \addplot [very thin, green!55!black, dashed, forget plot]
        table {%
        13 -2.0
        13 2.0
        };
        \addplot [very thin, green!55!black, dashed, forget plot]
        table {%
        22 -2.0
        22 2.0
        };
        
        \addplot [very thin, black, dashed, forget plot]
        table {%
        17 -2.0
        17 2.0
        };
        
        \addplot [very thin, black, dashed, forget plot]
        table {%
        18 -2.0
        18 2.0
        };
        
        \end{groupplot}
        
        \end{tikzpicture}

    \caption{GNN predictions and labels for one test example with phasors from two neighbouring PMUs removed. Vertical black lines indicate unobserved buses, while green lines represent buses that are affected by the loss of measurement data.}
    \label{PredictionsPerNode2PMUExcluded}
\end{figure}

To further assess the robustness of the proposed model, we test it by excluding several measurement phasors from the previously used test samples with optimally placed PMUs, resulting in an underdetermined system of equations that describes the SE problem. These scenarios are relevant even at higher levels of system redundancy, where partial grid observability can occur due to multiple component (PMU and communication link) failures caused by natural disasters or cyberattacks. To exclude a measurement phasor from the test sample, we remove its real and imaginary parts from the input data, which is equivalent to removing two factor nodes from the augmented factor graph. We use the previously used 100-sample test set to create a new test set by removing selected measurement phasors from each sample while preserving the same labels obtained as SE solutions of the system with all the measurements present. As an example, we consider a scenario where two neighbouring PMUs fail to deliver measurements to the state estimator. In this case, all eight measurement phasors associated with the removed PMUs are excluded from the GNN inputs. The average MSE for the test set of 100 samples created by removing these measurements from the original test set used in this section is $3.45\times 10^{-3}$. The predictions and labels for a single test set sample, per variable node index, are shown in Figure \ref{PredictionsPerNode2PMUExcluded}. The figure includes vertical black dashed lines that indicate the indices of unobserved buses 17 and 18. These buses have higher prediction errors due to the lack of input measurement data. Neighbouring buses that are not unobserved, but are affected by measurement loss, are indicated with vertical green lines and have lower prediction errors. It can be observed that significant deviations from the labels occur for some of the neighbouring buses, while the GNN predictions are a decent fit for the remaining node labels. This demonstrates the proposed model's ability to sustain error in the neighbourhood of the malfunctioning PMU, as well as its robustness in scenarios that cannot be solved using standard WLS approaches. A possible explanation for the higher susceptibility to errors in the imaginary parts of the voltage is related to their variance in the training set. The variance of real parts of voltages is $6.6 \times 10^{-4}$, while the variance of imaginary parts of voltages is $4.8 \times 10^{-3}$. This indicates that the imaginary parts of voltages have a higher variability and may therefore be more prone to errors in the prediction model.

\subsubsection{Comparison With the Feed-Forward Deep Neural Network-Based State Estimation} \label{subsec:compareWithDNN}

The main goal of this subsection is to compare the performance of the proposed GNN-based SE approach with a state-of-the-art deep learning-based approach on a variety of power systems. We used a 6-layer feed-forward DNN model, proposed by \cite{zhang2019}, with the same number of neurons in each layer as the number of input measurement scalars. This DNN architecture has similar performance as the best architecture proposed in the same work obtained by unrolling the iterative nonlinear SE solver, which cannot be applied directly to the linear SE problem we are considering.

We tested both approaches on the IEEE 30-bus, IEEE 118-bus, IEEE 300-bus, and the ACTIVSg 2000-bus power systems \cite{ACTIVSg2000}, with measurement variances set to $10^{-5}$. In contrast to previous examples, we used maximal measurement redundancies, ranging from $3.73$ to $4.21$. We provide a comparison of the number of trainable parameters for both GNN and DNN models for various power system sizes, which is often left out in similar analyses. To compare sample efficiencies between the GNN and DNN approaches, separate models for each of the test power systems were trained using smaller and larger datasets, containing 10 and 10000 samples. The results for all test power systems are presented in Table \ref{tbl_GNN_vs_DNN}. The first four rows show the 100-sample test set MSE for GNN and DNN models trained using smaller and larger datasets. The last two rows of the table show the number of trainable parameters for both approaches, depending on the power system size.

\begin{table}[]
\caption{A comparison of the performance of GNN and DNN models trained on different training set sizes, as measured by test set MSE and the number of trainable parameters.}
\begin{center}
\resizebox{13.0cm}{!}{
\begin{tabular}{|c|c|c|c|}
\hline
\textbf{}  & \textbf{IEEE 30}                                          &  \textbf{IEEE 300} & \textbf{ACTIVSg 2000} \\ \hline
Small training set GNN    & \boldmath{$4.73 \times 10^{-6}$} & \boldmath{$5.94\times 10^{-5}$} & \boldmath{$5.08\times 10^{-4}$} \\ \hline
Small training set DNN  & $9.29\times 10^{-4}$ & $5.92\times 10^{-3}$ & $4.77\times 10^{-3}$ \\ \cline{1-4}
Large training set GNN  & \boldmath{$2.48\times 10^{-6}$} & \boldmath{$6.62\times 10^{-6}$} & \boldmath{$3.91\times 10^{-4}$} \\ \hline
Large training set DNN & $6.28\times 10^{-6}$ & $2.91\times 10^{-3}$ & $2.61\times 10^{-3}$ \\ \cline{1-4}

GNN parameters & \boldmath{$4.99\times 10^{4}$} & \boldmath{$4.99\times 10^{4}$} & \boldmath{$4.99\times 10^{4}$}   \\ \hline
DNN parameters & $3.16\times 10^{5}$ & $3.15\times 10^{7}$ & {$1.77\times 10^{9}$}   \\ \hline
\end{tabular}
\label{tbl_GNN_vs_DNN}
}
\end{center}
\end{table}

The results show that the GNN approaches result in higher overall accuracy compared to the corresponding DNN approaches for all the power system and training set sizes. Furthermore, the number of trainable parameters (i.e., the model size) is constant\footnote{More precisely, the number of trainable parameters in the proposed GNN model remains nearly constant as the number of buses in the power system increases. This effect would only be noticeable for larger power systems. The only exception is the number of input neurons for the binary index encoding of the variable nodes, which grows logarithmically with the number of variable nodes. However, this increase is insignificant compared to the total number of GNN parameters.} and relatively low for GNN models, because the number of neurons in a layer is constant regardless of power system size. In contrast, the number of parameters grows quadratically for DNN models, because the number of neurons in a layer grows linearly with the input size, resulting in quadratic growth of the trainable parameter matrices. When expressed in computer memory units, the GNN models we used had a significantly smaller memory footprint, taking up only 0.19 MB. In comparison, the DNN model used for the ACTIVSg 2000 power system required a much larger 6.58 GB of memory, resulting in more challenging training and inference processes. The high number of trainable parameters required by DNN models increases their storage requirements, increases the dimensionality of the training process, and directly affects the inference speed and computational complexity. Since the proposed GNNs have a linear computational complexity in the prediction process, one training iteration of GNN also has a linear computational complexity. In contrast, one training iteration of DNN-based SE would have at least quadratic computational complexity per training iteration, making the overall training process significantly slower. To recall, the reason why the GNN has a constant number of parameters and generates predictions with linear computational complexity is that it takes measurements from a limited $K$-hop neighbourhood for every node, regardless of the size of the power system.

The results indicate that the quality of GNN and DNN model predictions improves with more training data. However, compared to GNNs, DNN models performed significantly worse on smaller datasets, suggesting that they are less sample efficient and more prone to overfitting due to their larger number of hyperparameters. While we used randomly generated training sets in the experiments, narrowing the learning space by selecting training samples based on historical load consumption data could potentially result in even better performance with small datasets.

The use of GNNs for power systems analysis has several additional advantages over using DNNs. One advantage is flexibility: spatial GNNs can produce results even if the input power system topology changes, whereas conventional DNN methods are trained and tested on the same topology of the power system. For example, if some measurements are removed from the inputs (as discussed in Subsection \ref{subsecPartiallyObservable}), a DNN would require retraining from scratch with the new topology. GNNs also have some theoretical advantages over other deep learning methods in that they are permutation invariant and equivariant. This means that the output of the GNN is the same regardless of the order in which the inputs are presented, and that the GNN output changes in a predictable and consistent way when the inputs are transformed. This property is useful for problems like SE, where the order of the nodes and edges is not important and the system can undergo topological changes. In addition, GNNs incorporate topology information into the learning process by design, whereas many other deep learning methods in power systems use node-level data as inputs while ignoring connectivity information. Finally, unlike most deep learning methods, spatial GNNs can be distributed for evaluation among edge devices.

Similar conclusions can be drawn when comparing GNNs with recurrent and convolutional neural networks for similar power systems' analysis algorithms, as they also require information from the entire power system as input. Overall, this comparison highlights the potential advantages of using GNNs for power system modelling and analysis.

\subsubsection{Robustness to Outliers}

To assess how well the proposed model can handle outliers in input data, we carried out experiments on two separate test sets, each containing samples with different degrees of outlier intensity. The experiments followed the setup described in Sec. \ref{subsec:optPlacedPMUs}, with the test samples initially generated using a measurement variance of $10^{-5}$. We replaced one of the existing measurements in each test sample with a randomly generated value, using a variance of either $\sigma_1 = 1.6$ or $\sigma_2 = 1.6\cdot10^2$. WLS SE solutions without outliers in the inputs are used as ground-truth values. The results, shown in Table \ref{tbl_outliers}, indicate the performance of four different approaches, as well as the WLS SE and the approximative WLS SE algorithm on the same test sets.

The first approach, which uses the already trained model from Sec. \ref{subsec:optPlacedPMUs}, results in the highest prediction error on tests set with outliers. The primary factor contributing to the MSE, particularly in the test set with outliers generated using larger variances, are significant mismatches from the ground-truth values in the $K$-hop neighbourhood of the outlier. This occurs because the ReLU activation function does not constrain its inputs during neighbourhood aggregation. To address this problem, we propose a second approach where we train a GNN model with the same architecture as the previous one, but which uses the tanh (hyperbolic tangent) activation function instead of ReLU. As presented in Table \ref{tbl_outliers}, this approach results in a significantly lower test set MSE for outliers generated using larger variances compared to the proposed GNN with ReLU activations, WLS SE, and the approximative WLS SE. The saturation effect of tanh prevents high values stemming from outliers from propagating through the GNN, but also reduces the training quality due to the vanishing gradient problem. Specifically, all the experiments we conducted under the same conditions for GNN with tanh activations required more epochs to converge to a solution with lower prediction quality compared to the GNN with ReLU activations. As a third approach, we propose training a GNN model with ReLU activations on a dataset in which half of the samples contain outliers, which are generated in the same manner as the test samples used in this subsection. This approach turned out to be the most effective for both test cases because the GNN learns to neutralize the effect of unexpected inputs from the dataset examples while maintaining accurate predictions in the absence of outliers in the input data. To confirm the validity of this methodology, we trained the DNN introduced in the subsection \ref{subsec:compareWithDNN} using the same datasets containing outliers. DNN was able to neutralize the effect of unexpected inputs because the input power system is small, resulting in the second-best approach in terms of robustness to outliers, trailing the GNN trained on the dataset containing outliers. 

In summary, as expected, all methods produced better results for the test set containing outliers with lower variances, while the GNN trained with outliers demonstrated the best performance for both higher and lower variance outliers. We note that these are only preliminary efforts to make the GNN model robust to outliers, and that future research could combine ideas from standard bad data processing methods in SE with the proposed GNN approach.

\begin{table}[htbp]
\caption{  A comparison of the results of various approaches for two test sets with different degrees of outlier intensity.}
\begin{center}
    \begin{tabular}{ | c | c | c | }
        \hline
        \textbf{Approach} & \begin{tabular}[c]{@{}c@{}}\textbf{Test set MSE} \\ $\mathbf{\sigma_1 = 1.6}$\end{tabular}  & \begin{tabular}[c]{@{}c@{}}\textbf{Test set MSE} \\ $\mathbf{\sigma_2 = 1.6 \cdot 10^2}$\end{tabular} \\
        \hline
        GNN & $9.48\times 10^{-3}$ & $1.60\times 10^{3}$ \\
        \hline
        GNN with tanh & $6.87\times 10^{-3}$ & $2.39\times 10^{-2}$ \\
        \hline
        GNN trained with outliers & \boldmath{$4.44\times 10^{-6}$} & \boldmath{$7.99\times 10^{-6}$} \\
        \hline
        DNN trained with outliers & $1.06\times 10^{-5}$ & $4.21\times 10^{-5}$ \\
        \hline
        WLS SE & $1.43\times 10^{-3}$ & $1.41\times 10^{-1}$ \\
        \hline
        Approximative WLS SE & $1.39\times 10^{-3}$ & $1.35\times 10^{-1}$ \\
        \hline
    \end{tabular}
\label{tbl_outliers}
\end{center}
\end{table}

\subsection{Scalability and Sample Efficiency Analysis of Linear State Estimation}

In the previous subsection, the GNN model for linear SE demonstrated good approximation capabilities under normal operating conditions and performed well in unobservable and underdetermined scenarios. This subsection extends the previous one in the following ways:
\begin{itemize}
    \item We conduct an empirical analysis to investigate how the same GNN architecture could be used for power systems of various sizes. We use the IEEE 30-bus system, the IEEE 118-bus system, the IEEE 300-bus system, and the ACTIVSg 2000-bus system \cite{ACTIVSg2000}, with measurements placed so that measurement redundancy is maximal. Our main assumption is that the local properties of the graphs in these systems are similar, leading to local neighbourhoods with similar structures which can be represented using the same embedding space size and the same number of GNN layers.
    \item To evaluate the sample efficiency of the GNN model, we run multiple training experiments on different sizes of training sets. Additionally, we assess the scalability of the model by training it on various power system sizes and evaluating its accuracy, training convergence properties, inference time, and memory requirements. For this purpose we create training sets containing 10, 100, 1000, and 10000 samples for each of the mentioned power systems, while the GNN models are tested on sets comprising 100 samples.
    \item As a side contribution, the proposed GNN model is tested in scenarios with high measurement variances, using which we simulate phasor misalignments due to communication delays \cite{zhao2019power}. While this is usually simulated by using variance values that increase over time, as an extreme scenario we fix the measurement variances to a high value of $5 \times 10^{-1}$.
\end{itemize}

It is important to note that the conclusions that will be made apply to GNN based-nonlinear SE as well.

\subsubsection{Properties of Power System Augmented Factor Graphs }

\begin{figure}[!t]
\centering
\begin{tikzpicture}
    \begin{axis} [
        grid=major,
        height=5cm,
        width=\linewidth,
        yscale=1,
        xscale=1,
        xlabel={Number of buses},
        ylabel style={align=center},
        ylabel={Power system\\graph properties},
        xlabel style={
            yshift={-5}, 
        },
        ymin=-0.5, ymax=15,
        ytick={0, 2, 4, 6, 8, 10, 12, 14},
        minor y tick num={1},
        xtick={0, 1, 2, 3},
        xticklabels={30, 118, 300, 2000},
        legend style={
            nodes={
                scale=0.8,
            },
            font=\footnotesize,
            anchor={north west},  
            at={(0.02,0.98)},     
            cells={anchor=west}, 
        },
        legend entries={Redundancy, Avg. Degree, Avg. Path Length, Avg. Cluster Coeff.},
        cycle list={
            {red, mark=square*},
            {blue, mark=*},
            {brown, mark=triangle*},
            {black, mark=pentagon*},
        }
    ]

    \addplot+ coordinates {
        (0, 3.733)
        (1, 4.153)
        (2, 3.740)
        (3, 4.206)
    };
    \addplot+ coordinates {
        (0, 6.894)
        (1, 7.507)
        (2, 6.940)
        (3, 7.527)
    };
    \addplot+ coordinates {
        (0, 4.114)
        (1, 7.093)
        (2, 10.581)
        (3, 13.751)
    };
    \addplot+ coordinates {
        (0, 0.639)
        (1, 0.666)
        (2, 0.641)
        (3, 0.674)
    };
    \end{axis}
\end{tikzpicture}
\caption{Properties of augmented factor graphs along with the system's measurement redundancy for different test power systems, labelled with their corresponding number of buses.}
\label{PowerSystemProperties}
\end{figure}
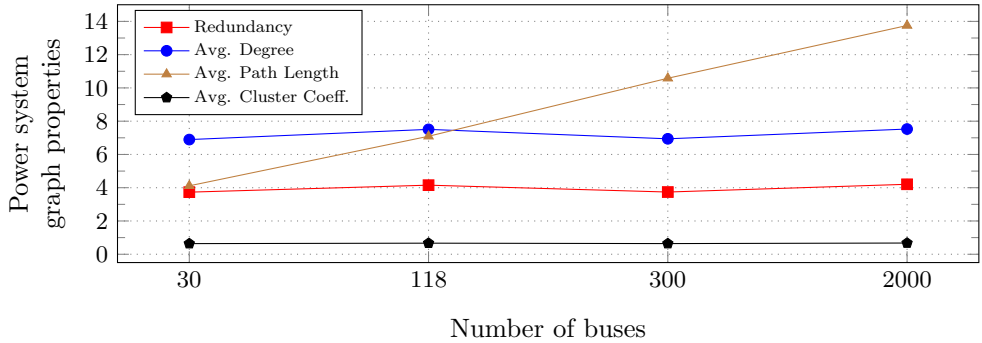

For all four test power systems, we create augmented factor graphs using the methodology described in Section~\ref{subsec:gnnBasedSe}. Fig.~\ref{PowerSystemProperties} illustrates how the properties of the augmented factor graphs, such as average node degree, average path length, average clustering coefficient, along with the system's maximal measurement redundancy, vary across different test power systems.

The average path length is a property that characterizes the global graph structure, and it tends to increase as the size of the system grows. However, as a design property of high-voltage networks, the other graph properties such as the average node degree, average clustering coefficient, as well as maximal measurement redundancy do not exhibit a clear trend of change with respect to the size of the power system. This suggests that the structures of local, $K$-hop neighbourhoods within the graph are similar across different power systems, and that they contain similar factor-to-variable node ratio. Consequently, it is reasonable to use the same GNN architecture (most importantly, the number of GNN layers and the node embedding size) for all test power systems, regardless of their size. In this way, the proposed model achieves scalability, as it applies the same set of operations to the local, $K$-hop neighbourhoods of augmented factor graphs of varying sizes without having to adapt to each individual case.

\subsubsection{Training Convergence Analysis}
First, we analyse the training process for the IEEE 30-bus system with four different sizes of the training set. As mentioned in \ref{subsec:gnnBasedSe}, the training loss is a measure of the error between the predictions and the ground-truth values for data samples used in the training process. The validation loss, on the other hand, is a measure of the error between the predictions and the ground-truth values on a separate validation set. In this analysis, we used a validation set of 100 samples.

\begin{figure}[!t]
\centering
\begin{tikzpicture}
    \begin{axis} [
        grid=major,
        yscale=1,
        xscale=1,
        height=5cm,
        width=\linewidth,
        ymin=-0.005, ymax=0.11,
        xlabel={Epoch},
        ylabel={Validation loss},
        minor y tick num={1},
        xtick={1, 30, 60, 90, 120, 150},
        yticklabels={0, 0, 0.05, 0.1},
        xticklabel style={rotate=-90},
        yticklabel style={
            /pgf/number format/.cd={
                fixed, 
                fixed zerofill,
                precision=2,
            },
        },
        scaled y ticks=false,
        unbounded coords=jump,
        legend style={
            anchor={north west},
            at={(0.2, 0.97)},
            font=\footnotesize,
            cells={anchor=west},    
            nodes={scale=0.9,},
            legend columns=2,
        },
        cycle list={
            {red, mark=square*},
            {blue, mark=*},
            {brown, mark=triangle*},
            {black, mark=pentagon*},
        },
        legend entries={$10$, $100$, $1000$, $10000$},
    ]

    \addplot+ [
        smooth,
        each nth point={30},
    ] table [
        x=n, y=validation_loss
    ]{chapter_04/dataLinear/10_sample_losses.txt};

    \addplot+ [
        smooth,
        each nth point={30},
    ] table [
        x=n, y=validation_loss
    ]{chapter_04/dataLinear/100_sample_losses.txt};

    \addplot+ [
        smooth,
        each nth point={30},
    ] table [
        x=n, y=validation_loss
    ]{chapter_04/dataLinear/1000_sample_losses.txt};

    \addplot+ [
        smooth,
        each nth point={30},
    ] table [
        x=n, y=validation_loss
    ]{chapter_04/dataLinear/10000_sample_losses.txt};
    \end{axis}
\end{tikzpicture}
\caption{Validation losses for trainings on four different training set sizes.}
\label{ieee30Convergence}
\end{figure}
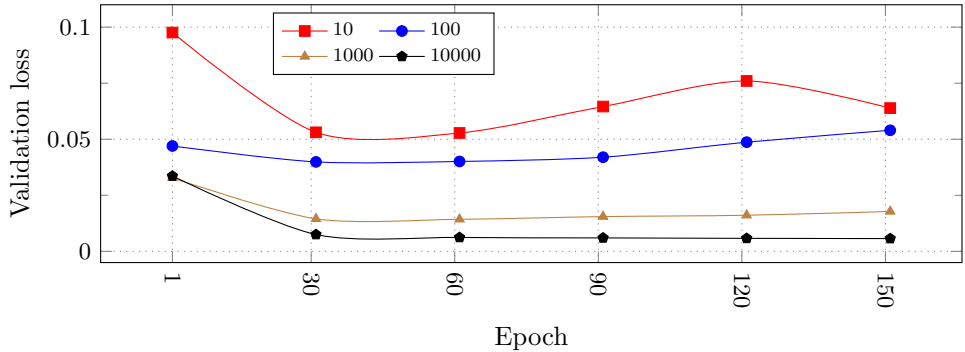

The training losses for all the training processes converged smoothly, so we do not plot them for the sake of clarity. Figure \ref{ieee30Convergence} shows the validation losses for 150 epochs of training on four different training sets. For smaller training sets, the validation loss decreases initially but then begins to increase, which is a sign of overfitting. In these cases, a common practice in machine learning is to select the model with the lowest validation loss value. As it will be shown in \ref{subsec:accuracy}, the separate test set results for models created using small training sets are still satisfactory. As the number of samples in the training set increases, the training process becomes more stable. This is because the model has more data to learn from and is therefore less prone to overfitting.

Next, in Table \ref{tbl_all_schemesEpochs}, we present the training results for the other power systems and training sets of various sizes. The numbers in the table represent the number of epochs after which either the validation loss stopped changing or began to increase. Similarly to the experiments on the IEEE 30-bus system, the trainings on smaller training sets exhibited overfitting, while others converged smoothly. For the former, the number in the table indicates the epoch at which the validation loss reached its minimum and stopped improving. For the latter, the number in the table represents the epoch when there were five consecutive validation loss changes less than $10^{-5}$.

\begin{table}[htbp]
\caption{Epoch until validation loss minimum for various power systems and training set sizes.}
\begin{center}
\resizebox{8.7cm}{!}{
    \begin{tabular}{ | c | c | c | c | }
        \hline
        \textbf{Power system}  & IEEE 118 & IEEE 300 & ACTIVSg 2000                                                       
        \\ \hline
        \textbf{10 samples}    & $61$     & $400$    & $166$                                                              
        \\ \hline
        \textbf{100 samples}   & $38$     & $84$     & $200$                                                              
        \\ \hline
        \textbf{1000 samples}  & $24$     & $82$     & $49$                                                               
        \\ \hline
        \textbf{10000 samples} & $12$     & $30$     & $15$
        \\ \hline

    \end{tabular}
\label{tbl_all_schemesEpochs}
}
\end{center}
\end{table}

Increasing the size of the training set generally results in a lower number of epochs until the validation loss reaches its minimum. However, the epochs until the validation loss reaches its minimum vary significantly between the different power systems. This could be due to differences in the complexity of the systems or the quality of the data used for training.

\subsubsection{Accuracy Assessment}
\label{subsec:accuracy}

\begin{figure*}[!ht]
    \hspace*{\fill}
    \begin{subfigure}[t]{\textwidth}
    \centering
        \begin{tikzpicture}
            \begin{customlegend}[legend entries={Approx. SE (baseline), GNN SE}, legend columns=2,]
                \addlegendimage{draw=red,mark=none,solid,line legend};
                \addlegendimage{blue,mark=*,solid};
            \end{customlegend}
        \end{tikzpicture}
    \end{subfigure}
    \hspace*{\fill}
    \newline
    \hspace*{\fill}
    \begin{subfigure}[t]{0.4\linewidth}
        \centering
        \begin{tikzpicture}
            \begin{axis} [
                title=(a) IEEE 30,
                title style={
                    anchor={north},
                    at={(0.5,-0.5)},
                    yshift=-1,
                },
                grid=major,
                yscale=1,
                xscale=1,
                width=0.9\linewidth,
                height=4cm,
                ymin=0, ymax=0.055,
                xmin=0.8, xmax=4.2,
                xtick=data,
                xlabel={Number of training samples},
                xticklabels={$10$, $10^{2}$, $10^{3}$, $10^{4}$},
                yticklabels={$0$, $0$, $0.02$, $0.04$,},
                ylabel={Test set MSE},
                minor y tick num={1},
                yticklabel style={
                    /pgf/number format/.cd={
                        fixed, 
                        fixed zerofill,
                        precision=2,
                    },
                },
                scaled y ticks=false,
            ]
            \addplot+ [
                color=blue,
            ] table [x=n, y=gnnmsse] {chapter_04/dataLinear/ieee30.txt};
        
            \addplot+ [
                color=red,
                mark=none,
                domain=0.8:4.2,
            ] {0.0386};
            \end{axis}
        \end{tikzpicture}
        \label{subf:ieee30}
    \end{subfigure}
    \hspace{0cm} 
    \begin{subfigure}[t]{0.4\linewidth}
        \centering
        \begin{tikzpicture}
            \begin{axis} [
                title=(b) IEEE 118,
                title style={
                    anchor={north},
                    at={(0.5,-0.5)},
                    yshift=-1,
                },
                grid=major,
                yscale=1,
                xscale=1,
                width=0.9\linewidth,
                height=4cm,
                ymin=0, ymax=0.055,
                xmin=0.8, xmax=4.2,
                ylabel={Test set MSE},
                xlabel={Number of training samples},
                xtick=data,
                xticklabels={$10$, $10^{2}$, $10^{3}$, $10^{4}$},
                yticklabels={$0$, $0$, $0.02$, $0.04$,},
                minor y tick num={1},
                yticklabel style={
                    /pgf/number format/.cd={
                        fixed, 
                        fixed zerofill,
                        precision=2,
                    },
                },
                scaled y ticks=false,
            ]
            \addplot+ [
                color=blue,
            ] table [x=n, y=gnnmsse] {chapter_04/dataLinear/ieee118.txt};
        
            \addplot+ [
                color=red,
                mark=none,
                domain=0.8:4.2,
            ] {0.0415};
            \end{axis}
        \end{tikzpicture}
        \label{subf:ieee118}
    \end{subfigure} 
    \hspace*{\fill}
    \newline\newline
    \hspace*{\fill}
    \begin{subfigure}[t]{0.4\linewidth}
        \centering
        \begin{tikzpicture}
            \begin{axis} [
                title=(c) IEEE 300,
                title style={
                    anchor={north},
                    at={(0.5,-0.5)},
                    yshift=-1,
                },
                grid=major,
                width=0.9\linewidth,
                height=4cm,
                yscale=1,
                xscale=1,
                width=0.9\linewidth,
                height=4cm,
                ymin=0, ymax=0.055,
                xmin=0.8, xmax=4.2,
                xlabel={Number of training samples},
                xtick={1,2,3,4},
                xticklabels={$10$, $10^{2}$, $10^{3}$, $10^{4}$},
                yticklabels={$0$, $0$, $0.02$, $0.04$,},
                ylabel={Test set MSE},
                minor y tick num={1},
                yticklabel style={
                    /pgf/number format/.cd={
                        fixed, 
                        fixed zerofill,
                        precision=2,
                    },
                },
                scaled y ticks=false,
            ]
            \addplot+ [
                color=blue,
            ] table [x=n, y=gnnmsse] {chapter_04/dataLinear/ieee300.txt};
        
            \addplot+ [
                color=red,
                mark=none,
                domain=0.8:4.2,
            ] {0.0514};
            \end{axis}
        \end{tikzpicture}
        \label{subf:ieee300}
    \end{subfigure}
    \hspace{0cm}
    \begin{subfigure}[t]{0.4\linewidth}
        \centering
        \begin{tikzpicture}
            \begin{axis} [
                title=(d) ACTIVSg 2000,
                title style={
                    anchor={north},
                    at={(0.5,-0.5)},
                    yshift=-1,
                },
                grid=major,
                yscale=1,
                xscale=1,
                width=0.9\linewidth,
                height=4cm,
                ymin=0, ymax=0.055,
                xmin=0.8, xmax=4.2,
                ylabel={Test set MSE},
                xlabel={Number of training samples},
                xtick={1, 2, 3, 4}, 
                xticklabels={$10$, $10^{2}$, $10^{3}$, $10^{4}$},
                yticklabels={$0$, $0$, $0.02$, $0.04$,},
                minor y tick num={1},
                yticklabel style={
                    /pgf/number format/.cd={
                        fixed, 
                        fixed zerofill,
                        precision=2,
                    },
                },
                scaled y ticks=false,
            ]
            \addplot+ [
                color=blue,
            ] table [x=n, y=gnnmsse] {chapter_04/dataLinear/activsg2000.txt};
        
            \addplot+ [
                color=red,
                mark=none,
                domain=0.8:4.2,
            ] {0.0379};
            \end{axis}
        \end{tikzpicture}
        \label{subf:activsg2000}
    \end{subfigure}
    \hspace*{\fill}
    \caption{Test set results for various power systems and training set sizes.}
    \label{mse_all_schemes}
\end{figure*}
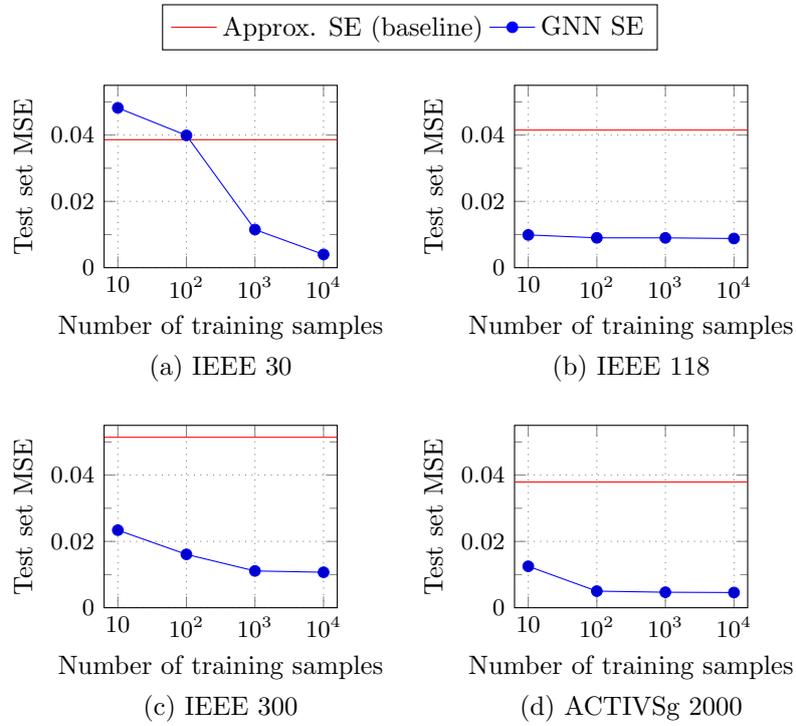

Fig.~\ref{mse_all_schemes} reports the MSEs between the predictions and the ground-truth values on 100-sample sized test sets for all trained models and the approximate WLS SE. These results indicate that even the GNN models trained on small datasets outperform the approximate WLS SE, except for the models trained on the IEEE 30-bus system with 10 and 100 samples. These results suggest that the quality of the GNN model's predictions and the generalization capabilities improve as the amount of training data increases, and the models with the best results (highlighted in bold) have significantly smaller MSEs compared to the approximate WLS SE. We assume that using carefully selected training samples based on historical load consumption data instead of randomly generated ones could potentially lead to even better results with small datasets.

\subsubsection{Inference Time}

\begin{figure}[!t]
\centering
\begin{tikzpicture}
    \begin{axis} [
        grid=major,
        ymode=log,
        xscale=0.9,
        yscale=0.5,
        xlabel={Number of buses},
        ylabel={Inference time ratio},
        xlabel style={
            yshift={-5}, 
        },
        xtick=data,
        xticklabels={30, 118, 300, 2000},
        minor y tick num={1},
        extra y ticks={2, 64},
        extra y tick labels={},
        ytick={1, 2, 10, 64, 100},
        yticklabels={1, 2, 10, 64, },
    ]

    \addplot+ [
        color=blue,
        mark=*,
    ] table [x expr=\thisrowno{0}, y expr=\thisrowno{2}/\thisrowno{3}] {chapter_04/dataLinear/inferenceTime.txt};
    \end{axis}
\end{tikzpicture}
\caption{A ratio of the execution times for WLS SE and GNN SE inference on a test set of 100 samples, as a function of the power system size.}
\label{InferTimeRatio}
\end{figure}
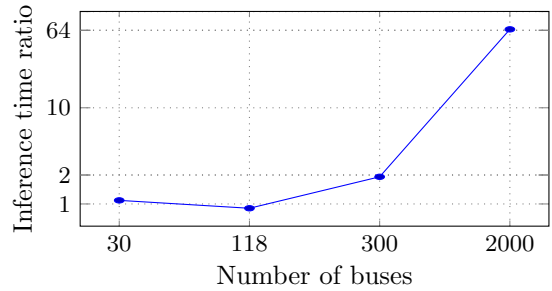

The plot in Fig.~\ref{InferTimeRatio} shows the ratio of execution times between WLS SE and GNN SE inference as a function of the number of buses in the system. These times are measured on a test set of 100 samples. The difference in computational complexity between GNN, with its linear complexity, and WLS, with more than quadratic complexity, becomes more apparent on the plot as the number of buses increases. From the results, it is clear that GNN significantly outperforms WLS in terms of inference time on larger power systems. 

Unlike with GNNs, the hardware implementation of matrix operations in WLS is a well-established field. However, the hardware implementation of GNNs is an active area of research, and it is possible that inference times may improve even further in the future \cite{Auten2020HardwareAcceleration}.

\subsection{Nonlinear State Estimation}
Finally, in this subsection, we present the numerical results of the proposed approach for the nonlinear SE problem formulation. We describe the GNN model's training process and test the trained model on various examples to validate its accuracy, and its robustness under measurement data loss due to communication failure and cyberattacks in the form of malicious data injections.

\subsubsection{Demonstration of prediction accuracy}

We conducted separate training experiments for IEEE 30 and IEEE 118-bus test cases, for which we generated a training set containing 10000 samples and validation and test sets containing 100 samples each. Similarly to the linear SE approach, each sample is created by randomly sampling the active and reactive power injections and solving the power flow problem. Measurement values are created by adding Gaussian noise to the power flow solutions, and the nonlinear SE is solved by GN to label the input measurement set in each sample. We used a Gaussian noise variance of $10^{-5}$ for phasor measurements, $10^{-3}$ for bus voltage magnitude and active and reactive power flow legacy measurements, and $10^{-1}$ for active and reactive injection legacy measurements.

\begin{figure}[ht]
	\centering
	\begin{tikzpicture}
	
	\begin{groupplot}[group style={group size=1 by 2, vertical sep=0.2cm}, ]
	\nextgroupplot[box plot width=0.6mm, ymode=log,
        	ylabel={MSE of $V_i$},
  	        grid=major,   	
  	        yscale=0.8,
  	        xmin=0, xmax=31, ymin=1e-13, ymax = 1e-2,	
            xtick={1,5,10,15,20,25,30}, 
            xticklabels={,,},
  	        ytick={1e-12, 1e-9, 1e-6, 1e-3},
  	        width=8.5cm,height=4.9cm,
  	        legend style={draw=black,fill=white,legend cell align=left,font=\tiny, at={(0.01,0.82)},anchor=west}
  	        ]
	
	        \boxplot [
                forget plot, fill=blue!20,
                box plot whisker bottom index=1,
                box plot whisker top index=5,
                box plot box bottom index=2,
                box plot box top index=4,
                box plot median index=3] {chapter_04/dataNonlinear/box_plot_magnitude_mse.txt};

	    \nextgroupplot[box plot width=0.6mm, ymode=log,
    	    xlabel={Bus Index $i$},
        	ylabel={MSE of ${\theta}_i$},
  	        grid=major, 
  	        yscale=0.8,
  	        xmin=0, xmax=31, ymin=1e-13, ymax = 1e-2,
            xtick={1,5,10,15,20,25,30}, 
  	        ytick={1e-12, 1e-9, 1e-6, 1e-3},
  	        width=8.5cm,height=4.9cm,
  	        legend style={draw=black,fill=white,legend cell align=left,font=\tiny, at={(0.01,0.82)},anchor=west}]
	
	        \boxplot [
                forget plot, fill=blue!20,
                box plot whisker bottom index=1,
                box plot whisker top index=5,
                box plot box bottom index=2,
                box plot box top index=4,
                box plot median index=3] {chapter_04/dataNonlinear/box_plot_angle_mse.txt};  
    \end{groupplot}        
	\end{tikzpicture}
	\caption{The test set MSE between the predictions and the labels per each bus for voltage magnitudes and angles in the IEEE 30-bus test case.}
	\label{plot2}
\end{figure}
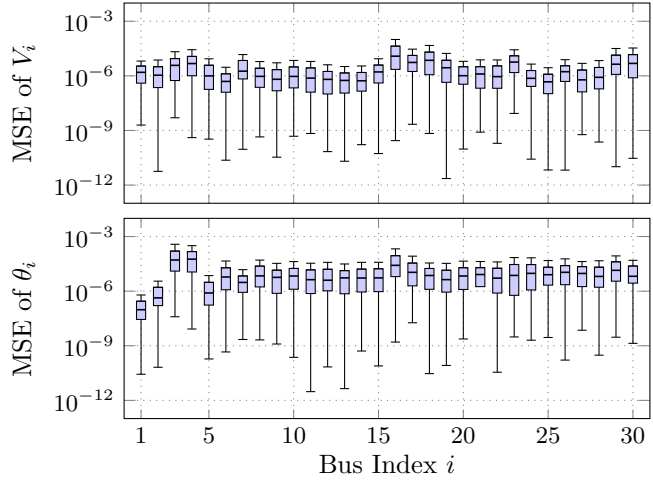

For the IEEE 30-bus test case, we placed $100$ legacy measurements and three PMUs (i.e., three bus voltage phasors and eight branch current phasors) in each sample, resulting in $2.03$ measurement redundancy. The trained model performed well on the test set, with the average test set mean square error of $1.233\times 10^{-5}$ between predictions and ground truth labels; the average test set MSE for voltage magnitudes of $5.221\times 10^{-6}$; the average test set MSE for voltage angles of $1.944\times 10^{-5}$. Fig.~\ref{plot2} shows the average test MSE per each bus, where the upper plot corresponds to voltage magnitudes and the lower one to voltage angles.

For the IEEE 118-bus test case, we placed $500$ legacy measurements and seven PMUs (i.e., seven bus voltage phasors and $26$ branch current phasors) in each sample, resulting in $2.39$ measurement redundancy. The average test set mean square error equals $2.038\times 10^{-5}$, with the average test set MSE for voltage magnitudes of $1.572\times 10^{-5}$ and the average test set MSE for voltage angles of $2.505\times 10^{-5}$. Based on the insights from both experiments, we can conclude that the proposed GNN model is a good approximator of the nonlinear SE solved by GN.

\subsubsection{Robustness to Loss of Input Data}

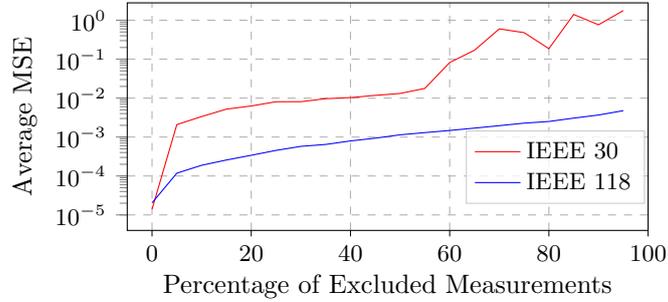
\begin{figure}[htbp]
    \centering
    \pgfplotstableread[col sep = comma,]{chapter_04/dataNonlinear/test_result_columns.csv}\testResultColumns
    \begin{tikzpicture}
    
    \begin{axis}[
    yscale=0.53,
    ymode=log,
    legend cell align={left},
    legend columns=1,
    legend style={
      fill opacity=0.8,
      font=\small,
      draw opacity=1,
      text opacity=1,
      at={(0.65,0.01)},
      anchor=north west,
      draw=white!80!black
    },
    tick align=outside,
    tick pos=left,
    x grid style={white!69.0196078431373!black},
    xlabel={Percentage of Excluded Measurements},
    xmin=-5, xmax=100, ymin=0.000004, ymax=2.8,
    xtick style={color=black},
    xlabel style={yshift=-4pt},
    y grid style={white!69.0196078431373!black},
    ylabel={Average MSE},
    ytick={0.00001, 0.0001, 0.001, 0.01, 0.1, 1},
    scaled y ticks=false,
    xmajorgrids=true,
    ymajorgrids=true,
    grid style=dashed
    ]
    \addplot [red]
    table [x={x-axis}, y={averageMSEsIEEE30}]{\testResultColumns};
    \addlegendentry{IEEE 30}
    \addplot [blue]
    table [x={x-axis}, y={averageMSEsIEEE118}]{\testResultColumns};
    \addlegendentry{IEEE 118}
    
    \end{axis}
    
    \end{tikzpicture}
    \caption{Average MSEs of test sets created by randomly excluding measurements.}
    \label{msesForBothModels}
\vspace{-4mm}    
\end{figure}

\begin{figure}[htbp]
    \centering
    \pgfplotstableread[col sep = comma,]{chapter_04/dataNonlinear/PredictionsPerNodeMeasFrom2Nodesxcluded.csv}\CSVPredVSObservdTwoPMUExcludedEdgesVarNodes
    
        \begin{tikzpicture}
        
        \begin{groupplot}[group style={group size=1 by 2, vertical sep=0.5cm}, ]
        \nextgroupplot[
        yscale=0.47,
        legend cell align={left},
        legend columns=1,
        legend style={
          fill opacity=0.8,
          font=\small,
          draw opacity=1,
          text opacity=1,
          at={(0.81,25.14)},
          anchor=south,
          draw=white!80!black,
          nodes={scale=0.98, transform shape}
        },
        scaled x ticks=manual:{}{\pgfmathparse{#1}},
        tick align=outside,
        tick pos=left,
        x grid style={white!69.0196078431373!black},
        xmin=-1.45, xmax=30.45,
        xtick style={color=black},
        xticklabels={},
        y grid style={white!69.0196078431373!black},
        ylabel={\text{Voltage Magnitude} ${V}_i$},
        ymin=0.96, ymax=1.1,
        ytick style={color=black},
        ytick={0.97, 1, 1.03, 1.06, 1.09},
        ]
        \addplot [red, mark=square*, mark options={scale=0.5}, mark repeat=2,mark phase=1]
        table [x expr=\coordindex, y={predictionsRE}]{\CSVPredVSObservdTwoPMUExcludedEdgesVarNodes};
        \addlegendentry{Predictions}
        
        \addplot [blue, mark=square*, mark options={scale=0.5}, mark repeat=2,mark phase=2]
        table [x expr=\coordindex, y={true_valuesRE}]{\CSVPredVSObservdTwoPMUExcludedEdgesVarNodes};
        \addlegendentry{Labels}
        
        
        
        \addplot [very thin, black, dashed, forget plot]
        table {%
        1 -2.0
        1 2.0
        };
        \addplot [very thin, black, dashed, forget plot]
        table {%
        0 -2.0
        0 2.0
        };
        \addplot [very thin, black, dashed, forget plot]
        table {%
        6 -2.0
        6 2.0
        };
        \addplot [very thin, black, dashed, forget plot]
        table {%
        5 -2.0
        5 2.0
        };
        \addplot [very thin, black, dashed, forget plot]
        table {%
        4 -2.0
        4 2.0
        };
        \addplot [very thin, black, dashed, forget plot]
        table {%
        3 -2.0
        3 2.0
        };

        \nextgroupplot[
        yscale=0.46,
        tick align=outside,
        tick pos=left,
        x grid style={white!69.0196078431373!black},
        xlabel={Bus Index $i$},
        xmin=-1.45, xmax=30.45,
        xtick style={color=black},
        xlabel style={yshift=-4pt},
        y grid style={white!69.0196078431373!black},
        ylabel={\text{Voltage Angle} ${\theta}_i$},
        ymin=-0.45, ymax=0.05,
        ytick style={color=black},
        ]
        
        \addplot [red, mark=square*, mark options={scale=0.5}, mark repeat=2,mark phase=1]
        table [x expr=\coordindex, y={predictionsIM}]{\CSVPredVSObservdTwoPMUExcludedEdgesVarNodes};
        
        \addplot [blue, mark=square*, mark options={scale=0.5}, mark repeat=2,mark phase=2]
        table [x expr=\coordindex, y={true_valuesIM}]{\CSVPredVSObservdTwoPMUExcludedEdgesVarNodes};
        
        
        
        \addplot [very thin, black, dashed, forget plot]
        table {%
        1 -2.0
        1 2.0
        };
        \addplot [very thin, black, dashed, forget plot]
        table {%
        0 -2.0
        0 2.0
        };
        \addplot [very thin, black, dashed, forget plot]
        table {%
        6 -2.0
        6 2.0
        };
        \addplot [very thin, black, dashed, forget plot]
        table {%
        5 -2.0
        5 2.0
        };
        \addplot [very thin, black, dashed, forget plot]
        table {%
        4 -2.0
        4 2.0
        };
        \addplot [very thin, black, dashed, forget plot]
        table {%
        3 -2.0
        3 2.0
        };
        
        \end{groupplot}
        
        \end{tikzpicture}

    \caption{GNN predictions and labels for one test example, with all measurements connected to two neighbouring buses removed. Dashed lines indicate the buses in the $1$-hop neighbourhood of the excluded measurements.}
    \label{PredictionsPerNodeMeasFrom2Nodesxcluded}
\end{figure}
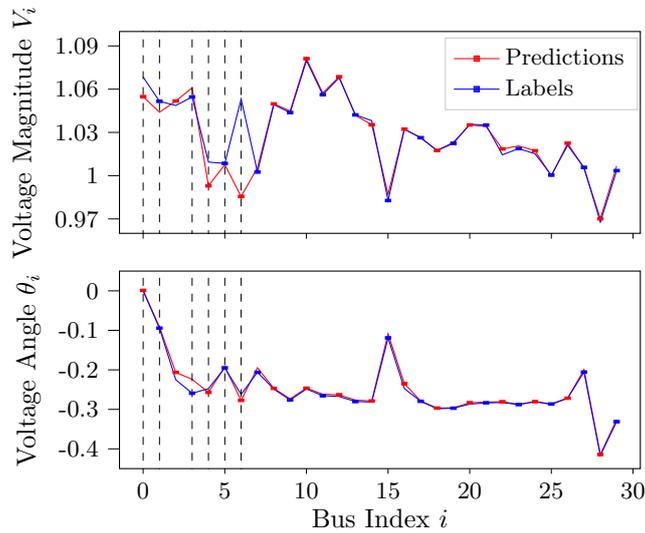

Next, we observe predictions of the GNN models previously trained on IEEE 30 and IEEE 118-bus test data when exposed to the loss of input data caused by communication failures or measurement device malfunctions. We simulate the described cases by randomly removing a percentage of all input measurements, ranging from $0\%$ to $95\%$ with a step of $5\%$. We create 20 test sets per IEEE test case, each containing samples with the same percentage of excluded measurements, and show the average test set MSEs in Fig.~\ref{msesForBothModels}. Proposed GNN models yields predictions in all examples, with an expected growing trend in MSE as the number of excluded measurements increases. In comparison, the GN method could not provide a solution for many examples due to underdetermined and ill-conditioned systems of nonlinear SE equations. A possible explanation for significantly lower MSEs for the IEEE 118-test case in these scenarios is that it contains a greater variety of subgraphs for GNN training.
To investigate the GNN predictions further, we create a test set by excluding five measurements connected to the two directly connected power system buses from each test sample, resulting in the average test set MSE of $1.488\times 10^{-4}$. Fig.~\ref{PredictionsPerNodeMeasFrom2Nodesxcluded} shows the results for one test sample, where vertical dashed lines correspond to the buses in the $1$-hop neighbourhood of the excluded measurements. We can observe that the deviation from the ground truth values manifests mainly in the vicinity of the excluded measurements, not affecting the prediction accuracy in the rest of the power system.

\vspace{5mm}

\subsubsection{Behaviour Under Malicious Data Injections}
\begin{figure}[htbp]
    \centering
    \pgfplotstableread[col sep = comma,]{chapter_04/dataNonlinear/PredictionsPerNodeWithAttacks.csv}\PredictionsPerNodeWithAttacks
    
        \begin{tikzpicture}
        
        \begin{groupplot}[group style={group size=1 by 2, vertical sep=0.5cm}, ]
        \nextgroupplot[
        yscale=0.54,
        legend cell align={left},
        legend columns=1,
        legend style={
          fill opacity=0.8,
          font=\small,
          draw opacity=1,
          text opacity=1,
          at={(0.79,0.99)},
          anchor=south,
          draw=white!80!black
        },
        scaled x ticks=manual:{}{\pgfmathparse{#1}},
        tick align=outside,
        tick pos=left,
        x grid style={white!69.0196078431373!black},
        xmin=-1.45, xmax=30.45,
        xtick style={color=black},
        xticklabels={},
        y grid style={white!69.0196078431373!black},
        ylabel={Voltage Magnitude ${V}_i$},
        ymin=0.94, ymax=1.147,
        ytick style={color=black},
        xmajorgrids=true,
        ymajorgrids=true,
        ytick={0.96, 1, 1.04, 1.08, 1.12},
        ]
        \addplot [red, mark=square*, mark options={scale=0.5}, mark repeat=3,mark phase=1]
        table [x expr=\coordindex, y={predictionsRE}]{\PredictionsPerNodeWithAttacks};
        \addlegendentry{Predictions}

        
        \addplot [black, mark=square*, mark options={scale=0.5}, mark repeat=3,mark phase=3]
        table [x expr=\coordindex, y={WLS_SE_with_attack_RE}]{\PredictionsPerNodeWithAttacks};
        \addlegendentry{GN based SE}
        
        \addplot [blue, mark=square*, mark options={scale=0.5}, mark repeat=3,mark phase=2]
        table [x expr=\coordindex, y={true_valuesRE}]{\PredictionsPerNodeWithAttacks};
        \addlegendentry{Ground truth}
        
        
        \nextgroupplot[
        yscale=0.42,
        tick align=outside,
        tick pos=left,
        x grid style={white!69.0196078431373!black},
        xlabel={Bus Index $i$},
        xlabel style={yshift=-4pt},
        xmin=-1.45, xmax=30.45,
        xtick style={color=black},
        y grid style={white!69.0196078431373!black},
        ylabel={Voltage Angle ${\theta}_i$},
        ymin=-0.44, ymax=0.04,
        ytick style={color=black},
        xmajorgrids=true,
        ymajorgrids=true,
        ytick={-0.4, -0.3, -0.2, -0.1, 0},
        ]
        
        \addplot [red, mark=square*, mark options={scale=0.5}, mark repeat=3,mark phase=1]
        table [x expr=\coordindex, y={predictionsIM}]{\PredictionsPerNodeWithAttacks};
        
        \addplot [blue, mark=square*, mark options={scale=0.5}, mark repeat=3,mark phase=2]
        table [x expr=\coordindex, y={true_valuesIM}]{\PredictionsPerNodeWithAttacks};
        
                
        \addplot [black, mark=square*, mark options={scale=0.5}, mark repeat=3,mark phase=3]
        table [x expr=\coordindex, y={WLS_SE_with_attack_IM}]{\PredictionsPerNodeWithAttacks};
        
        
        \end{groupplot}
        
        \end{tikzpicture}
    \caption{GNN predictions and GN based SE solutions for one test example with corrupted input data.}
    \label{PredictionsPerNodeCyberAttacks}
\end{figure}
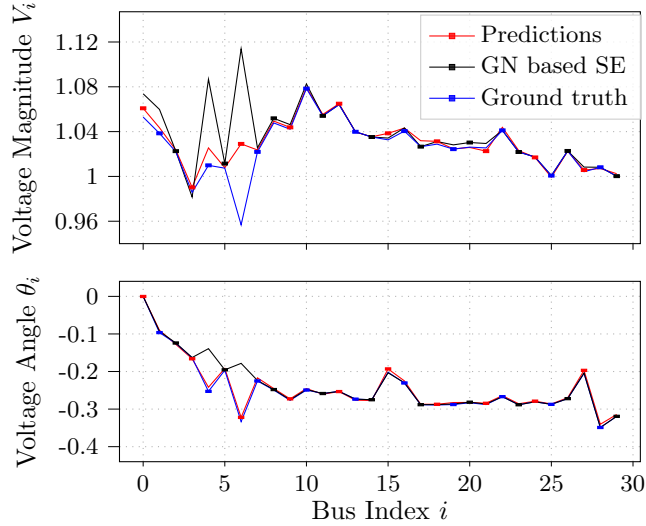

We examine the robustness of the proposed GNN model to malicious data injection type of cyberattacks by randomly altering the values of five neighbouring measurements in each test sample. We compare the proposed GNN model's predictions with the solutions of the GN method and the ground truth values obtained using the GN method applied on the uncorrupted measurement data. The GNN model demonstrated an order of magnitude better performance than the GN method, with the average test set MSEs $1.281\times 10^{-4}$ and $1.034\times 10^{-3}$, respectively. Fig.~\ref{PredictionsPerNodeCyberAttacks} depicts the comparison of the state variable predictions under corrupted input data for one example from the test set.

\section{Summary and future work}

In this chapter, we introduced methods for linear and nonlinear SE based on the GNN model specialized for operating on augmented power system factor graphs. The method avoids the problems that traditional SE solvers face, such as sensitivity to ill-conditioned cases, numerical instabilities and convergence time depending on the state variable initialization. By testing the GNN on power systems of various sizes, we observed the prediction accuracy in the normal operating states of the power system and the sensitivity when encountering false data injection cyberattacks and input data loss due to communication irregularities.

The results showed that the proposed approach provides good results for large power systems, and is an effective approximation method for traditional SE solutions even with a relatively small number of training samples, indicating its sample efficiency. The GNN model used in this approach is also fast and maintains constant memory usage, regardless of the size of the power system. More specifically, the computational complexity of the proposed GNN model regarding the number of state variables is linear during the inference phase, and it is possible to distribute the inference computation across multiple processing units. Given these characteristics, the approach is worthy of further consideration for real-world applications.

Since the proposed GNN model generates predictions even for underdetermined systems of equations describing the SE problem, it could be applied to highly unobservable distribution power systems. Another application of the proposed model for nonlinear SE could be the fast and accurate initialization of the traditional nonlinear SE solver, resulting in a hybrid approach that is both model-based and data-driven.

While our work shows promising results, an important limitation is the inability to quantify the uncertainty of the GNN predictions. However, we are encouraged by ongoing research efforts to address this issue, as quantifying uncertainty for GNN regression remains an open problem. For instance, \cite{MUNIKOTI20231} proposes a Bayesian framework that uses assumed density filtering to quantify aleatoric uncertainty and Monte Carlo dropout to capture epistemic uncertainty in GNN predictions. In light of this, we believe implementing a similar approach represents a promising future research direction.

\part{Dynamic Distribution Network Reconfiguration and Reinforcement Learning}

\chapter{Dynamic Distribution Network Reconfiguration}	\label{ch:ddnr}
\addcontentsline{lof}{chapter}{5 Dynamic Distribution Network Reconfiguration}

In this chapter, we introduce the foundations of static and dynamic distribution network reconfiguration, stating their importance in distribution management software. Furthermore, we provide a mathematical formulation of the dynamic distribution network reconfiguration problem, which will be transformed into the equivalent reinforcement learning formulation in Chapter \ref{ch:rl_ddnr}\footnote{Chapters \ref{ch:ddnr}, \ref{ch:rl}, and \ref{ch:rl_ddnr} introduce a new nomenclature.}. 

\section{Distribution Network Reconfiguration}
The electrical distribution network is the part of the electrical power system which delivers electric power from the transmission system to individual consumers. Traditional distribution networks consist of passive elements, with power flows directed only from the network supply point to the customers. Due to growing power demand, modern distribution networks exhibit changes such as energy deregulation, increased installation of the distributed generation coming mainly from renewable energy resources, and the deployment of controllable loads. These changes bring numerous challenges to the operation of modern distribution networks, such as bidirectional power flows, increased system dynamics, transient instabilities, short-circuit conditions supplied by multiple sources, overall operation inefficiency in terms of increased power losses, decreased reliability, etc.

To overcome these challenges and improve response time in unforeseen situations, modern distribution networks increase the level of network automation by using remotely controlled equipment and employing domain-specific software solutions such as DMS. DMS, which is usually tightly coupled with the SCADA system, is used for distribution network monitoring, analysis, optimization, and planning \cite{Dugan2010Distribution}. It includes functionalities like network model management and topology processing, load flow, state estimation, short circuit analysis, relay protection-related functionalities, Volt-VAR optimization, distributed energy resources monitoring and control, etc.

\begin{figure}[htbp]
	\centering
	\captionsetup[subfigure]{oneside,margin={0cm,0cm}}
	\subfloat[]{
	\centering
        \includegraphics[width=1.945in,trim={4cm 11cm 23cm 15cm},clip]{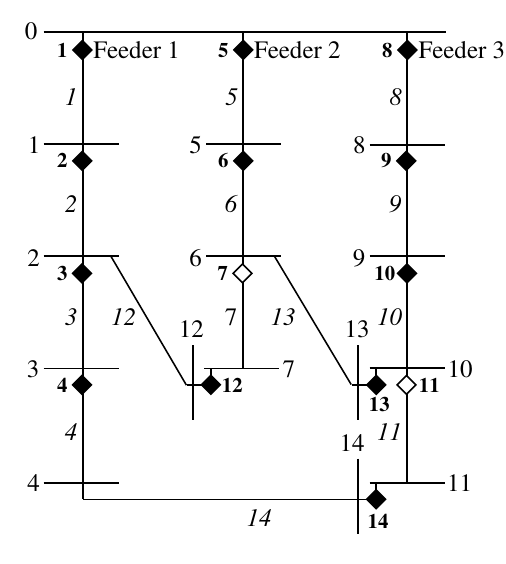}	
	}
	\subfloat[]{
	\centering
        \includegraphics[width=2.955in,trim={0 26.5cm 23cm 0},clip]{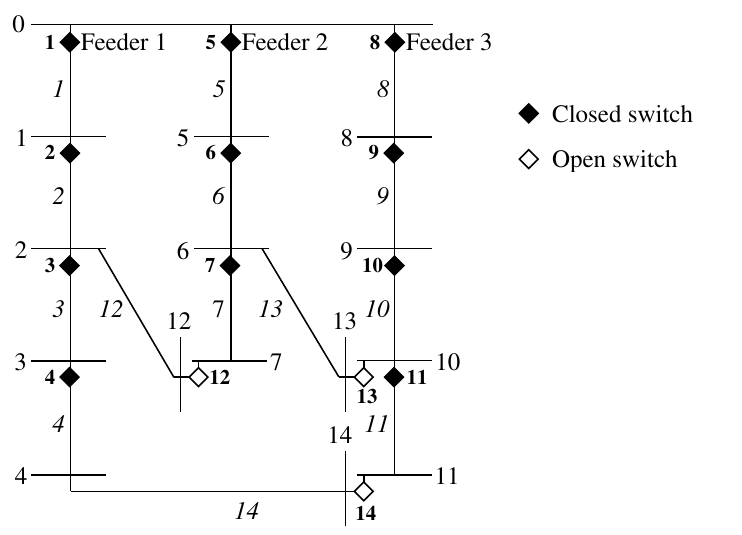}
	}
	\caption{An example of distribution network before (subfigure a) and after (subfigure b) the reconfiguration.}
	\label{beforeAndAfterReconfig}
\end{figure}

Distribution network reconfiguration is one of the most important DMS functionalities used for the optimization of distribution network operation. In general form, DNR minimizes the objective function, which usually includes power loss and voltage deviation, by changing the network topology using manipulations on the switching devices \cite{Fathi2020Reconfig}. During topology changes, it is necessary to satisfy multiple constraints, such as not exceeding the bus voltage, the apparent power of the branch, and the number of switching manipulation limits. Additional constraints related to the network topology are those that enforce the radiality of the network and ensure that all customer buses are connected to the supply point. Some of the reasons for enforcing network radiality are lower short circuit currents and simpler relay protection setup than in the distribution networks containing loops. Fig.~\ref{beforeAndAfterReconfig} represents an example of DNR on a simple 15-bus distribution network consisting of three feeders, with black squares representing closed switches and white squares representing open switches. The figure displays two distribution network topologies, before and after the DNR, both of which are radial and supply all customer buses. 

In the scope of the DMS software, DNR functionality can have secondary goals such as load balancing and Volt-VAR optimization \cite{Mishra2017ACR}, which are achieved by expanding the DNR objective function. In emergency conditions, DNR can be used to isolate the part of the distribution network where the fault occurred and restore supply to the rest of the affected customers. In these cases, since all the customers can not be supplied, DNR is used to minimize the number of the disconnected customers, or the amount of energy not supplied. However, in this thesis, we will not consider these emergency scenarios.

Static DNR is defined as DNR performed at a predefined time point, with fixed load and generation values. In traditional distribution networks with low network automation levels, where the customer load patterns change only seasonally, static DNR was usually performed a few times a year. Due to increasing load and generation dynamics, caused by controllable loads and renewable energy resources, a need for more frequent and more flexible DNR arises. Dynamic DNR optimizes the DNR objective function over the specified time interval by finding the optimal time points when DNR should be performed. DDNR is enabled by increased levels of network automation, as frequent changes to the network topology cannot be performed quickly in a manual way. In a usual formulation, the optimization interval is divided into time intervals, in each of which a network topology change can be performed. However, since switching manipulations have their own costs and can cause instabilities during the topology changes, their number is usually also a subject of minimization, in addition to being limited in the optimization problem constraints. In other words, as the switching costs can be larger than the network reconfiguration benefits, DDNR solves the trade-off problem between finding the optimal topology in each time interval and performing less frequent network topology changes to reduce the number of switching manipulations. Operation planning using DDNR can be performed on a daily or even hourly bases; therefore, it is necessary to develop fast algorithms that produce high-quality DDNR solutions.

\section{Mathematical Formulation of the DDNR Problem}

The mathematical model of DDNR, formulated as a mixed-integer nonlinear programming problem like in \cite{Lavorato2012Imposing, Kovacki2018ScalableAF} consists of a multi-objective function and constraints. The multi-objective function defined as the total cost of active energy loss and manipulation of switching devices is minimized, subject to the following constraints: active and reactive power injection constraints, bus voltage constraints, branch capacity constraints, switching operation constraints, and a network radiality constraint. Decision variables of the optimization problem consist of switch statuses in each of the time intervals, which define the distribution network topology in that time interval.

\subsection{Objective Function}
The DDNR problem is defined using the following multi-objective function:
\begin{equation} \label{eq:objFunction}
\min_{x_{sw}^t} \sum_{t=1}^{T}(C_{Loss} T_{int}^t P_{Loss}^t + C_{SWs} \, SW^t).
\end{equation}
In \eqref{eq:objFunction}, $t \in 1 \dots T$ denotes the index of a time interval, where $T$ is the total number of time intervals and $T_{int}^t$ is the duration of a time interval in hours. The decision variables of the optimization problem are the status of the switches per time interval, where the status of the switch $sw$ in the time interval $t$ is defined as follows:
\begin{equation} \label{eq:switch_status}
x_{sw}^t =
    \begin{cases}
      1, & \text{if switch $sw$ is closed in time interval $t$;}\\
      0, & \text{if switch $sw$ is opened in time interval $t$.}
    \end{cases}
\end{equation}
$C_{Loss}$ is the cost of energy losses, in $\$$ per $kWh$, $P_{Loss}^t$ denotes the active power losses in $kW$, and $C_{SWs}$ is the cost in $\$$ of the switching action for the $s^{th}$ switch. The total cost of switching actions in the time interval $t$ is calculated using the number of switches that had their status changed: 
\begin{equation} \label{eq:sum_status_changes}
SW^t = \sum_{sw=1}^{N_{sw}} y_{sw}^t,
\end{equation}
where $sw \in 1 \dots N_{sw}$ denotes the switch index and $N_{sw}$ is the total number of switches in the distribution network. $y_{sw}^t$ indicates the status change of a single switch $sw$ in time interval $t$:
\begin{equation} \label{eq:one_status_change}
y_{sw}^t =
    \begin{cases}
      1, & \text{if the switch $sw$ changed its status in time interval $t$;}\\
      0, & \text{otherwise.}
    \end{cases}
\end{equation}
The active and reactive powers of the branches in the time interval $t$ are defined as:
\begin{equation} \label{eq:P_and_Q_branch}
\begin{gathered}
P_b^t = g_{jk} (V_j^t)^2 - V_j^t V_k^t [g_{jk} cos(\theta_j^t-\theta_k^t)+b_{jk} sin(\theta_j^t-\theta_k^t)], \\
Q_b^t = -b_{jk} (V_j^t)^2+V_j^t V_k^t [b_{jk} cos(\theta_j^t-\theta_k^t)-g_{jk} sin(\theta_j^t-\theta_k^t)],
\end{gathered}
\end{equation}
where $b \in 1 \dots N_{b}$ denotes the branch index, $N_{b}$ is the total number of branches, while $j$ and $k$ denote indices of buses at the ends of the branch $b$. $V_j^t$ and $\theta_j^t$ denote the voltage magnitude and the phase angle at the bus $j$ in the time interval $t$, while $g_{jk}$ and $b_{jk}$ represent elements of the nodal conductance and susceptance matrices, respectively.
Active power losses in the time interval $t$ are defined as follows:
\begin{equation} \label{eq:Ploss}
P_{Loss}^t = \sum_{b=1}^{N_{b}} \alpha_b^t R_b \frac{(P_b^t)^2 + (Q_b^t)^2}{(V_j^t)^2},
\end{equation}
where $R_b$ denotes the resistance of the branch, and $\alpha_b^t$ combines information about the switch status and the existence of a switch on a branch:
\begin{equation} \label{eq:alphaa}
\alpha_b^t =
    \begin{cases}
      1, & \text{if branch $b$ does not have a switch;}\\
      x_{sw}^t, & \text{if branch $b$ has the switch with index $sw$.}
    \end{cases}
\end{equation}

\subsection{Constraints}

The formulation of the DDNR problem considers multiple constraints listed below.
\begin{itemize}
\item Active and reactive injection constraints, defined by the bus power balances per time interval, as in the classical load flow model \cite{Shirmohammadi1988LF}:
\begin{equation} \label{eq:P_Q_injection_constraints}
    \begin{gathered}
    P_j^t = V_j^t \sum_{k=1}^{N_n} V_k^t [g_{jk} cos(\theta_j^t-\theta_k^t)+b_{jk} sin(\theta_j^t-\theta_k^t)],\\
    Q_j^t =V_j^t \sum_{k=1}^{N_n} V_k^t [g_{jk} sin(\theta_j^t-\theta_k^t)-b_{jk} cos(\theta_j^t-\theta_k^t)],\\
    j = 1, \dots , N_n,
    \end{gathered}
\end{equation}
where $N_n$ represents the number of buses in the network. Active and reactive power injections are equal to the difference between the corresponding load and generation in bus $j$, and they are assumed constant during one time interval $t$.

\item Slack bus constraints, which specify the voltage magnitude and the phase angle in the root bus (i.e., the supply point) of the distribution network:
\begin{equation} \label{eq:slackBusConstraints}
    \begin{gathered}
    V_0^{t} = V_{spec}^{t}, \\
    \theta_0^{t} = 0,
    \end{gathered}
\end{equation}
where $V_{spec}^{t}$ represents the specified slack bus voltage magnitude value at time interval $t$. The slack bus provides an angular reference for all other buses and balances the system's active and reactive power \cite{Shirmohammadi1988LF}.

\item Bus voltage constraints:
\begin{equation} \label{eq:busVoltage_constraints}
    \begin{gathered}
    V_j^{min} \leq V_j^t \leq V_j^{max}, \\
    j = 1, \dots , N_n,
    \end{gathered}
\end{equation}
where $V_j^{min}$ and $V_j^{max}$ denote the minimum and maximum voltage magnitude allowed at the bus $j$.

\item Branch capacity constraints:
\begin{equation} \label{eq:branchCapacity_constraints}
    \begin{gathered}
    (P_b^t)^2 + (Q_b^t)^2 \leq (S_b^{max})^2, \\
    b = 1, \dots , N_b,
    \end{gathered}
\end{equation}
where $S_b^{max}$ represents the maximum apparent power allowed in the $b^{th}$ branch.

\item Switching operation constraints:
\begin{equation} \label{eq:switching operation_constraints}
    \begin{gathered}
    \sum_{t=1}^{T} |x_{sw}^t - x_{sw}^{t-1}| \leq N_{sw}^{max}, \\
    sw = 1, \dots , N_{sw},
    \end{gathered}
\end{equation}
where $N_{sw}^{max}$ represents the maximum number of allowed operations for the $sw^{th}$ switch during the optimization time interval, which depends on the type and the lifetime of the switch. $x_{sw}^0$ are the initial switch statuses, and they do not belong to the decision variables. Constraints containing absolute value operators can not be used directly in classical mixed-integer algorithms, but require reformulation by introducing additional variables. In the proposed RL approach, we will consider these constraints directly by adding a penalty term to the reward function, resulting in a simpler formulation of the problem.

\item Network radiality constraint, which assures there are no loops in the distribution network:
\begin{equation} \label{eq:radiality_constraints}
    \sum_{b=1}^{N_b} \alpha_b^t = N_n - 1.
\end{equation}
\end{itemize}

The DDNR problem formulated in this way is NP-hard, since it has $2^{N_{sw}T}$ possible solutions, and it can not be solved in polynomial time. In the forthcoming chapters, we will introduce RL algorithms which are trained to search the solution space based on the agent-environment interaction concept, and yield quality solutions with low computational effort during the algorithm's evaluation.

The proposed multi-objective formulation of the DDNR problem aims to minimize the total cost of active energy loss and manipulation of switching devices. However, in many practical applications the DDNR, apart from minimizing the cost of active energy loss and manipulation of switching devices, the DDNR is used for voltage deviation minimization \cite{Samman2020FastNr, Fathi2020Reconfig}, load balancing, Volt/Var optimization, supply restoration \cite{Mishra2017ACR}, the distribution network reliability maximization \cite{Kovacki2018ScalableAF}, limiting the budget \cite{Ahmadi2018Novel}, etc. Extension of the DDNR problem formulation assumes incorporating additional criteria into the objective function. That is achieved by adding the corresponding terms to the objective function while preserving the constraints \eqref{eq:P_Q_injection_constraints} - \eqref{eq:radiality_constraints}.
\chapter{Reinforcement Learning}	\label{ch:rl}
\addcontentsline{lof}{chapter}{6 Reinforcement learning}
Reinforcement learning (RL), as a machine learning technique, deals with how software agents learn to take actions in an environment through experience and exploration, with the goal of finding the optimal strategy that maximizes the long-term reward obtained \cite{Sutton1998}. In the RL framework, it is assumed that the agent interacts with the generally stochastic environment in discrete time steps. At the beginning of each time step, the agent observes the environment, that is, it receives the state variables from the environment. Based on the state variables, the agent takes an action and sends it to the environment. The environment then changes its state due to the action received, as well as due to its internal processes. After that, the environment sends the immediate reward signal for that time step and the state variables for the next time step to the agent. The goal of an RL algorithm is to find the (close to) optimal policy, i.e., the action selection that maximizes the long-term reward, while receiving feedback about its immediate performance. This chapter presents the theoretical background of RL in Sections \ref{finiteMDP} and \ref{qLearning} and the deep Q-learning algorithm in Section \ref{deepQLearning}, which is applied to DDNR in Chapter \ref{ch:rl_ddnr}.

\section{Finite Markov Decision Processes} \label{finiteMDP}
\begin{figure}[htbp]
    \centerline{\includegraphics[width=4in,trim={0 1.8cm 0 0},clip]{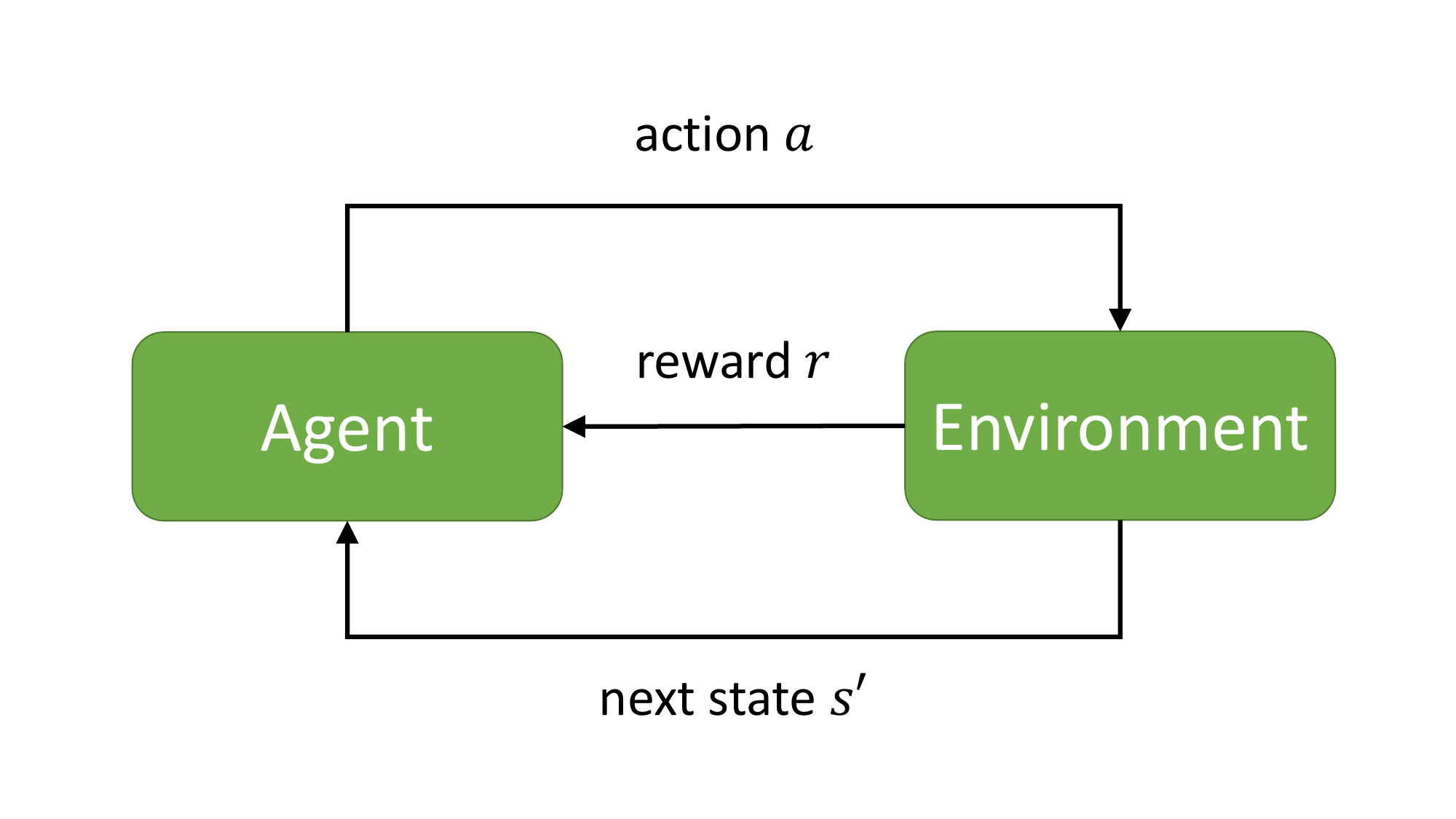}}
    \caption{The agent-environment interaction process.}
    \label{agentEnvInterraction}
\end{figure}

Finite Markov decision processes are discrete-time stochastic control processes that model decision-making in situations in which the outcome is partially stochastic and partially under the control of the decision-maker. The result of the solved MDP is the optimal sequence of actions, that is, the optimal policy. The RL problem can be mapped onto the MDP, which is defined as the following tuple:
\begin{itemize}
    \item $\mathcal{S}$ - finite set of states,
    \item $\mathcal{A}$ - finite set of actions,
    \item $\mathcal{R}$ - finite set of immediate rewards,
    \item $p\left(s'\,\lvert\, s, a\right) = \Pr\left(S_t=s'\,\lvert\, S_{t-1}=s, A_{t-1}=a\right)$ - transition probability function.
\end{itemize}
Random variables $S_t$, $S_{t-1}$, $R_t$, $A_{t-1}$ represent the new state, the previous state, received immediate reward and the action being taken, respectively, while $s', s \in \mathcal{S}$, $r \in \mathcal{R}$, $a \in \mathcal{A}$ denote the values of these random variables. The transition function represents the probability of being in the state $s'$ on the condition of being in the state $s$ previously and executing the action $a$. The agent and the environment interact in discrete time steps, as shown in Fig.~\ref{agentEnvInterraction}. In each time step the agent observes the state $s$, makes an action $a$, upon which the environment changes, and sends the feedback to the agent in the form of reward $r$, and the next state $s'$.

Additionally, a problem defined as MDP must satisfy the Markov property, meaning that the next state $s'$ depends only on the current state $s$ and the action $a$, and is independent of all previous states and actions:
\begin{equation} \label{eq:markovProperty}
    \Pr\left(S_t\,\lvert\, S_{0}, A_{0}, S_{1}, A_{1}, \dots, S_{t-1}, A_{t-1}\right) = \Pr\left(S_t\,\lvert\, S_{t-1}, A_{t-1}\right)
\end{equation}
Problems that do not satisfy the Markov property can be solved exactly using traditional MDP methods by expanding the state space with data from previous states or approximately using the RL methods.

Return in time step $t$, which represents the long-term reward starting from that time step, is the subject of optimization:
\begin{equation} \label{return}
G_{t} = R_{t+1} + \gamma R_{t+2} + \gamma^{2} R_{t+3} + \dots = \sum_{i=0}^{\infty}\gamma^{i}r_{t+i+1},
\end{equation}
where $\gamma \in [0,1]$ represents the discount factor. The case of $\gamma = 0$ corresponds to the greedy maximization of the immediate reward, while using $\gamma = 1$ implies equal weight on all rewards in the optimization horizon. Due to convergence problems in the case of long optimization horizons, the most often used value of the discount factor is $\gamma \in [0.9, 0.99]$.

Policy $\pi$ models the action selection in various states. When optimizing the policy, the long-term reward is optimized. The policy is in the general case stochastic, i.e., it maps the probability distribution of actions to states $\pi(a|s) = \Pr(A_t=a|S_t=s)$, but it can be also defined in a deterministic way $a=\pi(s)$. The quality of a policy $\pi$ in a state $s$ is usually expressed as an expectation of the discounted long-term reward $G_t$ given the state $s$, enabling the way to compare different policies and optimize them. This quantity is called the state value function, and it assigns higher values to the more desirable states in terms of the long-term reward, if following the policy $\pi$\footnote{Superscript $\pi$ denotes that the quantity is calculated with the assumption that the agent takes actions according to the policy $\pi$.}:
\begin{equation} \label{v_function}
v^{\pi}(s) = \mathop{\mathbb{E}}_{\pi}\,[G_{t} \,\lvert\, S_{t}=s] = \mathop{\mathbb{E}}_{\pi}\,[\sum_{i=0}^{\infty}\gamma^{i}R_{t+i+1} \,\lvert\, S_{t}=s].
\end{equation}

Similarly, the action value function (Q-function) is a mapping of states-action pairs to the real numbers, where the value of the state-action pair represents the expected discounted long-term reward starting from that state, taking that action, and following a concrete policy $\pi$ afterward:
\begin{equation} \label{q_function}
q^{\pi}(s, a) = \mathop{\mathbb{E}}_{\pi}\,[G_{t} \,\lvert\, S_{t}=s, A_{t}=a] = \mathop{\mathbb{E}}_{\pi}\,[\sum_{i=0}^{\infty}\gamma^{i}R_{t+i+1} \,\lvert\, S_{t}=s, A_{t}=a].
\end{equation}

The state value function has a more compact representation, while the Q-function provides a simpler way to determine the action that leads to the most desirable future state. The state value function and the Q-function can be recursively expressed using the Bellman equations:
\begin{equation} \label{eq:bellamnV}
\begin{split}
v^{\pi}(s) &= \mathop{\mathbb{E}}_{\pi}\,[R_{t+1} + \gamma v^{\pi}(S_{t+1})\,\lvert\, S_{t}=s] 
\\ &= \sum_{a\in\mathcal{A}} \pi\left(a \,\lvert\, s\right) (r + \gamma \sum_{s' \in \mathcal{S}} p(s'|s,a) v^{\pi}(s')),
\end{split}
\end{equation}
\begin{equation} \label{eq:bellamnQ}
\begin{split}
q^{\pi}(s, a) &= \mathop{\mathbb{E}}_{\pi}\,[R_{t+1} + \gamma q^{\pi}(S_{t+1}, A_{t+1})\,\lvert\, S_{t}=s] 
\\ &= r + \gamma \sum_{s'\in\mathcal{S}} p\left(s' \,\lvert\, s, a\right) \sum_{a'\in\mathcal{A}} \pi\left(a' \,\lvert\, s'\right) q^{\pi}(s', a').
\end{split}
\end{equation}

Finding the optimal state value function or the optimal Q-function results in finding the optimal policy. The optimal policy can be generated from e.g. the optimal Q-function by selecting the action with the largest Q-function value in each state. If the problem is formulated as an MDP, then at least one optimal solution exists, and an iterative procedure based on dynamic programming and the Bellman equations that converges to one of those solutions can be established. Some of the commonly used algorithms are the value iteration algorithm \cite{bellmanValueIteration}, and the policy iteration algorithm \cite{howard1960dynamic}.

To find the exact solution to the MDP, it has to be fully defined, i.e., all transition probabilities and immediate rewards have to be known, and it has to satisfy the Markov property. Additionally, for large state and action spaces, solving MDPs exactly could be computationally infeasible. Partial observability of the environment state additionally increases the computational time of the traditional algorithms that solve MDPs \cite{partially}. The main idea of RL is to overcome these problems by learning (close to) optimal policies based on the history of interactions of the agent with the environment.

\section{Q-Learning} \label{qLearning}

Our work considers model-free off-policy RL algorithms, where the optimal policy is learned directly from the accumulated experience, i.e., the history of the interaction process between the agent and the environment. On the other hand, in the model-based RL, the MDP, or the transition probabilities and immediate rewards for all state-action-next state triplets are learned from the accumulated experience and solved to obtain the policy.

Q-learning is a basic model-free RL algorithm, where the values of the Q-function for each state-action pair are stored in the Q-table and updated during algorithm training \cite{Watkins92qlearning}. During the evaluation of the algorithm, for each state the agent receives from the environment, the action with the greatest Q-function value in the table is selected. During the algorithm training, actions are selected randomly with the probability $\epsilon$, and the actions with the largest Q-function values for the corresponding states are selected with the probability $1-\epsilon$, where $\epsilon \in [0,1]$ is the exploration hyperparameter. This way the agent searches the state-action space and avoids the local optima problem. Algorithm training is performed by repeating the predefined number of episodes, which consist of time steps. One time step contains information about one interaction of the agent with the environment: the current state, the action selected by the agent, the received reward, and the next state. The length of episodes, i.e., the number of time steps in them, is generally variable.

Upon one interaction of the agent with the environment, the values in the Q-table are updated using the following rule, obtained using the idea from the Bellman equations:
\begin{equation} \label{eq:q_learning_update}
q(s, a) := (1-\alpha)q(s, a) + \alpha[r + \gamma \; \underset{a'}{\max} \; q(s', a')],
\end{equation}
where $\alpha$ is the learning rate hyperparameter. As well as MDPs, the Q-learning algorithm assumes discrete state and action spaces and for large state and action spaces it may be infeasible to learn the Q-function value for all state-action pairs.

\section{Deep Q-learning} \label{deepQLearning}

\begin{figure}[htbp]
    \centerline{\includegraphics[width=4in,trim={1cm 6.2cm 1cm 3cm},clip]{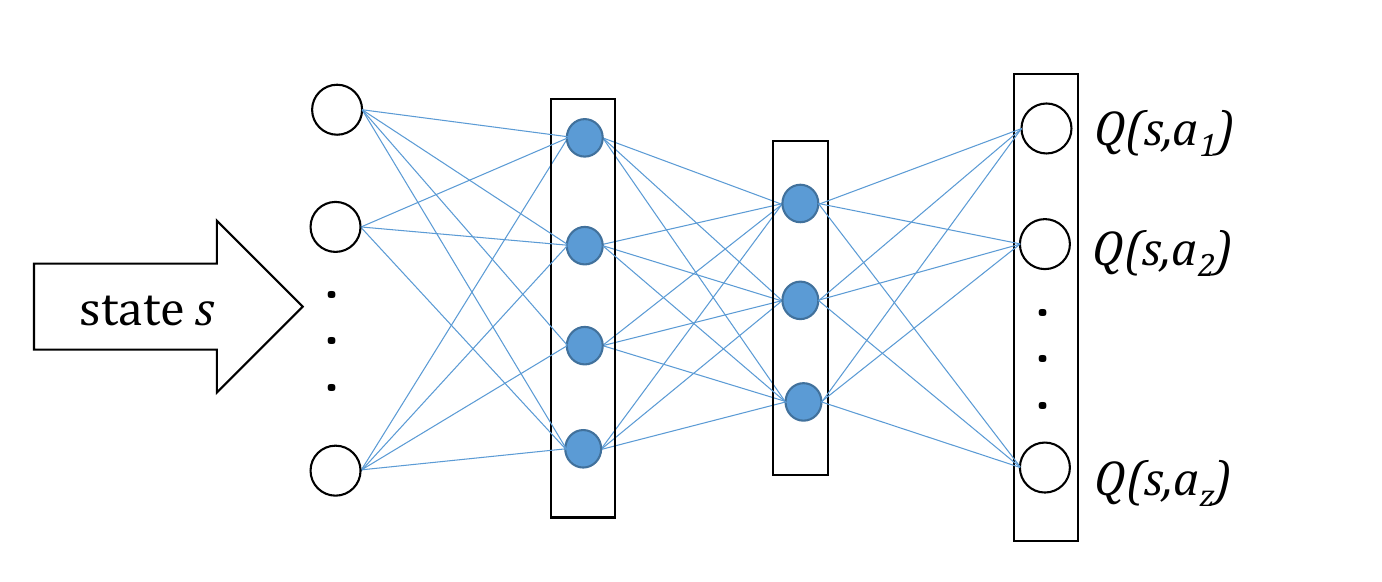}}
    \caption{An example of a deep Q-network.}
    \label{dqnTopology}
\end{figure}

The deep Q-Learning algorithm is one of the basic DRL algorithms, that utilizes the advances in the deep learning field to improve the traditional RL algorithms. The idea of the algorithm is to use a deep neural network, also called the deep Q-network (DQN), as an approximator of the Q-function \cite{mnih2015humanlevel}. Inputs to the DQN are state variables, while output neurons provide the approximation of Q-function values for each of the actions and for the input state. Therefore, the state variables can be continuous, which makes the learning feasible for large continuous state spaces which would have to be discretized when using the Q-learning algorithm. In deep Q-learning, the action space must be discrete and finite, since the number of output neurons is limited. An example of a DQN is shown in Fig.~\ref{dqnTopology}, where $Q(s,a_i)$ denotes the DQN output when the agent takes action $a_i, i=1, \dots, z$, while being in the state $s$, where $z = |\mathcal{A}|$ denotes the number of possible actions.

Deep Q-learning algorithm introduces target DQN $Q_{target}$ which has the same model architecture as DQN and in which the parameters of DQN are copied at the predefined period during the training process. Target DQN is used for determining the labels for the DQN training, as defined in \eqref{eq:dqn_labels}, which significantly improves the training process stability \cite{mnih2015humanlevel}. It reduces the oscillations of the training by fixing the label generation process during multiple training steps, allowing the DQN network weights to be updated steadily. Target networks are a widely used technique in RL, and recent RL algorithms propose more advanced variations, such as continuously updating the time-delayed target network \cite{Lillicrap2015ContinuousCW}.

On-policy RL algorithms \cite{Sutton1998} are trained by updating the same policy using which the sequence of actions is generated, which results in unstable and sample inefficient training due to correlation between the actions in the sequence. Deep Q-learning is an example of an off-policy RL algorithm, which stores the history of the agent’s interaction with the environment in the experience replay memory \cite{Fedus2020RevisitingFO}, and samples data randomly from it to perform the DQN training in a supervised learning manner. At each time step, tuple $(s,a,r,s')$ is stored in the replay memory, from which i.i.d. mini-batch data for DQN training is sampled.

Labels for DQN training are calculated in the following way, using the idea from the Bellman equations, similarly to the Q-learning algorithm:
\begin{equation} \label{eq:dqn_labels}
Q_{label}(s, a) = r + \gamma \; \underset{a'}{\max} \; Q_{target}(s', a' | \theta^{Q_{target}}).
\end{equation}
$\theta^{Q}$ and $\theta^{Q_{target}}$ denote the parameters (weights and biases) of the DQN and the target DQN. DQN is trained using the mini-batch gradient descent algorithm \cite{li2014MiniBatchGradDesc}, which minimizes the squared error loss function that expresses the distance between the labels and the DQN output during training:
\begin{equation} \label{eq:lossFunction}
L(\theta^{Q}) = \frac{1}{N_{mb}} \sum_{i=1}^{N_{mb}}(Q_{label}(s_i,a_i) - Q(s_i,a_i\,\lvert\, \theta^{Q}))^2.
\end{equation}
A trained DQN is evaluated by forwarding the input state $s$ through the network layers, obtaining the Q-function approximates $Q(s,a_i)$ for all actions $a_i, i=1, \dots, |\mathcal{A}|$, and selecting the action with the largest Q-function value.
\chapter{Reinforcement Learning based Dynamic Distribution Network Reconfiguration}	\label{ch:rl_ddnr}
\addcontentsline{lof}{chapter}{7 Reinforcement Learning based Dynamic Distribution Network Reconfiguration}
This chapter describes the way DDNR is expressed as an RL problem, how the objective function and constraints are considered, and the training and evaluation algorithms of the proposed DQN-based method. Finally, we evaluate the performance of the proposed approach on three distribution networks: 15-bus test benchmark, real-life large-scale distribution network, and the IEEE 33-bus network.

\section{Modelling Dynamic Distribution Network Reconfiguration as a Markov Decision Process} \label{modelingDDNRasMDP}

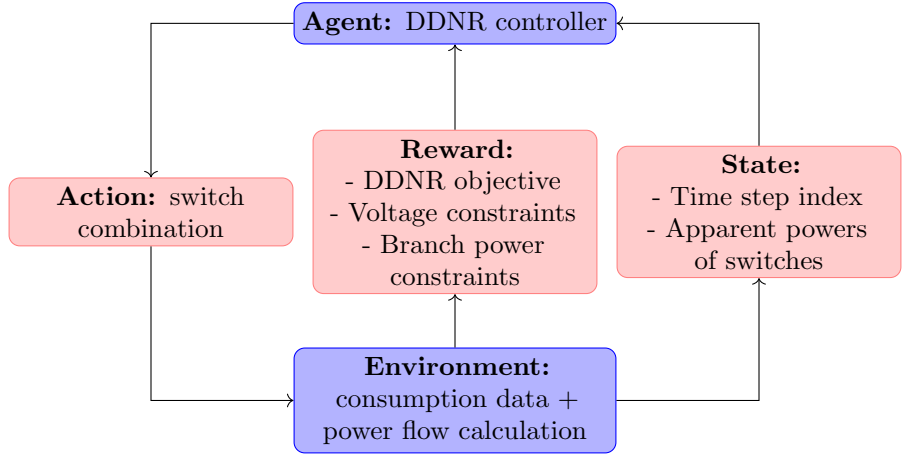
\begin{figure}[htbp]
    \begin{center}
    \begin{tikzpicture}[node distance=4cm]
\node[draw=blue, fill=blue!30, text width=4cm, align=center, text=black, rounded corners] (b1) at (4,2.5) {\textbf{Agent:} DDNR controller};
\node[draw=blue, fill=blue!30, text width=4cm, align=center, text=black, rounded corners] (b2) at (4,-2.5) {\textbf{Environment:} \\ consumption data + \\ power flow calculation };
\node[draw=red!50, fill=red!20, text width=3.5cm, align=center, text=black, rounded corners] (mb1) at (0,0) {\textbf{Action:} switch combination};
\node[draw=red!50, fill=red!20, text width=3.5cm, align=center, text=black, rounded corners] (mb2) at (4,0) {\textbf{Reward:} \\ - DDNR objective \\ - Voltage constraints \\ - Branch power constraints};
\node[draw=red!50, fill=red!20, text width=3.5cm, align=center, text=black, rounded corners] (mb3) at (8,0) {\textbf{State:} \\ - Time step index \\- Apparent powers \\ of switches};

\draw[->] (b1.west) -- (0,2.5) -- (mb1);
\draw[->] (mb1) -- (0,-2.5) -- (b2.west);
\draw[->] (b2) -- (mb2);
\draw[->] (mb2) -- (b1);
\draw[->] (b2.east) -- (8,-2.5) -- (mb3);
\draw[->] (mb3) -- (8,2.5) -- (b1.east);
    \end{tikzpicture}
    \end{center}
    \caption{The agent-environment interaction process for DDNR.}
    \label{agentEnvInterDDNR}
\end{figure}

The information flow between the DDNR agent and the environment during their interaction is presented in Fig.~\ref{agentEnvInterDDNR}. Episodes consist of $T$ time steps, where each episode corresponds to a separate instance of the DDNR applied to the one-day interval, and each RL agent's time step corresponds to the beginning of one time interval in DDNR problem formulation \eqref{eq:objFunction}. At each time step, active and reactive power consumption data for the next hour are loaded. Then, the power flow calculation is executed to create the state variables, which contain the time interval index $t$ and the apparent powers of all switches in the network $S_{sw}^t$, $sw \in 1,\dots,N_{sw}$, as shown in \eqref{eq:ddnr_state_variables}:
\begin{equation} \label{eq:ddnr_state_variables}
    s_t = (t, S_{1}^t, \dots, S_{N_{sw}}^t).
\end{equation}
This choice of state variables, motivated by the fact that switch statuses, and hence the network topology can be reconstructed using the apparent powers of the switches, reduces the state space dimensionality and DQN size. The current network configuration and power flow results are compressed into a single set of variables, from which the agent can make its own representation of the environment, and use it as an input in the decision-making process.

The action space contains all the switch combinations that lead the network in a feasible radial configuration, in which all the buses are energized. These switch combinations are enumerated uniquely so that one output neuron corresponds to one feasible radial combination. This action space definition implies that the radiality network constraint stated in \eqref{eq:radiality_constraints} is always satisfied, which accelerates the learning process.

The reward value for each time interval $t$ is equal to the negative sum of the following terms:
\begin{itemize}
    \item Price of active energy losses $C_{Loss} T_{int}^t P_{Loss}^t$, used to model the first term in the DDNR objective function stated in \eqref{eq:objFunction};
    \item Price of the switching manipulations needed to conduct the network from the previous configuration to the current one $C_{SWs} \, SW^t$, used to model the second term in the DDNR objective function stated in \eqref{eq:objFunction};
    \item Predefined penalty value $C_U$ if the bus voltage constraint is violated in any of the buses, used to model the bus voltage constraints in \eqref{eq:busVoltage_constraints};
    \item Predefined penalty value $C_S$ if the branch capacity constraint is violated for any of the branches, used to model the branch capacity constraints in \eqref{eq:branchCapacity_constraints}.
\end{itemize}

As an alternative to adding the predefined penalty to the reward function if the number of switch manipulations exceeds the predefined limit for any of the switches, we propose the following way to consider the switching operation constraints in \eqref{eq:switching operation_constraints}:
\begin{itemize}
    \item The subset of available actions is created at the beginning of each episode and is initially equal to the action set (i.e., all the switch combinations that lead the network into a feasible radial configuration, in which all the buses are energized);
    \item During the episode, the number of operations of each switch in that episode is updated. Prior to action selection, actions that would violate the switching operation constraints if selected are removed from the subset of available actions;
    \item In each time interval, the action with the largest Q-value is selected from the subset of available actions, instead of from the action set.
\end{itemize}
This approach improves the training efficiency compared to the approach that would penalize the exceeded number of switch operations. Since the actions that violate the switching operation constraints cannot be selected in the first place, the computational effort needed to learn Q-function values for those state-action pairs is eliminated. It is convenient to consider the switching operation constraints using this approach since only the selected action (switch combination) needs to be known, to conclude if the constraint is violated. The subset of available actions can be determined without executing the actions. Bus voltage and branch capacity constraints cannot be modelled using this approach generally, since the action must be executed and feedback from the environment is required for any conclusions about the constraint violations.
Active and reactive injection constraints in \eqref{eq:P_Q_injection_constraints}, as well as slack bus constraints in \eqref{eq:slackBusConstraints} are satisfied by the design of the power flow calculation \cite{Shirmohammadi1988LF} and, therefore, are not considered in the reward function. Values of the state variables and rewards are normalized, for the purpose of improving the numerical stability of neural network training.

\section{Training and Evaluation Algorithms}
During the algorithm training $N$ episodes are repeated, with each episode consisting of a predefined number of time steps $T$, where in each time step the interaction between the agent and the environment takes place, as described in \ref{modelingDDNRasMDP}. The variety of training scenarios is created by randomly sampling the daily load curves from some predefined distribution. The way the exploration hyperparameter $\epsilon$ is updated during the training can also be tuned. In this work, we used the linear decrease of $\epsilon$ until the $0.8N^\textrm{th}$ episode, and the constant value afterward. A detailed representation of the agent training procedure is displayed in Algorithm \ref{training_alg}.

\begin{algorithm} 
  \DontPrintSemicolon
    Initialize the deep Q-network's $Q(s\mid \theta^Q)$ parameters randomly\;
    Initialize the target deep Q-network's $Q'(s\mid \theta^{Q_{target}})$ parameters using the original network's parameters\;
    Initialize the experience replay buffer\;
    \For{$episode=1,2,\ldots,N$}{
        Sample daily load curves randomly\;
        Initialize $\epsilon$\;
        Initialize the subset of available actions to the action set\;
        Run the initial power flow calculation\;
        Send the initial state $s_1$ to the agent\;
        \For{$t=1,2,\ldots,T$}{
            $rand$ = random number between $0$ and $1$\;
            \uIf{$rand > \epsilon $}{
              Based on the current state $s_t$ select the action $a_t$ with the largest Q-value in the subset of available actions\;
            }
            \Else{
              Select random action $a_t$ from the subset of available actions\;
            }
            Update the subset of available actions\;
            Update the network configuration according to $a_t$\;
            Run the power flow calculation\;
            Collect the immediate reward $r_t$ and the next state $s_{t+1}$ data\;
            Store tuple $(s_t, a_t, r_t, s_{t+1})$ in the experience replay buffer\;
            Sample the mini-batch of tuples from the experience replay buffer\;
            Create labels for deep Q-network training using \eqref{eq:dqn_labels}\;
            Update deep Q-network parameters by minimizing the loss function given in \eqref{eq:lossFunction} \;
            Set loads for the next time interval for each bus\;
          }
        \uIf{update target network period}{
              $\theta^{Q_{target}}=\theta^{Q}$\;
            }
        Update $\epsilon$\;
    }
  \caption{Deep Q-network training}
  \label{training_alg}
\end{algorithm}

Once trained, the deep Q-network model can be evaluated multiple times by loading the saved trainable parameters, as shown in Algorithm \ref{eval_algorithm}. Note that during the algorithm evaluation target deep Q-network, experience replay buffer, neural network parameter updates, and the random exploration strategy are not used, reducing the memory storage and computational requirements per episode, when compared to the training algorithm. The trained algorithm's evaluation reduces to $T+1$ power flow calculations and $T$ neural network evaluations, which are almost instantaneous, resulting in a computationally efficient control procedure, which can be used either standalone or as a part of the more complex power systems' application.
\begin{algorithm} 
  \DontPrintSemicolon
    Load previously saved parameters of the trained deep Q-network $Q(s\mid \theta^Q)$\;
    Generate daily load curves for the evaluation example\;
    Initialize the subset of available actions to the action set\;
    Run the initial power flow calculation\;
    Send the initial state $s_1$ to the agent\;
    \For{$t=1,2,\ldots,T$}{
        Based on the current state $s_t$ select the action $a_t$ with the largest Q-value in the subset of available actions\;
        Update the subset of available actions\;
        Update the network configuration according to $a_t$\;
        Run the power flow calculation\;
        Retrieve the next state $s_{t+1}$\;
        Set loads for the next time interval for each bus\;
      }
  \caption{Deep Q-network evaluation}
  \label{eval_algorithm}
\end{algorithm}

\section{Numerical Results}
In this section, results and discussion are presented for benchmark test examples – \ref{sec:benchmark}, real-life large-scale distribution network – \ref{sec:realLife}, and the IEEE 33-bus radial system – \ref{sec:ieee33}, along with the choice of RL and deep learning hyperparameters. The proposed algorithms were implemented in Python, deep neural networks were modelled and trained using the PyTorch deep learning framework, and power system-related modelling and calculations were completed using OpenDSSDirect, the Python interface to OpenDSS distribution system simulation software \cite{dugan2011DSS}. The algorithms were executed on a 64-bit Windows 10 with the following hardware configuration: AMD A8-6410 APU with AMD Radeon R5 Graphics 2.00 GHz, 4 cores, and 8 GB of RAM.

\subsection{Benchmark Test Examples} \label{sec:benchmark}
\begin{figure}[htbp] 
    \centerline{\includegraphics[width=4in,trim={0 26.5cm 23cm 0},clip]{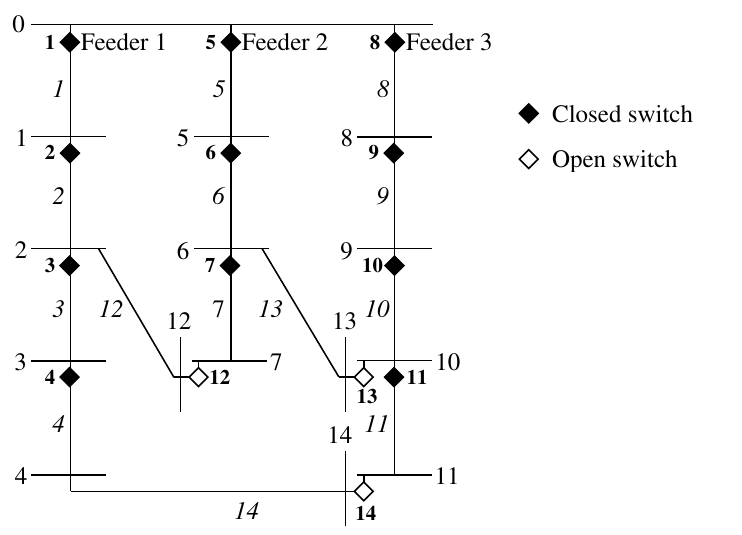}}
    \caption{Single-line diagram for $15$-bus test benchmark.}
    \label{benchmarkFig}
\end{figure}

\begin{figure}[htbp]
    \centering
    \pgfplotstableread[col sep = comma,]{./chapter_07/Figures/loadCurves.csv}\loadCurvess
    
        \begin{tikzpicture}
        
        \begin{groupplot}[group style={group size=1 by 3, vertical sep=0.5cm}, ]
        \nextgroupplot[
        yscale=0.3,
        width=5in,
        legend cell align={center},
        legend columns=2,
        legend style={
          fill opacity=0.8,
          draw opacity=1,
          text opacity=1,
          at={(0.7,0.1)},
          anchor=south,
          draw=white!80!black
        },
        scaled x ticks=manual:{}{\pgfmathparse{#1}},
        tick align=outside,
        tick pos=left,
        x grid style={white!69.0196078431373!black},
        xmin=0, xmax=23,
        xtick style={color=black},
        xticklabels={},
        y grid style={white!69.0196078431373!black},
        ylabel={\(\displaystyle \textrm{Load [MW]}\)},
        ymin=-0.05, ymax=1.4,
        ytick style={color=black},
        xmajorgrids=true,
        ymajorgrids=true,
        grid style=dotted
        ]
        \addplot [red, style={thick}]
        table [x expr=\coordindex, y={loads1}]{\loadCurvess};
        \addlegendentry{Feeder 1 load \; \;}
        
        \addplot [red, dashed, style={thick}]
        table [x expr=\coordindex, y={loads1max}]{\loadCurvess};
        \addlegendentry{Limits}
        
        \addplot [red, dashed, style={thick}]
        table [x expr=\coordindex, y={loads1min}]{\loadCurvess};
        
        \nextgroupplot[
        yscale=0.3,
        width=5in,
        legend cell align={left},
        legend columns=1,
        legend style={
          fill opacity=0.8,
          draw opacity=1,
          text opacity=1,
          at={(0.17,1.9)},
          anchor=south,
          draw=white!80!black
        },
        scaled x ticks=manual:{}{\pgfmathparse{#1}},
        tick align=outside,
        tick pos=left,
        x grid style={white!69.0196078431373!black},
        xmin=0, xmax=23,
        xtick style={color=black},
        xticklabels={},
        y grid style={white!69.0196078431373!black},
        ylabel={\(\displaystyle \textrm{Load [MW]}\)},
        ymin=-0.05, ymax=1.4,
        ytick style={color=black},
        xmajorgrids=true,
        ymajorgrids=true,
        grid style=dotted
        ]
        
        \addplot [blue, style={thick}]
        table [x expr=\coordindex, y={loads2}]{\loadCurvess};
        \addlegendentry{Feeder 2 load}
        
        \addplot [blue, dashed, style={thick}]
        table [x expr=\coordindex, y={loads2max}]{\loadCurvess};
        \addlegendentry{Limits}
        
        \addplot [blue, dashed, style={thick}]
        table [x expr=\coordindex, y={loads2min}]{\loadCurvess};
        
        \nextgroupplot[
        yscale=0.3,
        width=5in,
        legend cell align={left},
        legend columns=1,
        legend style={
          fill opacity=0.8,
          draw opacity=1,
          text opacity=1,
          at={(0.84,1.55)},
          anchor=south,
          draw=white!80!black
        },
        scaled x ticks=manual:{}{\pgfmathparse{#1}},
        tick align=outside,
        tick pos=left,
        x grid style={white!69.0196078431373!black},
        xlabel={\textrm{Hour [h]}},
        xlabel style={yshift=-5pt},
        xmin=0, xmax=23,
        xtick style={color=black},
        y grid style={white!69.0196078431373!black},
        ylabel={\(\displaystyle \textrm{Load [MW]}\)},
        ymin=-0.05, ymax=1.4,
        ytick style={color=black},
        xmajorgrids=true,
        ymajorgrids=true,
        grid style=dotted
        ]
        
        \addplot [black, style={thick}]
        table [x expr=\coordindex, y={loads3}]{\loadCurvess};
        \addlegendentry{Feeder 3 load}
        
        \addplot [black, dashed, style={thick}]
        table [x expr=\coordindex, y={loads3max}]{\loadCurvess};
        \addlegendentry{Limits}
        
        \addplot [black, dashed, style={thick}]
        table [x expr=\coordindex, y={loads3min}]{\loadCurvess};
        
        \end{groupplot}
        
        \end{tikzpicture}
    \caption{Daily load profiles for three feeders. Full lines represent average load values, and dashed lines represent limits between which training set loads are sampled.}
    \label{loadCurves}
\end{figure}
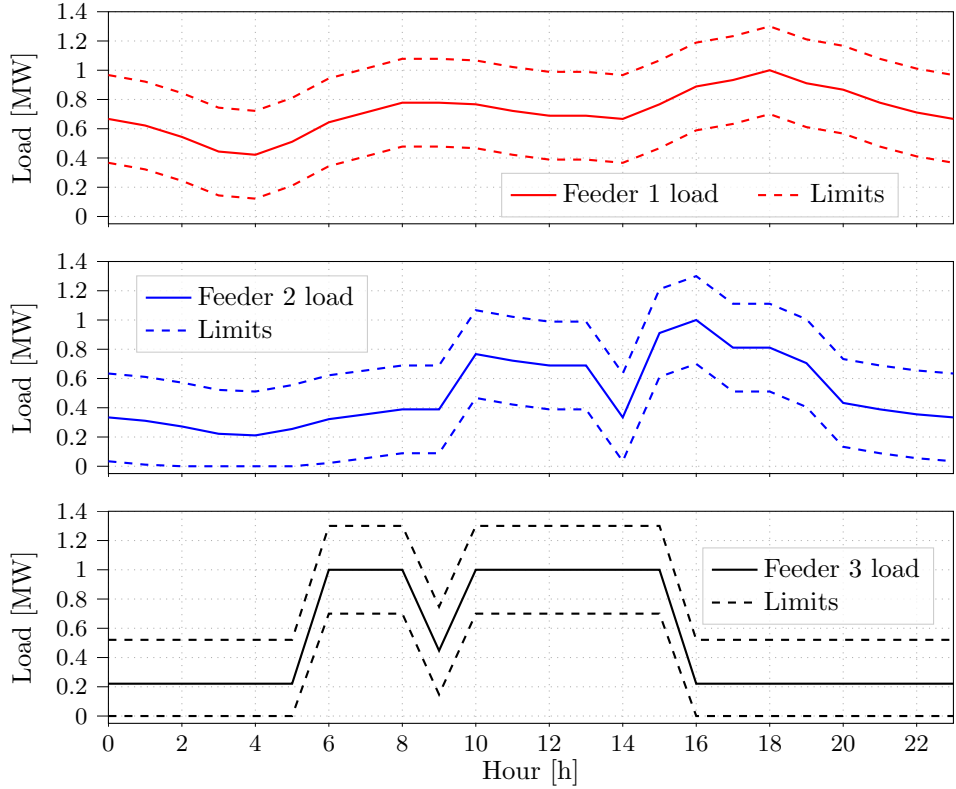

Fig.~\ref{benchmarkFig} illustrates a $15$-bus test benchmark where a slack bus is the bus with the marker $0$ and the other $14$ buses are of the PQ type. Loads are defined by the chronological daily diagrams, which are sampled uniformly from the intervals defined by dashed lines in Fig.~\ref{loadCurves}. The length of all branches is $4.5$km. All branches are balanced with the direct sequence impedance $r + jx = (0.224+j0.109) \Omega/$km. All branches have switching devices, and the number of switch manipulations is not constrained.

Penalty values used to model the bus voltage constraints in \eqref{eq:busVoltage_constraints} and the branch capacity constraints in \eqref{eq:branchCapacity_constraints} are: $C_U = C_S = 10$. The costs of energy losses and switching operations in the time interval are \cite{SHARIATKHAH20121, Yin2009Distribution}:
\begin{itemize}
    \item Cost of energy losses ($C_{Loss}$): \$6.5625 cents/kWh;
    \item Cost of switching manipulations ($C_{sw}$): \$1 per manipulation.
\end{itemize}

DQN used for this test example consists of the input layer, four hidden layers, and the output layer. The input layer has $15$ neurons, one for the time step index variable, and $14$ for the apparent powers of each switch. Each hidden layer has $256$ neurons, and the output layer consists of $186$ neurons, one for each switch combination that leads to a feasible radial configuration. Rectified Linear Unit (ReLU) activation function is applied to each of the hidden layers and the output layer. The neural network is trained using the Adam optimizer, with a learning rate of $10^{-5}$ and a mini-batch size of $128$. We experimented with adding batch normalization on several hidden layers, but it neither improved nor deteriorated the training results.

The number of training episodes we used is $60000$. During the training, loads for each hour were uniformly sampled from the intervals defined by dashed curves in Fig.~\ref{loadCurves}. The target DQN update frequency is $10$ episodes and the mini-batches for DQN training were sampled from experience replay memory, which has the capacity of $10^6$ samples. The initial value of the exploration parameter $\epsilon$ is $1$ and decreases linearly to the episode index until it reaches the value $0.1$ in $48000^\textrm{th}$ episode. The value of the discount factor $\gamma$ is $0.99$, as in \cite{mnih2015humanlevel}.

\begin{figure}[htbp] 
    \centerline{\includegraphics[width=4in,trim={0 0 0 0},clip]{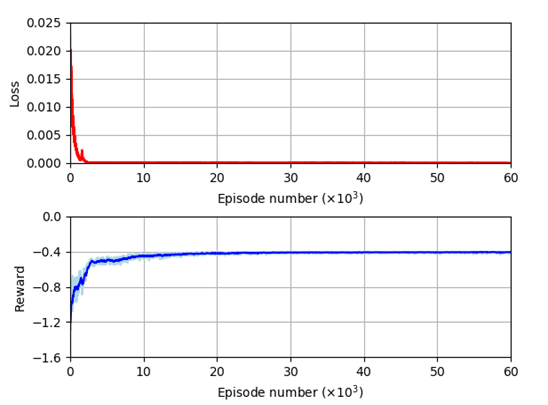}}
    \caption{Average DQN loss per episode (top) and total reward per episode along with its moving average (bottom).}
    \label{trainingConvergence}
\end{figure}

Fig.~\ref{trainingConvergence}. presents the average value of the DQN loss function, defined in \eqref{eq:lossFunction}, over the episodes. As the training advances, the DQN loss decreases, which implies that the Q-function is being approximated successfully. Additionally, the testing performance of the RL algorithm with increasing training episodes is presented in Fig.~\ref{trainingConvergence} by displaying the amount of the normalized received reward per episode and its moving average. These rewards were obtained by executing the episodes with the exploration parameter $\epsilon$ equal to zero, after each training episode. The training demonstrated asymptotic convergence within $20000$ episodes.

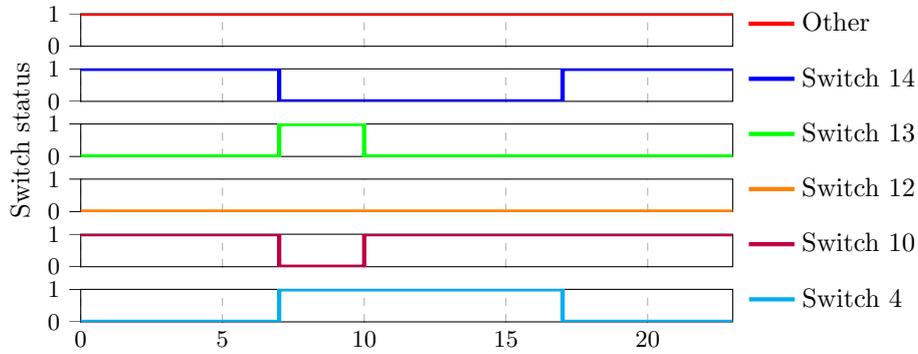
\begin{figure}[htbp]
    \centering
    \pgfplotstableread[col sep = comma,]{./chapter_07/Figures/fig5.csv}\figSwitchStatuses
    
        \begin{tikzpicture}
        
        \begin{groupplot}[group style={group size=1 by 6, vertical sep=0.3cm}, ]
        \nextgroupplot[
        yscale=0.06,
        width=4in,
        legend cell align={center},
        legend columns=2,
        legend style={
          fill opacity=0,
          draw opacity=1,
          text opacity=1,
          at={(1.12,0.1)},
          anchor=south,
          draw=white
        },
        scaled x ticks=manual:{}{\pgfmathparse{#1}},
        tick align=outside,
        tick pos=left,
        x grid style={white!69.0196078431373!black},
        xmin=0, xmax=23,
        xtick style={color=black},
        xticklabels={},
        y grid style={white!69.0196078431373!black},
        ymin=0, ymax=1,
        ytick style={color=black},
        ytick={0,1},
        xmajorgrids=true,
        grid style=dashed
        ]
        \addplot [const plot, red, style={ultra thick}]
        table [x expr=\coordindex, y={other}]{\figSwitchStatuses};
        \addlegendentry{Other}
        
        \nextgroupplot[
        yscale=0.06,
        width=4in,
        legend cell align={left},
        legend columns=1,
        legend style={
          fill opacity=0,
          draw opacity=1,
          text opacity=1,
          at={(1.155,0.1)},
          anchor=south,
          draw=white
        },
        scaled x ticks=manual:{}{\pgfmathparse{#1}},
        tick align=outside,
        tick pos=left,
        x grid style={white!69.0196078431373!black},
        xmin=0, xmax=23,
        xtick style={color=black},
        ytick={0,1},
        xticklabels={},
        y grid style={white!69.0196078431373!black},
        ymin=0, ymax=1,
        ytick style={color=black},
        xmajorgrids=true,
        grid style=dashed
        ]
        
        \addplot [const plot, blue, style={ultra thick}]
        table [x expr=\coordindex, y={sw14}]{\figSwitchStatuses};
        \addlegendentry{Switch 14}
        
        \nextgroupplot[
        yscale=0.06,
        width=4in,
        legend cell align={left},
        legend columns=1,
        legend style={
          fill opacity=0,
          draw opacity=1,
          text opacity=1,
          at={(1.155,0.1)},
          anchor=south,
          draw=white
        },
        scaled x ticks=manual:{}{\pgfmathparse{#1}},
        tick align=outside,
        tick pos=left,
        x grid style={white!69.0196078431373!black},
        xmin=0, xmax=23,
        xtick style={color=black},
        ytick={0,1},
        xticklabels={},
        y grid style={white!69.0196078431373!black},
        ylabel={\(\displaystyle \textrm{Switch status}\)},
        ymin=0, ymax=1,
        ytick style={color=black},
        xmajorgrids=true,
        grid style=dashed
        ]
        
        \addplot [const plot, green, style={ultra thick}]
        table [x expr=\coordindex, y={sw13}]{\figSwitchStatuses};
        \addlegendentry{Switch 13}

        \nextgroupplot[
        yscale=0.06,
        width=4in,
        legend cell align={left},
        legend columns=1,
        legend style={
          fill opacity=0,
          draw opacity=1,
          text opacity=1,
          at={(1.155,0.1)},
          anchor=south,
          draw=white
        },
        scaled x ticks=manual:{}{\pgfmathparse{#1}},
        tick align=outside,
        tick pos=left,
        x grid style={white!69.0196078431373!black},
        xmin=0, xmax=23,
        xtick style={color=black},
        ytick={0,1},
        xticklabels={},
        y grid style={white!69.0196078431373!black},
        ymin=0, ymax=1,
        ytick style={color=black},
        xmajorgrids=true,
        grid style=dashed
        ]
        
        \addplot [const plot, orange, style={ultra thick}]
        table [x expr=\coordindex, y={sw12}]{\figSwitchStatuses};
        \addlegendentry{Switch 12}
        
        \nextgroupplot[
        yscale=0.06,
        width=4in,
        legend cell align={left},
        legend columns=1,
        legend style={
          fill opacity=0,
          draw opacity=1,
          text opacity=1,
          at={(1.155,0.1)},
          anchor=south,
          draw=white
        },
        scaled x ticks=manual:{}{\pgfmathparse{#1}},
        tick align=outside,
        tick pos=left,
        x grid style={white!69.0196078431373!black},
        xmin=0, xmax=23,
        xtick style={color=black},
        ytick={0,1},
        xticklabels={},
        y grid style={white!69.0196078431373!black},
        ymin=0, ymax=1,
        ytick style={color=black},
        xmajorgrids=true,
        grid style=dashed
        ]
        
        \addplot [const plot, purple, style={ultra thick}]
        table [x expr=\coordindex, y={sw10}]{\figSwitchStatuses};
        \addlegendentry{Switch 10}
        
        \nextgroupplot[
        yscale=0.06,
        width=4in,
        legend cell align={left},
        legend columns=1,
        legend style={
          fill opacity=0,
          draw opacity=1,
          text opacity=1,
          at={(1.145,0.1)},
          anchor=south,
          draw=white
        },
        scaled x ticks=manual:{}{\pgfmathparse{#1}},
        tick align=outside,
        tick pos=left,
        x grid style={white!69.0196078431373!black},
        xmin=0, xmax=23,
        xtick style={color=black},
        ytick={0,1},
        y grid style={white!69.0196078431373!black},
        ymin=0, ymax=1,
        ytick style={color=black},
        xmajorgrids=true,
        grid style=dashed
        ]
        
        \addplot [const plot, cyan, style={ultra thick}]
        table [x expr=\coordindex, y={sw4}]{\figSwitchStatuses};
        \addlegendentry{Switch 4}

        \end{groupplot}
        
        \end{tikzpicture}
        \vspace*{-1mm}
    \caption{Switch status changes during the 24-hour period.}
    \label{figSwitchStatuses}
\end{figure}

Table~\ref{tbl_resultsSmallNetwork} presents load, losses, and switch status changes for the proposed approach in $24$-hour optimization horizon, as well as the comparison of the same results with the state-of-the-art method from \cite{Kovacki2018ScalableAF}. The method from \cite{Kovacki2018ScalableAF} minimizes the costs of energy losses, switching manipulations, and outages. For this comparison, the method from \cite{Kovacki2018ScalableAF} is executed with the cost of outages set to zero. A graphical representation of switch statuses is presented in Fig.~\ref{figSwitchStatuses}. The results are additionally compared with the method from \cite{ramos2005PathBased}, as is presented in Table~\ref{tbl_resultsSmallNetwork2}. By comparing the results of the proposed algorithm and the method from \cite{Kovacki2018ScalableAF}, it can be concluded that the proposed switching actions are not the same, but the total costs only differ slightly. The execution time of the proposed algorithm is $0.148$s, which is two orders of magnitude smaller than the execution time of the method from \cite{Kovacki2018ScalableAF}.

\begin{table}[]
\caption{Total load, active power losses and switch status changes in the 24-hour time optimization period for the $15$-bus test benchmark (O–open; C–close).}
\label{tbl_resultsSmallNetwork}
\begin{tabular}{|l|l|ll|ll|}
\hline
     &               & \multicolumn{2}{l|}{Proposed approach}                           & \multicolumn{2}{l|}{Method from \cite{Kovacki2018ScalableAF}}                       \\ \hline
Hour & Load {[}kW{]} & \multicolumn{1}{l|}{Losses {[}kW{]}} & \begin{tabular}[c]{@{}l@{}}Switch status\\ changes\end{tabular}  & \multicolumn{1}{l|}{Losses {[}kW{]}} & \begin{tabular}[c]{@{}l@{}}Switch status\\ changes\end{tabular} \\ \hline
0    & 4554.0        & \multicolumn{1}{l|}{114.06}          & 4(O), 14(C)               & \multicolumn{1}{l|}{131.73}          & No changes         \\ \hline
1    & 4305.0        & \multicolumn{1}{l|}{101.79}          & No changes          & \multicolumn{1}{l|}{115.32}          & No changes         \\ \hline
2    & 3876.0        & \multicolumn{1}{l|}{82.49}           & No changes          & \multicolumn{1}{l|}{89.97}           & No changes         \\ \hline
3    & 3326.0        & \multicolumn{1}{l|}{61.20}           & No changes          & \multicolumn{1}{l|}{62.95}           & No changes         \\ \hline
4    & 3205.0        & \multicolumn{1}{l|}{57.03}           & No changes          & \multicolumn{1}{l|}{57.81}           & No changes         \\ \hline
5    & 3693.0        & \multicolumn{1}{l|}{75.01}           & No changes          & \multicolumn{1}{l|}{80.35}           & No changes         \\ \hline
6    & 7542.0        & \multicolumn{1}{l|}{316.63}          & \begin{tabular}[c]{@{}l@{}}10(O), 14(O), \\ 4(C), 13(C)\end{tabular} & \multicolumn{1}{l|}{316.63}          & 10(O), 13(C)             \\ \hline
7    & 7909.0        & \multicolumn{1}{l|}{348.71}          & No changes          & \multicolumn{1}{l|}{348.71}          & No changes         \\ \hline
8    & 8279.0        & \multicolumn{1}{l|}{384.08}          & No changes          & \multicolumn{1}{l|}{384.08}          & No changes         \\ \hline
9   & 6067.0        & \multicolumn{1}{l|}{213.94}          & 10(C), 13(O)              & \multicolumn{1}{l|}{220.01}          & No changes         \\ \hline
10   & 9369.0        & \multicolumn{1}{l|}{465.82}          & No changes          & \multicolumn{1}{l|}{465.82}          & 10(C), 13(O)             \\ \hline
11   & 9054.0        & \multicolumn{1}{l|}{440.61}          & No changes          & \multicolumn{1}{l|}{440.61}          & No changes         \\ \hline
12   & 8823.0        & \multicolumn{1}{l|}{423.20}          & No changes          & \multicolumn{1}{l|}{423.20}          & No changes         \\ \hline
13   & 8823.0        & \multicolumn{1}{l|}{423.20}          & No changes          & \multicolumn{1}{l|}{423.20}          & No changes         \\ \hline
14   & 7670.0        & \multicolumn{1}{l|}{375.02}          & No changes          & \multicolumn{1}{l|}{324.18}          & 10(O), 13(C)             \\ \hline
15   & 9801.0        & \multicolumn{1}{l|}{493.87}          & No changes          & \multicolumn{1}{l|}{493.87}          & 10(C), 13(O)             \\ \hline
16   & 7440.0        & \multicolumn{1}{l|}{278.42}          & 4(O), 14(C)               & \multicolumn{1}{l|}{278.42}          & 4(O), 14(C)              \\ \hline
17   & 7049.0        & \multicolumn{1}{l|}{252.96}          & No changes          & \multicolumn{1}{l|}{252.96}          & No changes         \\ \hline
18   & 7317.0        & \multicolumn{1}{l|}{276.55}          & No changes          & \multicolumn{1}{l|}{276.55}          & No changes         \\ \hline
19   & 6643.0        & \multicolumn{1}{l|}{227.18}          & No changes          & \multicolumn{1}{l|}{227.18}          & No changes         \\ \hline
20   & 5651.0        & \multicolumn{1}{l|}{178.27}          & No changes          & \multicolumn{1}{l|}{178.27}          & No changes         \\ \hline
21   & 5163.0        & \multicolumn{1}{l|}{147.68}          & No changes          & \multicolumn{1}{l|}{147.68}          & No changes         \\ \hline
22   & 4793.0        & \multicolumn{1}{l|}{126.72}          & No changes          & \multicolumn{1}{l|}{126.72}          & No changes         \\ \hline
23   & 4554.0        & \multicolumn{1}{l|}{114.06}          & No changes          & \multicolumn{1}{l|}{114.06}          & No changes         \\ \hline
\end{tabular}
\end{table}

\begin{table}[]
\caption{Total losses, number of switch status changes, and total cost in the $24$-hour time optimization period.}
\label{tbl_resultsSmallNetwork2}
\begin{tabular}{|l|l|l|l|}
\hline
                                 & Proposed   approach & Method from \cite{Kovacki2018ScalableAF} & Method from   \cite{ramos2005PathBased} \\ \hline
Total   losses {[}kW{]}          & 5978.48             & 5980.28                & 5967.56                \\ \hline
\begin{tabular}[c]{@{}l@{}}Number of  switch\\ status changes\end{tabular} & 10                  & 10                     & 22                     \\ \hline
Total cost   {[}\${]}            & 402.3               & 402.4                  & 413.6                  \\ \hline
\end{tabular}
\end{table}

Based on Table~\ref{tbl_resultsSmallNetwork2}, it can be concluded that the proposed approach and the method from \cite{Kovacki2018ScalableAF} provide better solutions than the method from \cite{ramos2005PathBased}. Solutions from the proposed approach and the method from \cite{Kovacki2018ScalableAF} provide greater cost savings, and they achieve it with a lower number of switch manipulations.

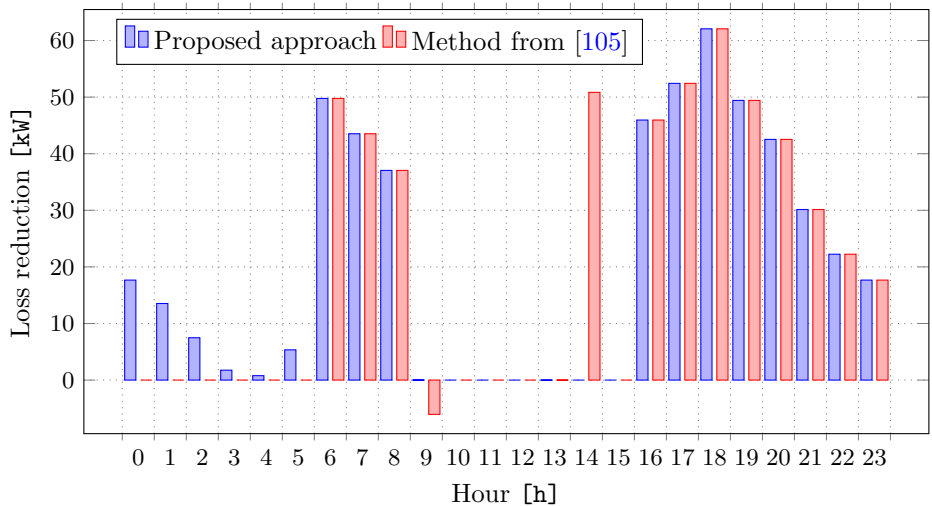
\begin{figure}[htbp]
    \centering
    \begin{tikzpicture}
    \begin{axis}[
        yscale=0.6,
        width=5in,
    	x tick label style={
    		/pgf/number format/1000 sep=},
    	xlabel=Hour \texttt{[h]},
    	xlabel style={yshift=-5pt},
    	ylabel=Loss reduction \texttt{[kW]},
    	enlargelimits=0.05,
    	legend style={at={(0.35,1.6)},
    		anchor=north,legend columns=-1},
    	ybar interval=0.7,
    	ymajorgrids=true,
    ]
    \addplot 
    	coordinates {(0,17.67) (1,13.53) (2,7.48) (3,1.75) (4,0.78) (5,5.34) (6,49.77) (7,43.53) (8,37.05) (9,0.01) (10,0) (11,0) (12,0) (13,-0.01) (14,0) (15,0) (16,45.95) (17,52.43) (18,62.07) (19,49.42) (20,42.53) (21,30.14) (22,22.24) (23,17.67) (24,0)};
    
    \addplot 
    	coordinates {(0,0) (1,0) (2,0) (3,0) (4,0) (5,0) (6,49.77) (7,43.53) (8,37.05) (9,-6.06) (10,0) (11,0) (12,0) (13,-0.01) (14,50.84) (15,0) (16,45.95) (17,52.43) (18,62.07) (19,49.42) (20,42.53) (21,30.14) (22,22.24) (23,17.67) (24,0)};
    
    \legend{Proposed approach,Method from \cite{Kovacki2018ScalableAF}}
    \end{axis}
    \end{tikzpicture}
    \vspace*{-3mm}
    \caption{Loss reduction using DDNR.}
    \label{figLossReduction}
\end{figure}

Total losses in the time period of $24$ hours for $15$-bus test benchmark without DNR are $6477.81$kW, with the total cost of $\$425.1$. With the proposed DDNR, losses are reduced to $5978.48$kW and the total cost is reduced to $\$402.3$. Fig.~\ref{figLossReduction} compares loss reduction per hour for the proposed approach and for the method from \cite{Kovacki2018ScalableAF}. In Fig.~\ref{figLossReduction} it can be seen that there is no reduction of losses in the interval from $10$ to $16$ hours. In Table~\ref{tbl_resultsSmallNetwork} it can be seen that the network configuration from 10 to 16 hours is the same as the basic network configuration, as in Fig.~\ref{benchmarkFig} (in the first hour switch 4 was opened and switch 14 was closed; in the seventh hour switch 4 was closed and switch 14 was opened and switch 10 was opened and switch 13 was closed; in the tenth hour switch 10 was closed and switch 13 was opened). Consequently, there is no reduction of losses from 10 to 16 hours. For the method from \cite{Kovacki2018ScalableAF} in the $10^{\textrm{th}}$ hour, there is a negative loss reduction (see Table~\ref{tbl_resultsSmallNetwork}). From 17 to $24$ hours, loss reduction is the same for both methods, because network configuration in that interval is the same, too.

Fig.~\ref{voltageProfile15Bus} shows the voltage profile before and after reconfiguration. It is noted that the voltage profile is improved after applying the configuration found by the proposed algorithm. Bus loads are taken from the daily load profile presented in Fig.~\ref{loadCurves} at $17^{\textrm{th}}$ hour. 

\begin{figure}[htbp]
    \centering
    \pgfplotstableread[col sep = comma,]{./chapter_07/Figures/fig8.csv}\figSwitchStatuses
    
        \begin{tikzpicture}
        
        \begin{groupplot}[group style={group size=1 by 8, vertical sep=0.3cm}, ]
        \nextgroupplot[
        yscale=0.06,
        width=4in,
        legend cell align={center},
        legend columns=2,
        legend style={
          fill opacity=0,
          draw opacity=1,
          text opacity=1,
          at={(1.12,0.1)},
          anchor=south,
          draw=white
        },
        scaled x ticks=manual:{}{\pgfmathparse{#1}},
        tick align=outside,
        tick pos=left,
        x grid style={white!69.0196078431373!black},
        xmin=0, xmax=23,
        xtick style={color=black},
        xticklabels={},
        y grid style={white!69.0196078431373!black},
        ymin=0, ymax=1,
        ytick style={color=black},
        ytick={0,1},
        xmajorgrids=true,
        grid style=dashed
        ]
        \addplot [const plot, red, style={ultra thick}]
        table [x expr=\coordindex, y={other}]{\figSwitchStatuses};
        \addlegendentry{Other}
        
        \nextgroupplot[
        yscale=0.06,
        width=4in,
        legend cell align={left},
        legend columns=1,
        legend style={
          fill opacity=0,
          draw opacity=1,
          text opacity=1,
          at={(1.155,0.1)},
          anchor=south,
          draw=white
        },
        scaled x ticks=manual:{}{\pgfmathparse{#1}},
        tick align=outside,
        tick pos=left,
        x grid style={white!69.0196078431373!black},
        xmin=0, xmax=23,
        xtick style={color=black},
        ytick={0,1},
        xticklabels={},
        y grid style={white!69.0196078431373!black},
        ymin=0, ymax=1,
        ytick style={color=black},
        xmajorgrids=true,
        grid style=dashed
        ]
        
        \addplot [const plot, blue, style={ultra thick}]
        table [x expr=\coordindex, y={sw14}]{\figSwitchStatuses};
        \addlegendentry{Switch 14}
        
        \nextgroupplot[
        yscale=0.06,
        width=4in,
        legend cell align={left},
        legend columns=1,
        legend style={
          fill opacity=0,
          draw opacity=1,
          text opacity=1,
          at={(1.155,0.1)},
          anchor=south,
          draw=white
        },
        scaled x ticks=manual:{}{\pgfmathparse{#1}},
        tick align=outside,
        tick pos=left,
        x grid style={white!69.0196078431373!black},
        xmin=0, xmax=23,
        xtick style={color=black},
        ytick={0,1},
        xticklabels={},
        y grid style={white!69.0196078431373!black},
        ylabel={\(\displaystyle \textrm{Switch status}\)},
        ymin=0, ymax=1,
        ytick style={color=black},
        xmajorgrids=true,
        grid style=dashed
        ]
        
        \addplot [const plot, green, style={ultra thick}]
        table [x expr=\coordindex, y={sw13}]{\figSwitchStatuses};
        \addlegendentry{Switch 13}

        \nextgroupplot[
        yscale=0.06,
        width=4in,
        legend cell align={left},
        legend columns=1,
        legend style={
          fill opacity=0,
          draw opacity=1,
          text opacity=1,
          at={(1.155,0.1)},
          anchor=south,
          draw=white
        },
        scaled x ticks=manual:{}{\pgfmathparse{#1}},
        tick align=outside,
        tick pos=left,
        x grid style={white!69.0196078431373!black},
        xmin=0, xmax=23,
        xtick style={color=black},
        ytick={0,1},
        xticklabels={},
        y grid style={white!69.0196078431373!black},
        ymin=0, ymax=1,
        ytick style={color=black},
        xmajorgrids=true,
        grid style=dashed
        ]
        
        \addplot [const plot, orange, style={ultra thick}]
        table [x expr=\coordindex, y={sw12}]{\figSwitchStatuses};
        \addlegendentry{Switch 12}
        
        \nextgroupplot[
        yscale=0.06,
        width=4in,
        legend cell align={left},
        legend columns=1,
        legend style={
          fill opacity=0,
          draw opacity=1,
          text opacity=1,
          at={(1.155,0.1)},
          anchor=south,
          draw=white
        },
        scaled x ticks=manual:{}{\pgfmathparse{#1}},
        tick align=outside,
        tick pos=left,
        x grid style={white!69.0196078431373!black},
        xmin=0, xmax=23,
        xtick style={color=black},
        ytick={0,1},
        xticklabels={},
        y grid style={white!69.0196078431373!black},
        ymin=0, ymax=1,
        ytick style={color=black},
        xmajorgrids=true,
        grid style=dashed
        ]
        
        \addplot [const plot, cyan, style={ultra thick}]
        table [x expr=\coordindex, y={sw11}]{\figSwitchStatuses};
        \addlegendentry{Switch 11}
        
        \nextgroupplot[
        yscale=0.06,
        width=4in,
        legend cell align={left},
        legend columns=1,
        legend style={
          fill opacity=0,
          draw opacity=1,
          text opacity=1,
          at={(1.155,0.1)},
          anchor=south,
          draw=white
        },
        scaled x ticks=manual:{}{\pgfmathparse{#1}},
        tick align=outside,
        tick pos=left,
        x grid style={white!69.0196078431373!black},
        xmin=0, xmax=23,
        xtick style={color=black},
        ytick={0,1},
        xticklabels={},
        y grid style={white!69.0196078431373!black},
        ymin=0, ymax=1,
        ytick style={color=black},
        xmajorgrids=true,
        grid style=dashed
        ]
        
        \addplot [const plot, purple, style={ultra thick}]
        table [x expr=\coordindex, y={sw10}]{\figSwitchStatuses};
        \addlegendentry{Switch 10}
        
        \nextgroupplot[
        yscale=0.06,
        width=4in,
        legend cell align={left},
        legend columns=1,
        legend style={
          fill opacity=0,
          draw opacity=1,
          text opacity=1,
          at={(1.145,0.1)},
          anchor=south,
          draw=white
        },
        scaled x ticks=manual:{}{\pgfmathparse{#1}},
        tick align=outside,
        tick pos=left,
        x grid style={white!69.0196078431373!black},
        xmin=0, xmax=23,
        xtick style={color=black},
        ytick={0,1},
        xticklabels={},
        y grid style={white!69.0196078431373!black},
        ymin=0, ymax=1,
        ytick style={color=black},
        xmajorgrids=true,
        grid style=dashed
        ]
        
        \addplot [const plot, brown, style={ultra thick}]
        table [x expr=\coordindex, y={sw7}]{\figSwitchStatuses};
        \addlegendentry{Switch 7}
        
        \nextgroupplot[
        yscale=0.06,
        width=4in,
        legend cell align={left},
        legend columns=1,
        legend style={
          fill opacity=0,
          draw opacity=1,
          text opacity=1,
          at={(1.145,0.1)},
          anchor=south,
          draw=white
        },
        scaled x ticks=manual:{}{\pgfmathparse{#1}},
        tick align=outside,
        tick pos=left,
        x grid style={white!69.0196078431373!black},
        xmin=0, xmax=23,
        xtick style={color=black},
        ytick={0,1},
        y grid style={white!69.0196078431373!black},
        ymin=0, ymax=1,
        ytick style={color=black},
        xmajorgrids=true,
        grid style=dashed
        ]
        
        \addplot [const plot, violet, style={ultra thick}]
        table [x expr=\coordindex, y={sw4}]{\figSwitchStatuses};
        \addlegendentry{Switch 4}

        \end{groupplot}
        
        \end{tikzpicture}
        \vspace*{-1mm}
    \caption{Switch status changes during the 24-hour period when the maximal number of switch manipulations is two.}
    \label{figSwitchStatusesConstraintTwo}
\end{figure}
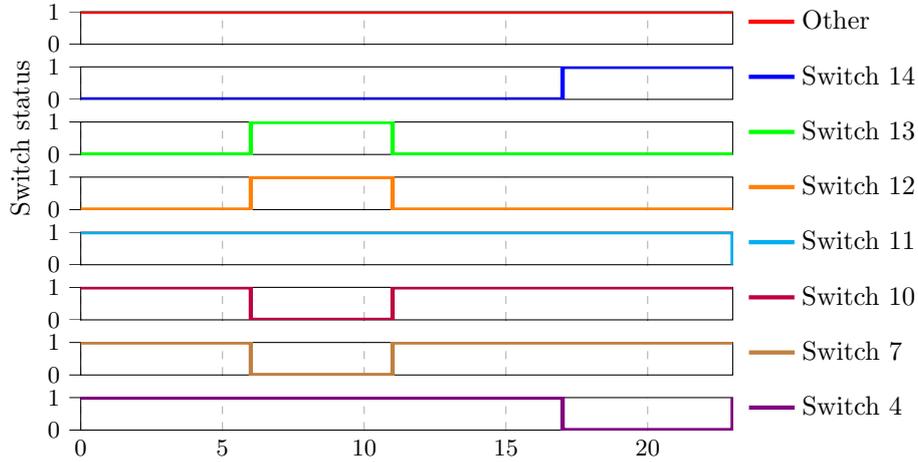

Next, we present the results of the proposed approach when the number of switch manipulations is limited to two. A graphical representation of the switch status changes in this case is given in Fig.~\ref{figSwitchStatusesConstraintTwo}. The cost of losses is equal to \$403.5, while the switching manipulation cost is \$12, summing up to the total cost of \$415.5, which means that this solution is more expensive than the solution presented in Table~\ref{tbl_resultsSmallNetwork}.

\begin{figure}[htbp]
    \centering
    \pgfplotstableread[col sep = comma,]{./chapter_07/Figures/voltagesBenchmark.csv}\testResultColumns
    \begin{tikzpicture}
    
    \begin{axis}[
    width = 5in,
    yscale=0.5,
    legend cell align={left},
    legend columns=1,
    legend style={
      fill opacity=0.8,
      font=\small,
      draw opacity=1,
      text opacity=1,
      at={(0.63,1.85)},
      anchor=north west,
      draw=white!80!black
    },
    tick align=outside,
    tick pos=left,
    x grid style={white!69.0196078431373!black},
    xlabel={Bus index},
    xlabel style={yshift=-5pt},
    xmin=-0.3, 
    xmax=14.3,
    xtick style={color=black},
    y grid style={white!69.0196078431373!black},
    ylabel={Bus voltages [p.u.]},
    ytick style={color=black},
        yticklabel style={
        /pgf/number format/fixed,
        /pgf/number format/precision=5
    },
    scaled y ticks=false,
    xmajorgrids=true,
    ymajorgrids=true,
    grid style=dashed
    ]
    \addplot [red]
    table [x expr=\coordindex, y={baseCase}]{\testResultColumns};
    \addlegendentry{Base case}
    \addplot [blue]
    table [x expr=\coordindex, y={after}]{\testResultColumns};
    \addlegendentry{After reconfiguration}
    
    \end{axis}
    
    \end{tikzpicture}
    \vspace{-3mm} 
    \caption{Voltage profile for 15-bus test benchmark.}
    \label{voltageProfile15Bus}
\end{figure}
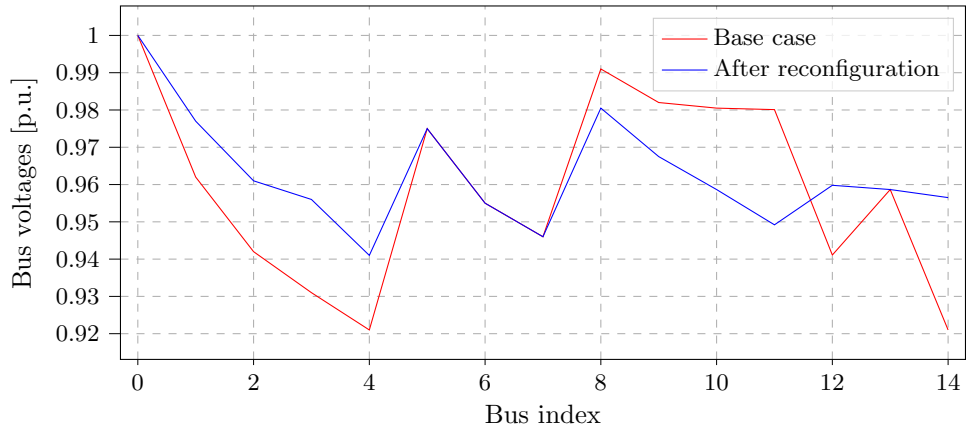

\subsection{Real-Life Large-Scale Distribution Network} \label{sec:realLife}
To assess the scalability of the proposed approach, we evaluate its results using a real-life large-scale radial distribution network, which consists of four feeders, 1015 branches, and 1008 loads. The distribution network is equipped with 31 remotely controlled switches, where 24 of them are normally closed and seven are normally open. This large number of remotely controlled switches is used to demonstrate how well does the developed algorithm scale with the increase in the number of decision variables. DQN used for this test example consists of the input layer, four hidden layers, and the output layer. The input layer has 32 neurons, one for the time step index variable and 31 for the apparent powers of each switch. Each hidden layer has 1024 neurons, and the output layer consists of 3567 neurons, one for each switch combination that leads to a feasible radial configuration. The algorithm was trained on 100000 episodes. The initial value of the exploration parameter $\epsilon$ is 1 and decreases linearly to the episode index, until it reaches the value 0.1 in $80000^{\mathrm{th}}$ episode. The rest of the parameters are the same as in the case of the 15-bus test benchmark network. Costs of energy losses and switching manipulations are set to the same values as in Section \ref{sec:benchmark}.

Table \ref{tbl_largeNetwork} presents the resulting losses and switch manipulations of the proposed approach for the 24-hour optimization horizon. The execution time of the proposed algorithm is 3.635s, which is two orders of magnitude smaller than the execution time of the method from \cite{Kovacki2018ScalableAF} when used for networks with 31 remotely controlled switches. The training of the RL algorithm demonstrated convergence after 30000 episodes, as evidenced by the analysis of its test time performance after each training episode.

\begin{table}[] 
\caption{Active power losses and switch status changes in the 24-hour time optimization period for the large-scale radial distribution network (O–open; C–close).}
\centering
\begin{tabular}{|l|l|l|}
\hline
Hour & Losses   {[}kW{]} & Switch status changes                 \\ \hline
0   & 182.30            & 238(O),   900 (O), 994(O), 1011 (C), 1013(C), 1014 (C) \\ \hline
1    & 160.98            & No changes                                     \\ \hline
2    & 127.19            & No changes                                     \\ \hline
3    & 92.46             & No changes                                     \\ \hline
4    & 86.08             & No changes                                     \\ \hline
5    & 196.43            & No changes                                     \\ \hline
6    & 742.05            & 695(O),   742(O), 900 (C), 1015 (C)                    \\ \hline
7    & 783.53            & No changes                                     \\ \hline
8    & 826.54            & No changes                                     \\ \hline
9    & 368.21            & 947   (O), 1014 (O), 695 (C), 994 (C)                  \\ \hline
10   & 1008.27           & No changes                                     \\ \hline
11   & 965.66            & No changes                                     \\ \hline
12   & 932.39            & No changes                                     \\ \hline
13   & 891.17            & 1011   (O), 1013 (O), 238 (C), 947 (C)                 \\ \hline
14   & 746.81            & No changes                                     \\ \hline
15   & 1044.24           & No changes                                     \\ \hline
16   & 786.83            & 947   (O), 1013 (C)                                    \\ \hline
17   & 553.33            & No changes                                     \\ \hline
18   & 744.87            & No changes                                     \\ \hline
19   & 645.53            & 1015   (O), 1014 (C)                                   \\ \hline
20   & 509.13            & 191   (O), 1010 (C)                                    \\ \hline
21   & 455.83            & No changes                                     \\ \hline
22   & 313.87            & No changes                                     \\ \hline
23   & 178.78            & No changes                                     \\ \hline
\end{tabular}
\label{tbl_largeNetwork}
\end{table}

The total active power losses are equal to 13342.49 kW, with a cost of \$875.6. The switching manipulation cost is equal to \$24, which sums up to the total cost of \$899.6. The results demonstrate that the proposed DDNR is applicable to real-life large-scale distribution networks.

\subsection{IEEE 33-bus Radial System} \label{sec:ieee33}

\begin{figure}[htbp] 
    \centerline{\includegraphics[width=5in,trim={0 0cm 0cm 0},clip]{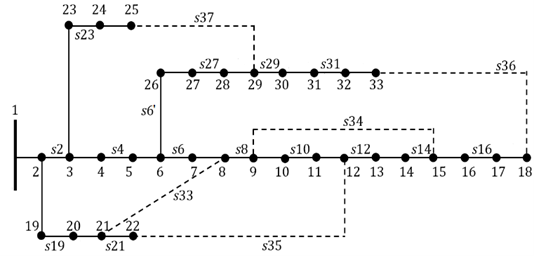}}
    \caption{IEEE 33‐bus radial system.}
    \label{figIeee33}
\end{figure}

In this section, a numerical test is performed by using the IEEE 33‐bus radial system \cite{Kashem2000ANovel}, displayed in Fig.~\ref{figIeee33}. We used the same branch resistances, reactances, and peak loads as in \cite{Kashem2000ANovel}. The loads between buses 2 and 18 are scaled with the blue full line in Fig.~\ref{loadCurves}, loads between buses 19 and 25 are scaled with the black full line in Fig.~\ref{loadCurves}, and the loads between buses 26 and 33 are scaled with the red full line in Fig.~\ref{loadCurves}. This network is modified to match the developed DDNR. The IEEE33 test system is equipped with 20 remotely controlled switches, where 15 of them are normally closed and 5 are normally open, as shown in Fig.~\ref{figIeee33}. Remotely controlled switches are marked with "s" and the unique index. If the remotely controlled switch is placed on the full line, it is closed, otherwise, it is open. Five lines were added to the original scheme and are presented with dashed lines in Fig.~\ref{figIeee33}. Costs of energy losses and switching manipulations are the same as in Section \ref{sec:benchmark}.

The algorithm was trained with the same training hyperparameters as in the case of the 15-bus test benchmark network. Table \ref{tbl_ieee33results} presents the losses and switch manipulations for the proposed approach in the 24-hour time optimization period. The execution time of the algorithm is 0.272s.

\begin{table}[] 
\caption{Active power losses and switch status changes in the 24-hour time optimization period for the IEEE 33-bus radial system (O–open; C–close).}
\centering
\begin{tabular}{|l|l|l|}
\hline
Hour & Losses   {[}kW{]} & Switch status changes \\ \hline
0   & 34.87             & 6   (O), 35 (C)            \\ \hline
1   & 30.41             & No changes         \\ \hline
2   & 23.50             & No changes         \\ \hline
3   & 16.03             & No changes         \\ \hline
4   & 14.60             & No changes         \\ \hline
5   & 20.85             & No changes         \\ \hline
6   & 56.49             & No changes         \\ \hline
7   & 62.11             & 8   (O), 33 (C)            \\ \hline
8   & 70.44             & No changes         \\ \hline
9   & 50.05             & No changes         \\ \hline
10  & 95.42             & 14   (O), 34 (C)           \\ \hline
11  & 87.05             & No changes         \\ \hline
12  & 81.22             & No changes         \\ \hline
13  & 81.22             & No changes         \\ \hline
14  & 57.06             & No changes         \\ \hline
15  & 108.47            & No changes         \\ \hline
16  & 86.87             & 27   (O), 37 (C)           \\ \hline
17  & 75.20             & No changes         \\ \hline
18  & 81.43             & No changes         \\ \hline
19  & 65.80             & No changes         \\ \hline
20  & 47.51             & No changes         \\ \hline
21  & 38.90             & No changes         \\ \hline
22  & 33.00             & No changes         \\ \hline
23  & 29.45             & No changes         \\ \hline
\end{tabular}
\label{tbl_ieee33results}
\end{table}

\begin{figure}[htbp]
    \centering
    \pgfplotstableread[col sep = comma,]{./chapter_07/Figures/voltagesIEEE33.csv}\testResultColumns
    \begin{tikzpicture}
    
    \begin{axis}[
    width = 5in,
    yscale=0.5,
    legend cell align={left},
    legend columns=1,
    legend style={
      fill opacity=0.8,
      font=\small,
      draw opacity=1,
      text opacity=1,
      at={(0.19,1.85)},
      anchor=north west,
      draw=white!80!black
    },
    tick align=outside,
    tick pos=left,
    x grid style={white!69.0196078431373!black},
    xlabel={Bus index},
    xlabel style={yshift=-5pt},
    xmin=-0.9, 
    xmax=32.45,
    xtick style={color=black},
    y grid style={white!69.0196078431373!black},
    ylabel={Bus voltages [p.u.]},
    ytick style={color=black},
        yticklabel style={
        /pgf/number format/fixed,
        /pgf/number format/precision=5
    },
    scaled y ticks=false,
    xmajorgrids=true,
    ymajorgrids=true,
    grid style=dashed
    ]
    \addplot [red]
    table [x expr=\coordindex, y={baseCase}]{\testResultColumns};
    \addlegendentry{Base case}
    \addplot [blue]
    table [x expr=\coordindex, y={after}]{\testResultColumns};
    \addlegendentry{After reconfiguration}
    
    \end{axis}
    
    \end{tikzpicture}
    \vspace{-3mm} 
    \caption{Voltage profile for IEEE 33-bus radial system.}
    \label{voltagesIEEE33}
\end{figure}
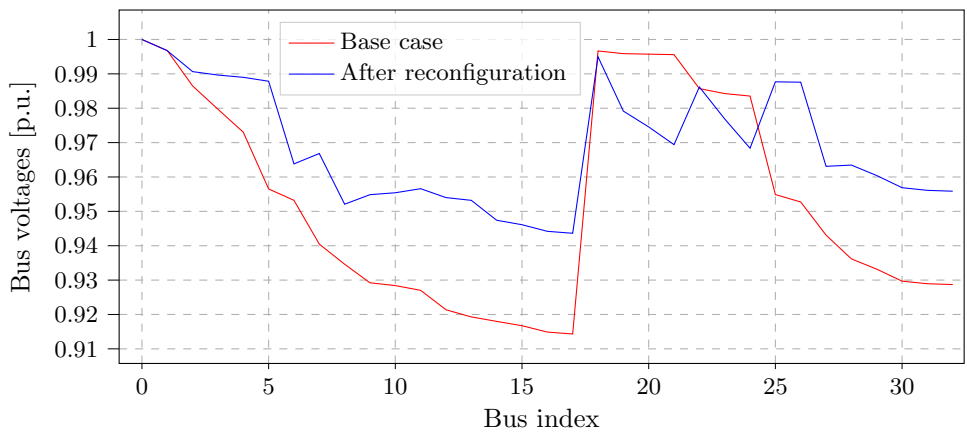

Total active power losses are equal to 1347.95 kW, with a cost of \$88.5. The cost of switching manipulation is equal to \$8, which sums up to the total cost of \$96.5. The RL algorithm was found to converge after 15000 training episodes, as determined by an analysis of the test time performance with increasing episodes. Fig.~\ref{voltagesIEEE33} shows the voltage profile before and after the reconfiguration. It is noted that in most of the buses, the voltage profile is enhanced after applying the configuration found by the proposed algorithm. Bus loads are taken from the daily load profile presented in Fig.~\ref{loadCurves} at $17^{\mathrm{th}}$ hour.

\section{Summary and future work}

In this chapter, we present a multi-objective formulation of the DDNR aimed at reducing the total cost of energy losses and switching operations. Our solution is based on DRL, and we demonstrate that the proposed definition of state variables, which demands lower observability, enables successful treatment of the DDNR problem within the RL framework. Additionally, the proposed computationally efficient method of addressing switching operation constraints by creating a subset of available actions and updating it during the episode was proven to be effective for this problem. The ideas from the suggested way of modelling the DDNR problem as MDP can be used for solving similar power system control and optimization problems as well. Once trained, the RL algorithm demonstrates faster execution compared to the state-of-the-art method, while yielding approximately equal total cost savings. The presented results indicate that the developed algorithm is scalable, i.e., the computation time does not increase exponentially with the problem dimension. Therefore, the developed DDNR algorithm can handle large-scale real-life distribution networks well.

The major drawback of applying typical RL algorithms in power systems is the need for the usage of power system simulators that enable the exploration process of an RL agent. Training an RL agent on a real-world power system is not a common practice due to the high cost of exploration actions. On the other hand, power system simulators do not represent the real-world power system ideally, which can cause the lower performance of a trained RL model when deployed in reality. A possible solution and a topic for future work could be using offline reinforcement learning \cite{offlineRL}, which eliminates the need for using the action exploration, by optimizing the policy based on the variety of historical actions, collected during the real-world distribution network operation. Another promising area of research is using safe reinforcement learning \cite{Garca2015ACS} and worst-case reinforcement learning \cite{Yang2021WCSACWS}, which enables safe exploration during the training while guaranteeing the constraint satisfaction, therefore causing no harm to the power system.

\chapter{Conclusions}	

Deep learning has demonstrated great potential to improve various tasks in power systems, including monitoring tasks such as stability assessment and fault detection, as well as for optimization tasks like Volt-Var optimization, optimal power flow, etc. One of the current trends in the field is the use of GNNs and DRL. In this thesis, we applied these methods to SE and DDNR problems and shown that these methods exhibit high levels of accuracy and improved performance when compared to traditionally used techniques. As the field continues to evolve, it is expected that more research and development will be conducted in these areas, with a focus on implementing these techniques in real-world power systems to demonstrate their practical potential.

This thesis presents two main contributions. As the first one, we investigate how GNN can be used as fast and accurate solvers of linear and nonlinear SE. The proposed graph attention network-based model, specialized for the newly introduced heterogeneous augmented factor graphs, recursively propagates the input measurements from the factor nodes to generate predictions in the variable nodes. Evaluating the trained model on unseen data samples confirms that the proposed GNN approach can be used as a highly accurate approximator of the traditional SE solutions, with the added benefit of linear computational complexity at inference time. The model is robust in unobservable scenarios that are not solvable using traditional SE and deep learning methods, such as when individual measurements or entire measurement units fail to deliver measurement data to the proposed SE solver. Furthermore, the GNN model performs well when measurement variances are high or outliers are present in the input data. The proposed approach demonstrates scalability and sample efficiency when tested on power systems of various sizes, as it makes good predictions even when trained on a small number of randomly generated samples. Finally, the proposed GNN model outperforms the more conventional deep learning-based SE approach in terms of prediction accuracy and significantly lower number of trainable parameters, especially as the size of the power system grows. In this work, we focused on using GNNs to solve a linear and nonlinear transmission system SE model. However, the proposed learning framework, graph augmentation techniques, and conclusions can be applied to a wide range of SE formulations. For example, the GNN's ability to provide relevant solutions in underdetermined scenarios suggests that it could be useful for GNN-based SE in highly unobservable distribution systems.

As a second main contribution, this thesis has explored a multi-objective formulation for DDNR, aimed at minimizing the total cost of energy losses and switching operations. The proposed solution, based on DRL, demonstrated successful treatment of DDNR as an MDP through the use of state variables with reduced observability requirements and a computationally efficient approach for handling switching operation constraints. The results showed that the developed algorithm is scalable, performs faster than the state-of-the-art method while yielding comparable cost savings, and is capable of handling large-scale real-world distribution networks. This work contributes to the field of power system control and optimization by providing a novel and effective solution for DDNR.


\small
\cleardoublepage
\newpage
\phantomsection
\addcontentsline{toc}{chapter}{Bibliography}
\bibliographystyle{IEEEtran} 
\bibliography{thesis_bibliography}


\end{document}